\documentclass{article}

\usepackage{microtype}
\usepackage{graphicx}
\usepackage{subcaption}

\usepackage{wrapfig}

\usepackage{multirow}
\usepackage{booktabs} % for professional tables

\usepackage{hyperref}

\usepackage[accepted]{icml2026}

\usepackage{amsmath}
\usepackage{amssymb}
\usepackage{mathtools}
\usepackage{amsthm}

\usepackage{threeparttable}
\usepackage[table]{xcolor} % cellcolor

\definecolor{RowColor}{RGB}{222,239,247}

\usepackage[capitalize,noabbrev]{cleveref}

\theoremstyle{plain}

\theoremstyle{definition}

\theoremstyle{remark}

\usepackage[textsize=tiny]{todonotes}

\newcommand{\R}{\mathbb{R}}

\newcommand{\norm}[1]{\left\|#1\right\|}
\DeclarePairedDelimiterX{\inner}[1]{\langle}{\rangle}{#1}
\newcommand{\sign}{\text{sign}}

\newcommand{\vh}{\mathbf{h}}

\newcommand{\vk}{\mathbf{k}}

\newcommand{\vm}{\mathbf{m}}

\newcommand{\vp}{\mathbf{p}}
\newcommand{\vq}{\mathbf{q}}
\newcommand{\vr}{\mathbf{r}}
\newcommand{\vs}{\mathbf{s}}

\newcommand{\vu}{\mathbf{u}}
\newcommand{\vv}{\mathbf{v}}
\newcommand{\vw}{\mathbf{w}}

\newcommand{\mH}{\mathbf{H}}
\newcommand{\mI}{\mathbf{I}}

\newcommand{\mK}{\mathbf{K}}

\newcommand{\mM}{\mathbf{M}}

\newcommand{\mP}{\mathbf{P}}

\newcommand{\mR}{\mathbf{R}}

\icmltitlerunning{Training-Free Hashing-Based Attention via Binary Principal Components}

\begin{document}

\twocolumn[
  \icmltitle{Training-Free Hashing-Based Attention via Binary Principal Components}



  \icmlsetsymbol{equal}{*}

  \begin{icmlauthorlist} 
    \icmlauthor{Daohai Yu}{sch}
    \icmlauthor{Zhanpeng Zeng}{sch}
    \icmlauthor{Keyu Chen}{comp}
    \icmlauthor{Wenhao Li}{sch}
    \icmlauthor{Zhifeng Shen}{comp}

    \icmlauthor{Luxi Lin}{sch}
    \icmlauthor{Ruizhi Qiao}{comp}
    \icmlauthor{Xing Sun}{comp}
    \icmlauthor{Rongrong Ji}{sch,sch1}
  \end{icmlauthorlist}

  % \icmlaffiliation{sch}{Department of XXX, University of YYY, Location, Country}
  \icmlaffiliation{sch}{Key Laboratory of Multimedia Trusted Perception and Efficient Computing, Ministry of Education of China, Xiamen University, 361005, P.R. China.}

  \icmlaffiliation{comp}{Tencent YouTu Lab, Shenzhen, China}
  \icmlaffiliation{sch1}{Sino-Russian Research Center for Digital Economy}
  % Sino-Russian Research Center for Digital Economy

  \icmlcorrespondingauthor{Zhanpeng Zeng}{zzeng@xmu.edu.cn}

  \icmlkeywords{Machine Learning, ICML}

  \vskip 0.3in
]



\printAffiliationsAndNotice{}  % no special notice (required even if empty)

\begin{abstract}
  
Long-context large language models (LLMs) are increasingly deployed in real-world applications, yet self-attention remains a major efficiency bottleneck -- especially during decoding -- due to the necessity of repeatedly processing ever-growing key-value (KV) caches. 
Existing sparse attention reduce computation by attending to fewer KV pairs, but often suffer from substantial accuracy degradation, require additional training, or rely on expensive hashing.
In this work, we present \textbf{BinaryPC}, a training-free, data-aware hashing-based sparse attention for long-context LLMs. BinaryPC constructs compact binary hash codes and corresponding hash function by computing binary principal components of data. 
Unlike Locality-Sensitive Hashing (LSH) with data-independent random projections or learned non-linear hashing methods, BinaryPC constructs binary codes that explicitly preserve the structural information of data without requiring gradient-based training.
Comprehensive experiments across multiple model families and long-context benchmarks show that BinaryPC preserves accuracy relative to full attention while achieving superior performance among sparse and hashing-based baselines.
On modern GPUs, BinaryPC improves end-to-end decoding throughput by 3.56$\times$ over the FlashAttention kernel.
Our code is available at 
\href{https://github.com/yudaohai666/BPC}{https://github.com/yudaohai666/BPC}.
\end{abstract}

\section{Introduction}
\begin{figure}[t]
\centering
\includegraphics[width=0.8\linewidth]{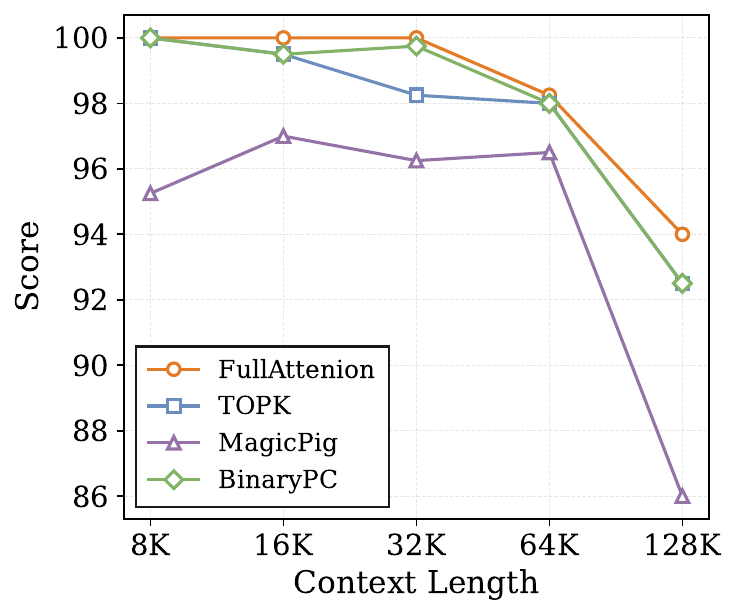}
\caption{Performance comparison on the RULER~\cite{hsieh2024ruler} NIAH multi-value task. Oracle TOPK selects top-$k$ keys using exact full-precision attention with 2\% budget.}
\label{fig:motivate-1}
\vspace{-0.2in}
\end{figure}

% 介绍，引出解码阶段是主要瓶颈

With the rapid advancement of large language models (LLMs), handling long-range dependencies has become essential for applications such as multi-document question answering~\cite{wang2024leave}, conversational agents, and complex reasoning 
tasks~\cite{achiam2023gpt,anthropic2024claude,yang2025qwen3}. 
% anthropic2024claude
Modern LLM inference is typically divided into two stages: the prefill stage, where input tokens are processed in parallel to construct the key-value (KV) cache~\cite{pope2023efficiently}, and the decoding stage, where tokens are generated autoregressively. 
% In long-context scenarios, the decoding stage suffers from significant latency overhead as it repeatedly accesses and processes an ever-expanding KV cache, making it the primary bottleneck in inference pipelines.
Compared to the highly parallel prefill stage, decoding involves frequent memory transfers for the growing KV cache while per-token generation remains inherently sequential. 
This results in low GPU utilization and suboptimal hardware efficiency, severely limiting throughput for long sequences~\cite{he2024fastdecode}. 
% For instance, when deploying Llama-3.1-8B-Instruct~\cite{grattafiori2024llama} with a 128K context on data-center GPUs, the prefill stage reaches a throughput of approximately 2024 tokens per second, whereas the decoding stage is constrained to only 30.54 tokens per second—revealing a substantial throughput gap and exposing decoding as the dominant bottleneck in long-context inference. 

% 先前的研究，Snapkv,Quest
% 为缓解这一问题，先前研究探索了多种稀疏注意力机制以减少解码阶段的计算开销。
To mitigate this issue, prior research has explored a range of sparse attention mechanisms that reduce computational overhead during decoding. 
Static selection strategies~\cite{ge2024model, li2024snapkv, cai2025pyramidkv, qin2025cake, lin2025compresskv} typically aggregate or prune KV entries after prefilling, enabling subsequent decoding to operate on a compressed representation. 
Alternative methods maintain a fixed memory budget during decoding by continuously discarding or down-weighting less critical tokens~\cite{zhang2023h2o, oren2024transformers, xiao2024efficient, adnan2024keyformer}. 
Additionally, the query-aware selection~\cite{tang2024quest} dynamically identifies salient token subsets at each decoding step, using heuristics derived from the current query.
However, these techniques often incur performance degradation, as query-agnostic eviction can remove relevant evidence, while heuristic saliency estimates may fail to align with true attention affinity.

\begin{figure}[t]
\centering
\begin{minipage}{0.32\linewidth}
    \centering
    \includegraphics[width=\textwidth]{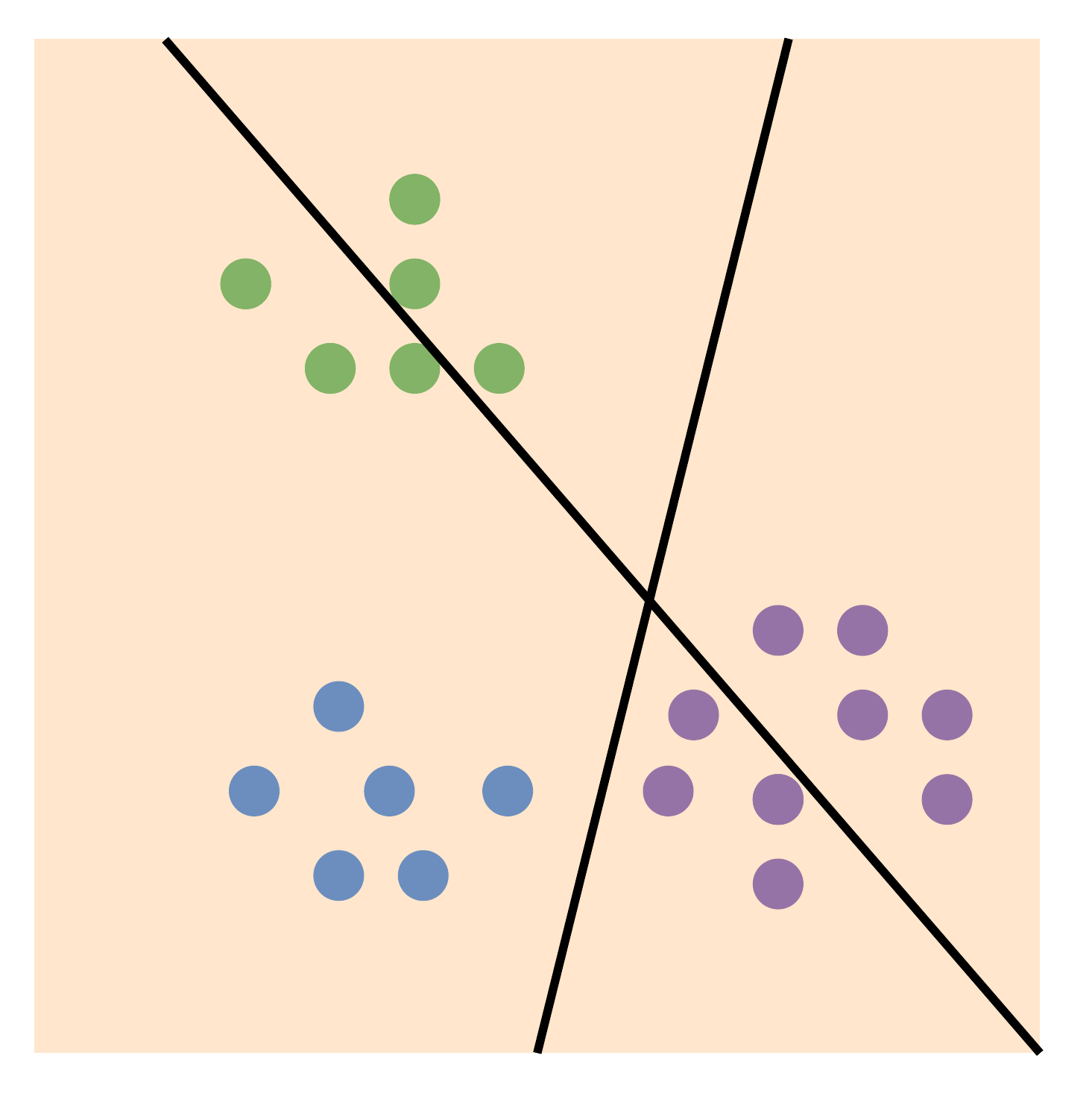}
    \subcaption{}
    \label{fig:motivate-2}
\end{minipage}
\hfill
\begin{minipage}{0.32\linewidth}
    \centering
    \includegraphics[width=\textwidth]{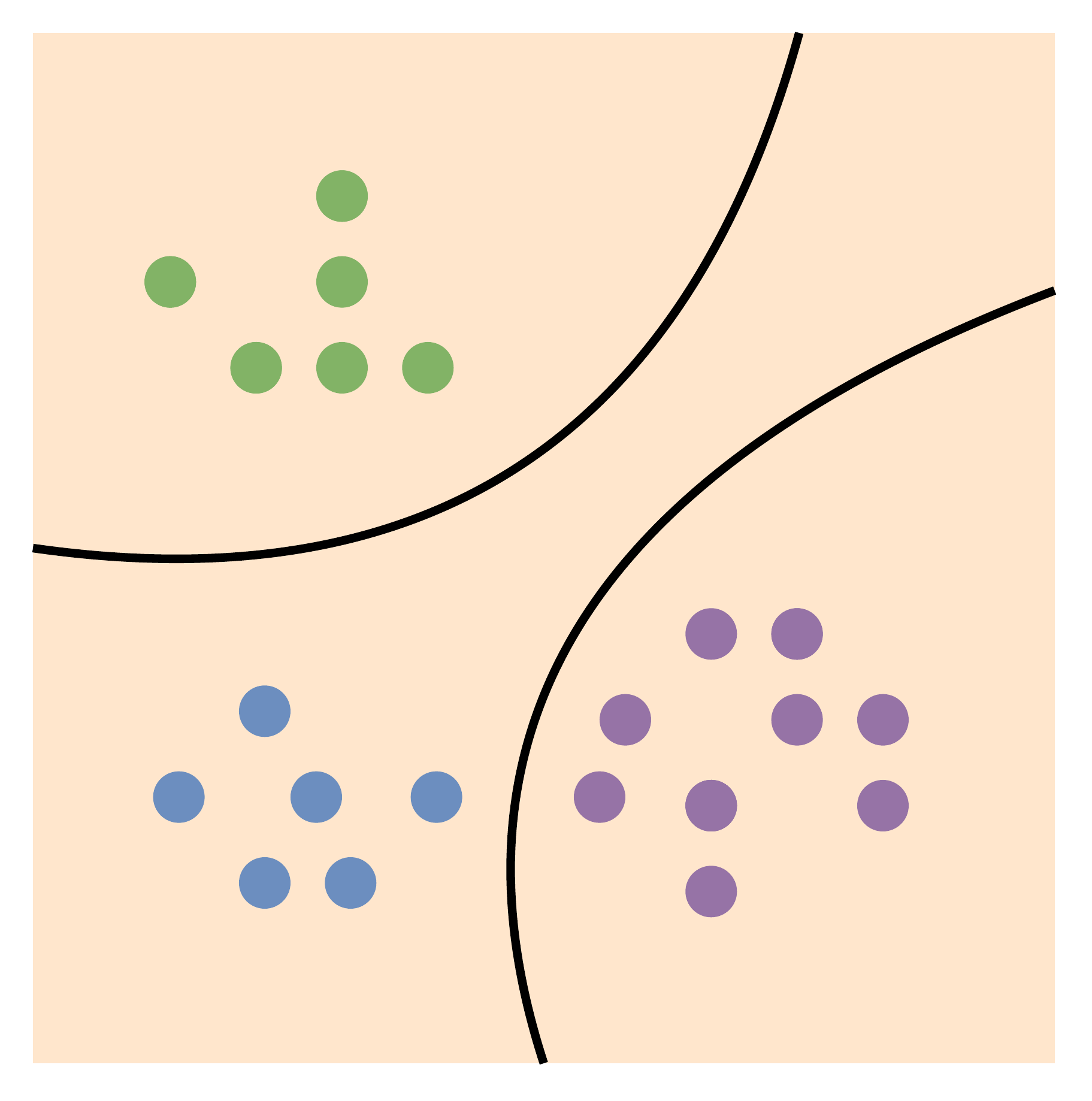}
    \subcaption{}
    \label{fig:motivate-3}
\end{minipage}
\hfill
\begin{minipage}{0.32\linewidth}
    \centering
    \includegraphics[width=\textwidth]{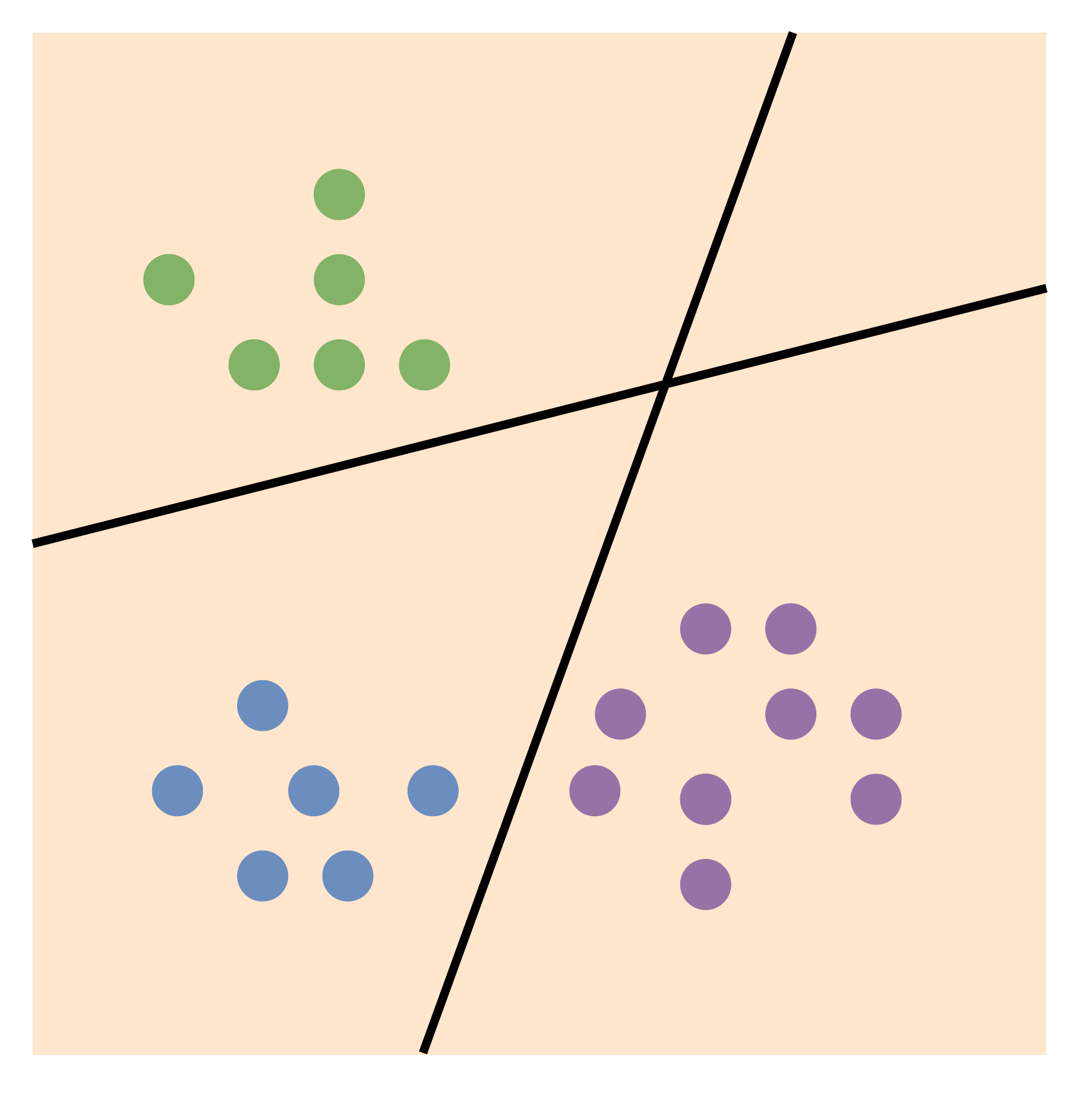}
    \subcaption{}
    \label{fig:motivate-4}
\end{minipage}
\caption{\textbf{Illustration of different hashing methods.} (a) Locality-Sensitive Hashing with random projections~\cite{chen2025magicpig}; (b) Learned hashing with trained mappings~\cite{li2025spotlight}; (c) BinaryPC with binary principal directions.}
% 中文注释：\textbf{哈希方法的潜在空间可视化。}(a) 使用随机投影的 LSH；(b) 使用训练映射的学习型哈希；(c) 使用主方向的 BinaryPC。
% \caption{\textbf{Hashing methods comparison.} (a) LSH~\cite{chen2024magicpig} disrupts semantic clusters with random projections; (b) Learned hashing~\cite{li2025spotlight} preserves structure but incurs training costs; (c) BinaryPC achieves structural alignment in a training-free manner by exploiting principal data directions.}
% % 中文注释：
% % \textbf{哈希方法比较。}(a) LSH 通过随机投影破坏了语义簇；(b) 学习型哈希保留了结构但产生了训练成本；(c) BinaryPC 通过利用数据的主成分方向，以无需训练的方式实现了结构对齐。
\label{fig:motivate-hash}
\vspace{-0.2in}
\end{figure}

% 基于哈希的方案，结合movtivation图，引出我们的方法
Recently, hashing-based sparse attention, such as MagicPIG~\cite{chen2025magicpig} and Spotlight~\cite{li2025spotlight}, has emerged as a promising paradigm for fine-grained salient token selection. 
By leveraging compact binary representations, these methods enable efficient similarity computation by significantly reducing memory transfer and enabling low cost bitwise operations.
However, MagicPIG relies on data-independent projections, requiring extremely long hash codes to compensate for precision loss and may still limit its ability to accurately identify true attention affinity. 
%Although MagicPIG reports competitive performance on certain RULER subtasks, this observation does not generalize \cite{chen2024magicpig}. 
As illustrated in Figure~\ref{fig:motivate-1}, on the NIAH multi-value task, MagicPIG consistently underperforms oracle TOPK across all context lengths. At a 128K context length, MagicPIG falls 6.5 percentage points below oracle TOPK, which itself remains nearly identical to the full-attention baseline. This performance gap highlights the limitations of data-agnostic projections, which fail to capture the structural properties inherent in LLM activations.
% 图中的潜在空间可视化进一步阐明了这些现象。
As illustrated in Figure~\ref{fig:motivate-2}, data-independent random hyperplanes that are oblivious to the underlying data structure lead to suboptimal hash partitions. 
In contrast, learned non-linear hashing functions such as Spotlight~\cite{li2025spotlight} can better capture the structure of the data (Figure~\ref{fig:motivate-3}), but incur substantial training overhead, requiring calibration data and extensive optimization for each model variant.
% 这些观察促使我们探索一种有原则的、免训练的、同时又是数据感知的哈希方案。
These observations motivate the need for a training-free yet data-aware hashing scheme. 

% 为解决这些局限性，我们提出 \textbf{BinaryPC}，优点，相较于其他方法，达到的加速比
To address these limitations, we propose \textbf{BinaryPC}, a training-free data-aware hashing-based sparse attention. 
Given key vectors $\mK$, BinaryPC constructs compact binary hash codes $\mH$ and a hashing projection $\mP$ by computing binary principal components of $\mK$, which minimizes the reconstruction error $\norm{\mK - \mH\mP}_F$. In this formulation, the binary hash codes $\mH$ capture the structural information of $\mK$ through the projection $\mP$. 
Unlike LSH with data-independent random projections or learned hashing methods requiring per-model optimization, BinaryPC derives hash codes and projections in a single forward pass without gradient-based training.
As shown in Figure~\ref{fig:motivate-4}, BinaryPC leverages the principal directions of the data to achieve structural alignment. 

% 如图~\ref{fig:motivate-4} 所示，BinaryPC 利用键分布的主方向实现结构对齐，保持语义邻域关系。
% 具体而言，对于 8K 上下文的 Llama-3-8B，计算投影矩阵仅需对 180 个样本进行单次前向传播，在单张 GPU 上耗时不到五分钟。
%Concretely, for Llama-3-8B with an 8K context, computing the projection matrix requires only a single forward pass over 180 samples, taking less than five minutes on a single GPU.

Notably, BinaryPC achieves competitive retrieval fidelity with compact 64-bit codes—over 10$\times$ shorter than MagicPIG~\cite{chen2025magicpig} and 2$\times$ shorter than Spotlight~\cite{li2025spotlight}—substantially reducing memory transfer and computational overhead.
Furthermore, the reconstruction-based formulation provides interpretable error signals, enabling an error-aware safeguard that preserves recall for hard-to-hash tokens.
BinaryPC is model-agnostic and can be integrated seamlessly into pre-trained LLMs via online projection during inference or offline calibration, yielding up to 3.56$\times$ improvement in end-to-end decoding throughput over FlashAttention~\cite{dao2022flashattention,dao2024flashattention} and 5.04$\times$ speedup when FlashAttention falls back to standard implementations.

% 本文贡献总结如下：
Our contributions are summarized as follows:
\vspace{-0.1in}
\begin{itemize}

    \item We propose \textbf{BinaryPC}, a training-free data-aware hashing-based sparse attention mechanism that constructs compact binary codes by computing binary principal components of data, enabling efficient retrieval via fast bitwise operations.

    \item We develop a lightweight procedure that minimizes the reconstruction error $\norm{\mK - \mH\mP}_F$, allowing the binary hash codes $\mH$ to capture the structural information of $\mK$ through the projection $\mP$. As a result, salient token retrieval can be performed both efficiently and accurately using $\mH$. An error-aware safeguard is incorperated to preserve recall for hard-to-hash tokens.
    
    \item Extensive experiments show that BinaryPC maintains accuracy across multiple models and benchmarks while substantially improving decoding throughput.
\end{itemize}

\section{Related Works}

\subsection{Sparse Attention}
% 稀疏注意力
To mitigate the computational bottlenecks of long-context LLMs, sparse attention mechanisms selectively reduce the number of KV pairs during attention calculation. 

A primary category relies on static selection strategies determined during the prefill stage. 
StreamingLLM~\cite{xiao2024efficient} retains attention sinks with a sliding window; FastGen~\cite{ge2024model} adaptively selects policies based on prefill attention patterns; SnapKV~\cite{li2024snapkv} identifies significant clusters within an observation window to retain salient KV representations; PyramidKV~\cite{cai2025pyramidkv} implements layer-wise pyramidal budget allocation; CAKE~\cite{qin2025cake} and CompressKV~\cite{lin2025compresskv} leverage attention entropy and semantic retrieval heads for fine-grained optimization. 
However, these methods are query-agnostic: token importance is assessed only once after prefilling, leading to irreversible information loss when subsequent queries depend on unprioritized context.

% 动态选择方法
Dynamic selection methods perform token filtering at each decoding step for query-awareness. 
H2O~\cite{zhang2023h2o} tracks heavy-hitter tokens via cumulative attention scores; TOVA~\cite{oren2024transformers} approximates query-key inner products for online selection; Keyformer~\cite{adnan2024keyformer} utilizes regularized importance estimates for stability. 
Quest~\cite{tang2024quest} partitions tokens into blocks for page-level selection based on approximated maximum attention scores. 
These approaches often rely on coarse-grained or heuristic scoring that may misalign with true attention affinity. 
Moreover, block-level strategies suffer from intra-page fragmentation, requiring entire pages to be retained even when only a small subset of tokens is salient.

\subsection{Hashing-Based Sparse Attention}
% 基于哈希的稀疏注意力
By leveraging compact binary representations, hashing-based sparse attention enables efficient token-level retrieval via bitwise operations, facilitating finer-grained token selection than coarse-grained heuristic strategies. 

% MagicPIG
MagicPIG~\cite{chen2025magicpig} employs
% LSH
Locality-Sensitive Hashing (LSH) 
with data-independent random projections to select salient tokens.
% via Hamming distance. 
However, LSH's random hyperplanes are often misaligned with the intrinsic structure observed in LLM activations. 
% This misalignment necessitates extremely long hash codes (often exceeding 1000 bits) to maintain acceptable recall, and further requires auxiliary structures
% such as local windows and sink tokens 
% to stabilize retrieval, increasing system complexity and memory overhead.
This misalignment necessitates extremely long hash codes (exceeding 1000 bits) to maintain acceptable recall, and further requires auxiliary structures to stabilize retrieval, increasing system complexity and memory overhead.
% Spotlight
Learned hashing methods~\cite{desai2025hashattention,gong2025hata,li2025spotlight} incorporate trainable hashing functions to more effectively capture query-key similarity.
% To improve code efficiency, 
Spotlight~\cite{li2025spotlight} replaces the data-independent LSH with a learned MLP-based hashing function trained with ranking-oriented objectives. 
By fitting the underlying data distribution, Spotlight reduces the hash length while preserving the retrieval accuracy. 
However, it introduces considerable training overhead. It requires around 8 hours of optimization on 8,192 samples even for 8K context length. Moreover, the hashing module must be retrained for each model, restricting its practicality as a drop-in replacement.

% 与现有方法的对比总结
In contrast to MagicPIG that require long codes and Spotlight that demands extensive training, BinaryPC achieves superior performance with compact 64-bit codes.
This is accomplished through a lightweight training-free data-aware procedure that explicitly minimizes the reconstruction error $\|\mK - \mH\mP\|_F$, 
% enabling the binary hash codes $\mH$ to encode the structural information of $\mK$, 
which in turn allows salient tokens to be identified both efficiently and accurately.

\section{BinaryPC}

\subsection{Hashing-Based Retrieval}
\label{sec:hash_retrieval}

Efficiently and accurately identifying salient tokens is fundamental to sparse attention mechanisms. Hashing-based methods approximate attention scores with substantially reduced computational cost to efficiently retrieve the top-$k$ most relevant key-value (KV) pairs. 

Given query $\vq$ and key $\vk$, existing methods typically transform these vectors into compact binary codes $\vh \in \{-1, 1\}^H$ via a hashing function $\Phi(\cdot)$, yielding $\vh_q = \Phi(\vq)$ and
$\vh_k = \Phi(\vk)$, where $H < D$. For instance, MagicPIG \cite{chen2025magicpig} employs LSH with random projections, whereas Spotlight \cite{li2025spotlight} adopts a learnable MLP-based hashing scheme to adapt to data distributions.

The objective of these methods is to ensure that the inner product between $\vh_q$ and $\vh_k$ serves as a high-fidelity proxy for the full precision query-key inner product: 
\begin{equation}
\vh_q \vh_k^\top \sim \vq \vk^\top,
\label{eq:ideal_condition}
\end{equation}
When this relationship holds, relevant KV pairs can be identified via $\vh_q \vh_k^\top$, which can be calculated with significantly reduced memory transfer and lower computational cost. 

% The common paradigm underlying these methods involves two stages: (1) projecting queries and keys from $\mathbb{R}^D$ to a lower-dimensional space $\mathbb{R}^H$ (where $H \ll D$) via a projection function $\Phi(\cdot)$, and (2) quantizing the projected vectors to enable efficient similarity computation. Formally, given an input vector $\vx \in \mathbb{R}^D$, the projection can be expressed as:
% \begin{equation}
% \Phi(\vx) =
% \begin{cases}
% \vx \mR, & \text{(LSH)} \\[4pt]
% \mW_{2}\,\text{SiLU}(\mW_{1}\vx + \vb_{1}), & \text{(MLP)}
% \end{cases}
% \label{eq:hash_mapping}
% \end{equation}
% where $\mR \in \mathbb{R}^{D \times H}$ denotes a random Gaussian matrix as in MagicPIG~\cite{chen2024magicpig}, and $\{\mW_1, \mW_2, \vb_1\}$ are learnable parameters as in Spotlight~\cite{li2025spotlight}.

% The core objective is to ensure that the similarity computed in the compressed space faithfully approximates the full-precision inner product. This motivates the following ideal retrieval condition:
% \begin{equation}
% \tilde{\vq}\, \tilde{\vk}^\top \sim \vq\, \vk^\top,
% \label{eq:ideal_condition}
% \end{equation}
% where $\tilde{\vq}$ and $\tilde{\vk}$ denote the quantized representations of the projected query and key, respectively. Existing methods define $\vh_q = \sign(\Phi(\vq))$ and $\vh_k = \sign(\Phi(\vk))$, enabling efficient retrieval via Hamming distance using bitwise XOR and population count operations.

\subsection{BinaryPC} 
\label{sec:BinaryPC}

Obtaining a high-fidelity proxy for the full-precision query–key inner product in practice is challenging, as existing approaches either rely on data-independent projections that poorly align with model activations or require costly training to adapt to specific models.

To address this gap, we develop BinaryPC, a training-free and data-aware hashing-based attention mechanism. Rather than applying data-aware hashing symmetrically to both query and keys, BinaryPC adopts an asymmetric design that processes queries and keys differently. 
Since the memory transfer of keys dominates the cost due to the large number of KV pairs relative to a single query, we focus first on efficiently encoding keys into binary representations. 
Specifically, BinaryPC derives binary hash codes for keys directly from the geometry of the key vectors, constructing compact binary hash codes that preserve the structural information of the key vectors. 
We will discuss query encoding and the complete retrieval procedure later in this section.

% BinaryPC directly addresses the ideal condition in Eq.~\eqref{eq:ideal_condition} in a training-free, data-aware manner. 
% Unlike LSH, which employs random projections oblivious to the underlying data distribution, and MLP-based hashing, which requires gradient-based training, BinaryPC constructs a data-dependent projection by exploiting the intrinsic structure of key vectors at inference time.

% 问题形式化
\textbf{Problem Formulation.}
Given $N$ key vectors $\mK \in \mathbb{R}^{N \times D}$, BinaryPC computes binary hash codes $\mH \in \{-1,1\}^{N \times H}$ and a real-valued projection matrix $\mP \in \mathbb{R}^{H \times D}$ that minimizes the reconstruction error:
\begin{equation}
\min_{\mH, \mP} \norm{\mK - \mH \mP}_F.
\label{eq:BinaryPC_obj}
\end{equation}
Once optimal $\mH$ and $\mP$ are found, for any key $\vk$ in $\mK$ with corresponding hashcode $\vh$ in $\mH$, since $\vk \approx \vh \mP$, for any query $\vq$, a high-fidelity proxy can be achieved via
\begin{equation}
(\vq \mP^\top) \vh^\top = \vq (\vh \mP)^\top \approx \vq \vk^\top.
\label{eq:retrieval_expansion}
\end{equation}
Note that $\vq \mP^\top$ is real-valued. Accordingly, the second component of this asymmetric design maps $\vq \mP^\top$ into a binary representation through quantization and bit arrangement, enabling fully bitwise operations for computational efficiency. We will discuss this procedure later. 
%Later, we further quantize $\vq \mP^\top$ to enable bitwise XOR and population count operations. 

\begin{figure}[!hptb]
\centering
\includegraphics[width=\columnwidth]{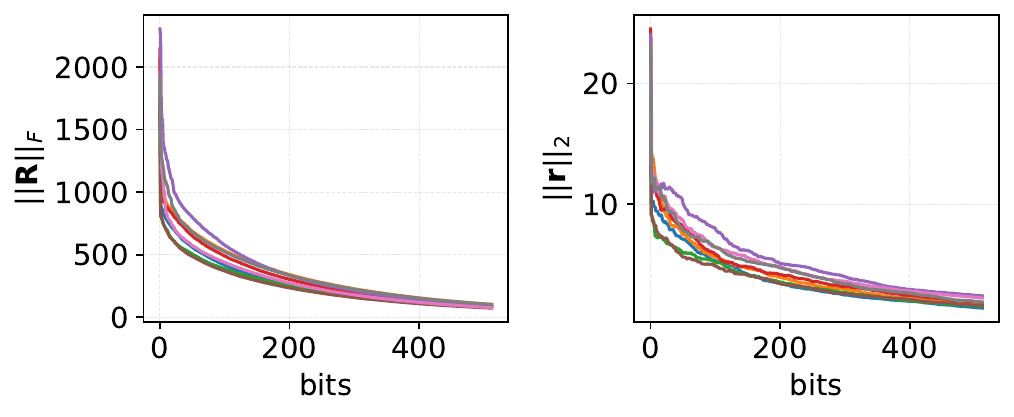}
\caption{Left: the visualization of Frobenius norm of $\mR$ vs hashcode length $H$ for Algorithm~\ref{alg:binary_proj}. Right: plot is the visualization of L2 norm of $\vr$ vs hashcode length $H$ for Algorithm~\ref{alg:compute_hash}. Different curves represent different data. The algorithms converge in all tested data. }
\label{fig:residual_norms}
% \vspace{-0.1in}
\end{figure}

% 贪心迭代算法
\textbf{Binary Principal Components Finding.}
Note that Eq. \eqref{eq:BinaryPC_obj} is closely related to Principal Component Analysis (PCA). PCA can be solved via iteratively finding the principal component (or the largest singular vector) and removing the component from the signal. Inspired by PCA, we solve Eq. \eqref{eq:BinaryPC_obj} by iteratively finding binary principal component and removing this component from $\mK$, as detailed in Algorithm~\ref{alg:binary_proj}. 
Note that given a randomly sampled vector $\vv^\ast$, let $\mR$ be the residual signal (initially $\mR = \mK$), the principal component of $\mR$ can be discovered via 
% \begin{equation}
% \vu = (\mR \mR^\top)^n \mR \vv^\top
% \end{equation}
\begin{equation}
\vu = (\mR \mR^\top)^n \mR {\vv^\ast}^\top
% $\vv^\ast$
\end{equation}
for sufficiently large $n$, since the singular values of $\mR$ will grow exponentially fast as $n$ increases making the largest singular vector dominate. 
However, since we are interested in the binary quantized component, exact principal component is less useful. We use
% \begin{equation}
% \vu = \sign(\mR \vv^\top)
% \end{equation}
\begin{equation}
\vu = \sign(\mR {\vv^\ast}^\top)
\end{equation}
as a sufficiently good binary component to save computation since the largest singular value ensure $\mR {\vv^\ast}^\top$ will lean towards the principal component with high probability.
Then, 
\begin{equation}
\vv = \vu \mR / N
\end{equation}
finds the magnitudes of projecting columns of $\mR$ onto $\vu$. Lastly, the component $\vu$ is removed from columns of $\mR$ via $\mR - \vu^\top \vv$. 
Through this iterative procedure, we can construct both hash codes $\mH$ and projection $\mP$.
Appendix~\ref{sec:binarypc_procedure_overview} provides a visual overview of this iterative rank-1 binary decomposition process.
The left plot of Figure \ref{fig:residual_norms} shows that as the iterative procedure progress, $||\mR||_F$ become progressively smaller, indicating the effectiveness of Algorithm~\ref{alg:binary_proj} in solving Eq. \eqref{eq:BinaryPC_obj}.

% \setlength{\textfloatsep}{10pt}
% \begin{algorithm}[tb]
%    \caption{Constructing hashcodes and corresponding projection}
%    \label{alg:binary_proj}
%    {\small
% \begin{algorithmic}
%    \STATE {\bfseries Input:} key vectors $\mK \in \R^{N \times D}$ and target hashcode length $H$
%    \STATE Initialize empty $\mH \in \{-1, 1\}^{N \times H}$ and $\mP \in \R^{H \times D}$
%    \STATE Initialize residual $\mR \gets \mK$
%    \FOR{$i = 1$ {\bfseries to} $i = H$}
%    \STATE sample $\vv \sim N(0, \mI_D)$
%    \STATE compute $\vu \gets \sign(\mR \vv)$ and $\vv \gets \vu \mR / N$
%    \STATE update $\mR \gets \mR - \vu \vv^{\top}$
%    \STATE $\mH[:, i] \gets \vu$ and $\mP[i, :] \gets \vv$
%    \ENDFOR
%   \STATE {\bfseries Output:} hashcodes $\mH$ and projection $\mP$
% \end{algorithmic}}
% \end{algorithm}

\setlength{\textfloatsep}{10pt}
\begin{algorithm}[tb]
   \caption{Constructing hash codes and corresponding hashing projection}
   \label{alg:binary_proj}
   {\small
\begin{algorithmic}
   \STATE {\bfseries Input:} key vectors $\mK \in \R^{N \times D}$ and target hash code length $H$
   \STATE Initialize empty $\mH \in \{-1, 1\}^{N \times H}$ and $\mP \in \R^{H \times D}$
   \STATE Initialize residual $\mR \gets \mK$
   \FOR{$i = 1$ {\bfseries to} $i = H$}
   \STATE sample $\vv^\ast \sim N(0, \mI_D)$
   \STATE compute $\vu \gets \sign(\mR {\vv^\ast}^\top)$ and $\vv \gets \vu \mR / N$
   \STATE update $\mR \gets \mR - \vu^{\top} \vv$
   \STATE $\mH[:, i] \gets \vu$ and $\mP[i, :] \gets \vv$
   \ENDFOR
  \STATE {\bfseries Output:} hash codes $\mH$ and projection $\mP$
\end{algorithmic}}
\end{algorithm}

 % {\vv^\ast}

\setlength{\textfloatsep}{10pt}
\begin{algorithm}[hptb]
   \caption{Given projection, computing hashcode}
   \label{alg:compute_hash}
   {\small
\begin{algorithmic}
   \STATE {\bfseries Input:} key vector $\vk \in \R^{D}$ and projection $\mP \in \R^{H \times D}$
   \STATE Initialize empty $\vh \in \{-1, 1\}^{H}$
   \STATE Initialize residual $\vr \gets \vk$
   \FOR{$i = 1$ {\bfseries to} $i = H$}
   \STATE let $\vv \gets \mP[i, :]$
   \STATE compute $u \gets \sign(\vr \vv^\top)$
   \STATE update $\vr \gets \vr - u \vv$
   \STATE $\vh[i] \gets u$
   \ENDFOR
  \STATE {\bfseries Output:} hashcode $\vh$
\end{algorithmic}}
\end{algorithm}

In an alternative setting where the projection matrix $\mP$ is given, we compute the hash code of a key vector $\vk$ by finding a binary representation of $\vk$ using rows of $\mP$ as the "basis". Since the rows of $\mP$ might not be orthogonal. This procedure should be done iteratively. Let $\vr$ be the residual signal (initially $\vr = \vk$), for each row $\vv$ of $\mP$, we find the binarized scalar $\sign(\vr \vv^\top)$ and subtract the corresponding component from $\vr$. The full procedure is summarized in Algorithm~\ref{alg:compute_hash}. 
The right plot of Figure \ref{fig:residual_norms} shows that as the iterative procedure progress, $||\vr||_2$ become progressively smaller, indicating the effectiveness of Algorithm~\ref{alg:compute_hash} in representing $\vk$ using rows of $\mP$ as the "basis".

\textbf{Offline Projection Calibration (OPC).}
Prior text discusses the procedure of constructing the projection $\mP$ given $\mK$ in an online setting. Alternatively, $\mP$ can be computed offline using a small calibration dataset prior to deployment to save some computational overhead.
To ensure diverse coverage across different textual domains, we construct a calibration set comprising 180 samples drawn from three representative corpora: PG19~\cite{rae2020compressive} for long-form literary text, ProofPile~\cite{Zhangir2022proofpile} for mathematical and formal reasoning content, and CodeParrot~\cite{Zhang2024long-llm-data} for programming code, with 60 samples from each source. Then, we perform a forward pass to collect key vectors from each attention layer. Algorithm~\ref{alg:binary_proj} is then applied to the aggregated keys to derive these projections. 
% This calibration is not training: it involves only a single forward pass to collect key activations, followed by applying Algorithm~\ref{alg:binary_proj}—an efficient greedy procedure, requiring no gradient computation or loss-based optimization.

% \textbf{Retrieval Phase.}
% During inference, the learned projection basis $\mP$ is applied to incoming queries. Given the reconstruction $\vk \approx \vh_k \mP$ from the quantization phase, the query-key inner product can be approximated as:
% \begin{equation}
%     \vq \vk^\top \approx \vq (\vh_k \mP)^\top = (\vq \mP^\top) \vh_k^\top.
%     \label{eq:retrieval_expansion}
% \end{equation}
% This formulation directly satisfies the ideal condition in Eq.~\eqref{eq:ideal_condition} by leveraging the data-dependent structure captured in $\mP$. 
% % During the decoding stage, we compute the quantized query projection and evaluate the weighted similarity against all cached key hashes in $\mH$, subsequently selecting the top-$k$ candidates for sparse attention computation.
% During the decoding stage, the projected query $\vq \mP^\top$ is quantized to multi-bit integers and evaluated against all cached binary key hashes in $\mH$, subsequently selecting the top-$k$ candidates for sparse attention computation.

\setlength{\textfloatsep}{10pt}
\begin{algorithm}[hptb]
   \caption{Computing hashing score}
   \label{alg:hash_score}
   {\small
\begin{algorithmic}
   \STATE {\bfseries Input:} query vector $\vq \in \R^D$, hashing projection $\mP \in \R^{H \times D}$, and hashcode $\vh \in \{-1, 1\}^{H}$
   \STATE Compute quantized $\vq \mP^\top$, where sign bits are represented as $\vs \in \{-1, 1\}^{H}$ and magnitude bits are represented as $\mM \in \{0, 1\}^{7 \times H}$
   \STATE Compute $\vs \gets \vh \text{ XOR } \vs$
   \STATE Initialize scores $c = 0$
   \FOR{$i = 1$ {\bfseries to} $i = 7$}
   \STATE $m \gets \text{count-bits}(( \vs \land \mM[i, :]) - \text{count-bits}(\neg \vs \land \mM[i, :])$
   \STATE Update $c \gets c + m \ll i$
   \ENDFOR
  \STATE {\bfseries Output:} hash score $c$
\end{algorithmic}}
\end{algorithm}

\begin{table*}[t]
  \centering
  \scriptsize
  \setlength{\tabcolsep}{2pt}
    \caption{Performance on InfiniteBench~\cite{zhang2024infty} (left) and LongBench v2~\cite{bai2025longbench} (right).}
    
  \label{tab:infinitebench}
  \resizebox{\textwidth}{!}{%
  \begin{tabular}{l|c|ccccccc|c|ccccc|c}
    \toprule
        \multirow{2}{*}{Methods} & & \multicolumn{8}{c|}{InfiniteBench} & \multicolumn{6}{c}{LongBench v2} \\
         &Token      &En.Sum  &En.QA &En.MC &En.Dia&Zh.QA &Math.F&R.PK & Avg.                         & Easy & Hard & Short& Medium& Long& Avg. \\
        \midrule
      \textbf{Llama-3.1-8B} &Full      &32.13   &25.78 &69.00 &19.50 &31.90 &25.43 &99.32 &43.30                        &31.2  &28.3  &33.3  &27.0  &27.8  &29.4   \\
              PyramidKV     &2K        &27.87   &25.83 &69.00 &19.00 &31.51 &25.43 &99.32 &42.57                        &30.7  &28.0  &33.9  &26.0  &26.9  &29.0   \\
              Cake          &2K        &28.58   &25.52 &69.00 &18.00 &30.80 &25.43 &99.32 &42.38                        &31.8  &28.6  &33.9  &27.4  &27.8  &\textbf{29.8}   \\
              CompressKV    &2K        &29.84   &25.79 &69.00 &19.50 &31.60 &25.43 &99.32 &42.92                        &31.2  &28.3  &33.3  &26.5  &28.7  &29.4   \\
              Quest         &2K        &17.63   &24.61 &69.43 &18.50 &31.29 &25.43 &99.32 &40.89                        &28.1  &26.0  &31.1  &24.7  &24.1  &26.8   \\
              MagicPIG      &Default   &32.10   &25.71 &69.00 &20.00 &30.70 &25.43 &99.32 &\underline{43.18}            &29.7  &26.4  &33.9  &24.7  &23.1  &27.6   \\
              BinaryPC     &2\%       &33.06   &24.82 &69.00 &19.00 &31.87 &25.43 &99.32 &\textbf{43.21}               &32.8  &28.0  &35.0  &26.0  &28.7  &\textbf{29.8}   \\
              BinaryPC     &2K        &32.30   &25.19 &69.00 &18.50 &31.98 &25.43 &99.32 &43.10                        &32.8  &28.0  &33.9  &27.0  &28.7  &\textbf{29.8}   \\
        BinaryPC w/ OPC    &2\%       &32.61   &25.44 &69.00 &17.50 &32.35 &25.43 &99.32 &43.09                        &31.8  &28.6  &33.9  &27.4  &28.7  &29.4   \\
        BinaryPC w/ OPC    &2K        &32.23   &25.28 &69.00 &18.00 &32.32 &25.43 &99.32 &43.08                        &31.8  &27.7  &34.4  &25.1  &28.7  &29.2   \\
      %   \midrule
      % \textbf{Qwen2.5-7B}  &Full       &34.02   &12.14 &68.12 &19.00 &9.90  &40.29 &34.58 &31.15                        &34.9  &27.7  &37.8  &26.5  &25.9  &30.4   \\
      %         PyramidKV    &2K         &27.48   &11.17 &67.69 &10.00 &9.20  &39.43 &34.58 &28.50                        &35.4  &27.3  &37.2  &26.0  &27.8  &\underline{30.4}   \\
      %         Quest        &2K         &15.57   &11.60 &67.69 &13.00 &10.56 &40.00 &34.58 &27.57                        &31.8  &25.7  &37.8  &22.3  &23.1  &28.0   \\
      %         MagicPIG     &Default    &27.31   &11.35 &69.00 &14.00 &10.45 &29.71 &10.17 &24.57                        &32.8  &25.7  &31.7  &28.4  &23.1  &28.4   \\
      %         BinaryPC    &2\%        &31.40   &11.43 &67.25 &14.00 &10.45 &39.71 &34.58 &\textbf{29.83}               &35.9  &28.3  &38.9  &27.4  &25.9  &\textbf{31.2}   \\
      %         BinaryPC    &2K        &30.61   &11.45 &67.69 &15.50 &10.21 &40.57 &34.58 &\textbf{29.83}                   &35.4  &27.7  &40.0  &27.0  &22.2  &\underline{30.6}   \\
      %         BinaryPC w/ OPC &2\%    &31.07   &11.57 &69.00 &15.50 &10.22 &33.43 &34.41 &\underline{29.31}            &35.4  &27.3  &38.9  &27.0  &23.1  &\underline{30.4}   \\
      %         BinaryPC w/ OPC &2K    &--   &11.57 &69.00 &15.50 &10.22 &33.43 &34.41 &\underline{29.31}                &35.9  &28.0  &40.0  &27.9  &22.2  &\underline{31.0}   \\
        \bottomrule
  \end{tabular}
  }% end resizebox
  % \vspace{-0.1in}
\end{table*}

% Note that $\vq \mP^\top$ is real-valued, so that second part of this asymmetric design to mapping $\vq \mP^\top$ into a binary representation via quantization and bit arrangement to enable full bitwise operations for computational efficiency, which will in discussed later. 

\textbf{Retrieval Phase.}
% 【检索阶段】
During inference, the projection $\mP$ is applied to the incoming query, $\vq \mP^\top$, and a hash score will be calculated: 
\begin{equation}
c = (\vq \mP^\top) \vh^\top
\end{equation}
As discussed, we map $\vq \mP^\top$ into a binary representation. 
Specially, the projected query $\vq \mP^\top \in \R^{H}$ is quantized into $H$ sign bits and $H$ 7-bit magnitudes and these bits are packed into a $H$-bits word for signs and 7 $H$-bits word for magnitudes as illustrated in Algorithm~\ref{alg:hash_score}. 
Then, the hash score is calculated via fast bitwise operations, such as bitwise XOR, AND, NOT, bit count, and bit shift instructions as shown in Algorithm~\ref{alg:hash_score}. 
Finally, the top-$k$ candidates with highest hash scores are subsequently selected for sparse attention computation.
% As described in Eq. \eqref{eq:alter_condition}, since $\vk \approx \vh \mP$, the query-key inner product can be approximated as:

% \begin{equation}
%     \vq \vk^\top \approx \vq (\vh \mP)^\top = (\vq \mP^\top) \vh^\top.
%     \label{eq:retrieval_expansion}
% \end{equation}
% This formulation directly satisfies the ideal condition in Eq.~\eqref{eq:ideal_condition} by leveraging the data-dependent structure captured in $\mP$. 
% This formulation aligns with the ideal condition in Eq.~\eqref{eq:ideal_condition} by leveraging the data-dependent structure captured in $\mP$.
% To enable fast bit-parallel computation, the projected query $\vq \mP^\top$ is quantized into sign bits and 7-bit magnitudes. The procedure of calculating similarity in Eq. \eqref{eq:retrieval_expansion} is described in Algorithm~\ref{alg:hash_score}. 
% The entire computation is performed via bitwise XOR and bit count instructions. Finally, the top-$k$ candidates with highest similarity scores are subsequently selected for sparse attention computation.

% The similarity against each cached binary key hash $\vh_k \in \mH$ is computed as an asymmetric integer-binary inner product:

% \begin{equation}
%     s(\hat{\vq}, \vh_k) = \langle \hat{\vq}, \vh_k \rangle = \sum_{i=1}^{m} \hat{q}_i \cdot h_{k,i},
%     \label{eq:similarity_computation}
% \end{equation}
% where $m$ is the length of the hash code. To efficiently compute this asymmetric inner product, each quantized query is decomposed into bit-planes, enabling the entire computation to be performed via bitwise XOR and population count instructions. 

\textbf{Error-Aware Safeguard (EAS).}
% A distinctive advantage of BinaryPC's reconstruction-based formulation is that it naturally provides interpretable error signals. This motivates an error-aware safeguard mechanism that is not an external heuristic, but rather an integral component intrinsically enabled by our quantization framework.

% While hashing-based retrieval significantly reduces computational overhead, approximating attention scores via binary codes inevitably introduces precision loss. Since $\vk \approx \vh_k \mP$, the inner product approximation in Eq.~\eqref{eq:retrieval_expansion} is highly accurate for tokens with small reconstruction residuals; conversely, tokens with large residuals may compromise retrieval fidelity.

A key advantage of BinaryPC's reconstruction-based design is its error signal:
\begin{equation}
||\vk - \vh \mP||_2
\label{eq:reconstruction_error}
\end{equation}
It naturally enables an error-aware safeguard mechanism integrated within the retrieval mechanism. 
Although hashing-based retrieval greatly reduces computation, approximating attention scores via binary codes incurs inevitable precision loss: for tokens with small reconstruction errors, the inner product approximation in Eq.~\eqref{eq:retrieval_expansion} remains accurate, whereas large errors may degrade retrieval fidelity.
As a result, during the prefill phase, we compute the per-token reconstruction error via Eq. \eqref{eq:reconstruction_error} and identify the tokens with the top-$m$ largest errors as a hard-to-hash set $\mathcal{S}_{\text{err}}$. During decoding, the final set of tokens selected for sparse attention computation is the union:
\begin{equation}
    \mathcal{S}_{\text{attn}} = \mathcal{S}_{\text{hash}} \cup \mathcal{S}_{\text{err}},
    \label{eq:safeguard_union}
\end{equation}
where $\mathcal{S}_{\text{hash}}$ denotes the top-$k$ candidates retrieved using the hash codes, ensuring that critical tokens are preserved and providing a robust safeguard against approximation errors.

Experimental results demonstrate that even with
% a compact bit-width of $H=64$
compact 64-bit codes
, BinaryPC approaches or matches full attention accuracy across various long-context benchmarks, validating the effectiveness of our approach.

\section{Experiments}

We empirically validate that BinaryPC substantially reduces decoding cost for large language models while preserving task accuracy. In Section~\ref{sec:accuracy}, we evaluate BinaryPC on short-, medium-, and long-context benchmarks, demonstrating that it consistently approaches or matches full-attention accuracy across diverse task categories and model families. In Section~\ref{sec:efficiency}, we show that BinaryPC with a 2\% attention budget yields significant throughput improvement.
In Section~\ref{sec:Ablation}, we conduct ablation study to analyze the effect of the EAS mechanism.

\subsection{Accuracy Evaluation}
\label{sec:accuracy}

\textbf{Setup.} We evaluate BinaryPC on four widely used large language models: Llama-3-8B, Llama-3.1-8B-Instruct~\cite{grattafiori2024llama}, Mistral-7B-Instruct-v0.3~\cite{jiang2023mistral7b}, and Qwen2.5-7B-Instruct-1M~\cite{yang2025qwen2}. Experiments are organized into four parts: (1) Short-context: three tasks from LM-Eval-Harness~\cite{eval-harness} (GSM8K-CoT~\cite{cobbe2021training}, MMLU-Flan-Cot-Fewshot~\cite{hendrycks2021measuring}, and CoQA~\cite{reddy2019coqa}); (2) Medium-context: LongBench~\cite{bai2024longbench}; (3) Long-context: InfiniteBench~\cite{zhang2024infty} and LongBench v2~\cite{bai2025longbench}; (4) Scalability from 8K to 128K: RULER~\cite{hsieh2024ruler} and Needle-in-a-Haystack (NIAH)~\cite{needle-in-haystack}. Detailed dataset descriptions are provided in Appendix~\ref{sec:dataset_details}.

\textbf{Baselines.} We compare BinaryPC and its offline calibrated variant, BinaryPC w/ OPC, against several representative sparse-attention approaches, including static selection methods (PyramidKV~\cite{cai2025pyramidkv}, CAKE~\cite{qin2025cake}, CompressKV~\cite{lin2025compresskv}), query-aware selection (Quest~\cite{tang2024quest}), and hashing-based methods (MagicPIG~\cite{chen2025magicpig}, Spotlight~\cite{li2025spotlight}). 
For MagicPIG, the effective token budget is data-dependent and difficult to estimate.
We therefore report results using its default configuration, whose effective budget is approximately 5–7\% on three LM-Eval-Harness~\cite{eval-harness} tasks, while it remains around 2–3\% on the other tasks, based on our empirical statistics. 
Because different methods adopt different token-budgeting strategies—some using a fixed percentage of the input length (e.g., 2\%), while others use a fixed number of tokens (e.g., 1K) regardless of input length—we follow the most commonly used budget-setting strategy among the baselines when configuring the size of $\mathcal{S}_{\text{attn}}$ for BinaryPC to ensure a fair comparison. In addition, 10\% of the token budget is reserved for EAS. 
% For MagicPIG, the effective token budget is data-dependent and difficult to estimate. We therefore report results using its default configuration.
% ; nevertheless, MagicPIG remains slower than BinaryPC, as discussed in Section~\ref{sec:efficiency}.
Detailed baseline methods settings are provided in Appendix~\ref{sec:Baseline_Methods}.
Among sparse attention methods, the best result is highlighted in \textbf{bold}, and the second-best is \underline{underlined}. 

% \textcolor{red}{
% To ensure that the token budget scales appropriately with the context length, we set it to 2\%; however, under a 128K context, this may exceed 2K tokens. Appendix~\ref{sec:Ablation_appendix} reports ablations under a fixed 2K-token budget and with/without EAS, while implementation details and dataset-specific budgets are provided in Appendices~\ref{sec:Baseline_Methods} and~\ref{sec:dataset_details}.
% }

\begin{table}[t]
    \centering
    \footnotesize
    \setlength{\tabcolsep}{3pt}
    \caption{Short-context evaluation on three benchmarks from LM-Eval-Harness~\cite{eval-harness}. 
    %BinaryPC and Spotlight are configured with a 5\% token budget and a minimum retention of 20 tokens. MagicPIG uses 4 sink tokens and a local window of 24 tokens. Quest uses a token budget of 64.
    }

    \label{tab:shortcontext}
    \begin{tabular}{l|c|ccc|c}
        \toprule
                    Method    &Token  & GSM8K & COQA & MMLU  & Avg. \\
        \midrule
        \textbf{Llama-3-8B}   &Full   & 54.50 & 80.53& 59.73 & 64.92 \\
        Quest                 &64    & 3.10  & 75.69& 27.87 & 35.55 \\
        MagicPIG              &Default& 40.40 & 76.77& 52.21 & 56.46 \\
        Spotlight             &5\%    & 40.00 & 80.11& 56.81 & 58.97 \\
        BinaryPC             &5\%    & 42.60 & 79.63& 55.70 & \underline{59.31} \\
        BinaryPC w/ OPC      &5\%    & 52.00 & 80.28& 59.18 & \textbf{63.82} \\
        \midrule
        \textbf{Llama-3.1-8B} &Full   & 73.80 & 78.82& 65.60 & 72.74 \\
        Quest                 &64   & 12.00 & 71.82& 25.78 & 36.53 \\
        MagicPIG              &Default& 60.90 & 75.34& 56.46 & 64.23 \\
        BinaryPC             &5\%    & 69.20 & 78.74& 64.50 & \underline{70.81} \\
        BinaryPC w/ OPC      &5\%    & 70.60 & 78.80& 64.08 & \textbf{71.16} \\
        \bottomrule
    \end{tabular}
\end{table}
\textbf{Short-context.} Table~\ref{tab:shortcontext} compares BinaryPC and its variants against full-attention baselines and sparse attention methods on three short-context benchmarks from LM-Eval-Harness~\cite{eval-harness}.
Notably, Quest~\cite{tang2024quest}, which operates under a comparable token budget (64 tokens, exceeding 5\% of the input length), exhibits a substantial performance drop.
%due to in-page fragmentation, making it challenging to maintain quality under constrained token budgets. 
In contrast, BinaryPC maintains strong task performance.
On Llama-3-8B, BinaryPC achieves an average score of 59.31, outperforming MagicPIG~\cite{chen2025magicpig} and even the training-based Spotlight~\cite{li2025spotlight}. The offline variant further narrows the gap to full attention, reaching 63.82. 
%This improvement is partly attributed to better alignment of the principal directions of key distributions when more adaptation data are available in the short-context regime. 
On Llama-3.1-8B, BinaryPC achieves 70.81 average accuracy, substantially surpassing MagicPIG while closely tracking the full-attention baseline. 
These results demonstrate that BinaryPC effectively reduces attention computation without significantly degrading task-level quality in short-context scenarios.

\begin{table}[t]
  \centering
  \scriptsize
  \setlength{\tabcolsep}{1.4pt}
  % \caption{LongBench~\cite{bai2024longbench} evaluation on Llama-3-8B. 
  % \caption{LongBench~\cite{bai2024longbench} evaluation results.}
  \caption{LongBench~\cite{bai2024longbench} evaluation results. Scores are averaged by task category.}
    \label{tab:longbench_llama38b}
    \resizebox{\columnwidth}{!}{%
    \begin{tabular}{l|c|cccccc|c}
    \toprule
    Methods                &Token      & S-Doc & M-Doc & Sum.  &F-shot & Syn.  & Code  & Avg. \\
    \midrule
    \textbf{Llama-3-8B}    &Full       & 18.25 & 9.62  & 18.05 & 68.99 & 4.99  & 67.66 & 30.63 \\
    MagicPIG               &Default    & 16.96 & 9.18  & 16.67 & 68.20 & 5.17  & 66.55 & 29.78 \\
    Spotlight              &2\%        & 18.26 & 9.19  & 18.13 & 69.27 & 5.06  & 65.19 & \underline{30.31} \\
    BinaryPC              &2\%        & 17.99 & 9.12  & 18.36 & 68.60 & 4.75  & 65.58 & 30.18 \\
    BinaryPC w/ OPC       &2\%        & 17.61 & 9.70  & 17.89 & 69.02 & 4.65  & 66.87 & \textbf{30.35} \\
    \midrule
    \textbf{Llama-3.1-8B}  &Full       & 43.40 & 46.46  & 28.95 & 69.25 & 55.50  & 59.57 & 49.64 \\
    PyramidKV              &1K         & 42.53 & 45.69  & 25.32 & 68.13 & 55.38  & 57.35 & 48.15 \\
    Cake                   &1K         & 42.83 & 45.74  & 25.87 & 68.45 & 55.46  & 58.32 & 48.51 \\
    CompressKV             &1K         & 43.45 & 46.00  & 26.09 & 68.63 & 55.38  & 58.97 & 48.82 \\
    Quest                  &1K         & 42.45 & 46.36  & 28.84 & 68.34 & 55.30  & 57.14 & 48.93 \\
    MagicPIG               &Default    & 43.06 & 46.13  & 28.11 & 68.76 & 54.84  & 58.11 & 49.01 \\
    BinaryPC              &2\%        & 43.59 & 46.33  & 29.00 & 69.16 & 55.25  & 58.61 & 49.50 \\
    BinaryPC              &1K         & 43.38 & 46.22  & 29.02 & 69.27 & 55.48  & 59.98 & \underline{49.66} \\
    BinaryPC w/ OPC       &2\%        & 43.58 & 46.24  & 28.74 & 68.93 & 55.38  & 59.42 & 49.50 \\
    BinaryPC w/ OPC       &1K         & 43.45 & 46.34  & 28.73 & 69.33 & 55.43  & 60.13 & \textbf{49.67} \\
    \midrule
    \textbf{Mistral-7B}    &Full       & 38.63 & 39.66  & 28.63 & 70.74 & 52.00  & 60.24 & 47.31 \\
    PyramidKV              &1K         & 36.96 & 37.62  & 25.25 & 70.11 & 50.00  & 57.88 & 45.35 \\
    Cake                   &1K         & 37.69 & 38.15  & 26.39 & 70.10 & 51.00  & 59.38 & 46.11 \\
    CompressKV             &1K         & 38.50 & 38.75  & 26.52 & 70.68 & 51.25  & 59.44 & 46.55 \\
    Quest                  &1K         & 37.52 & 37.55  & 27.70 & 69.96 & 49.54  & 58.36 & 45.87 \\
    MagicPIG               &Default    & 38.53 & 38.91  & 28.47 & 70.76 & 51.25  & 60.24 & 47.06 \\
    BinaryPC              &2\%        & 38.32 & 38.66  & 28.70 & 70.49 & 52.00  & 60.32 & 47.07 \\
    BinaryPC              &1K         & 38.69 & 39.44  & 28.74 & 70.90 & 52.00  & 60.39 & \underline{47.38} \\
    BinaryPC w/ OPC       &2\%        & 39.23 & 38.88  & 28.60 & 71.04 & 51.50  & 59.78 & 47.24 \\
    BinaryPC w/ OPC       &1K         & 38.53 & 39.30  & 28.80 & 70.98 & 52.50  & 60.48 & \textbf{47.42} \\

    \bottomrule
  \end{tabular}
}
% \vspace{-0.1in}
\end{table}

\begin{figure*}[htbp]
\centering
\begin{subfigure}{0.43\linewidth}
\centering
\includegraphics[width=\linewidth]{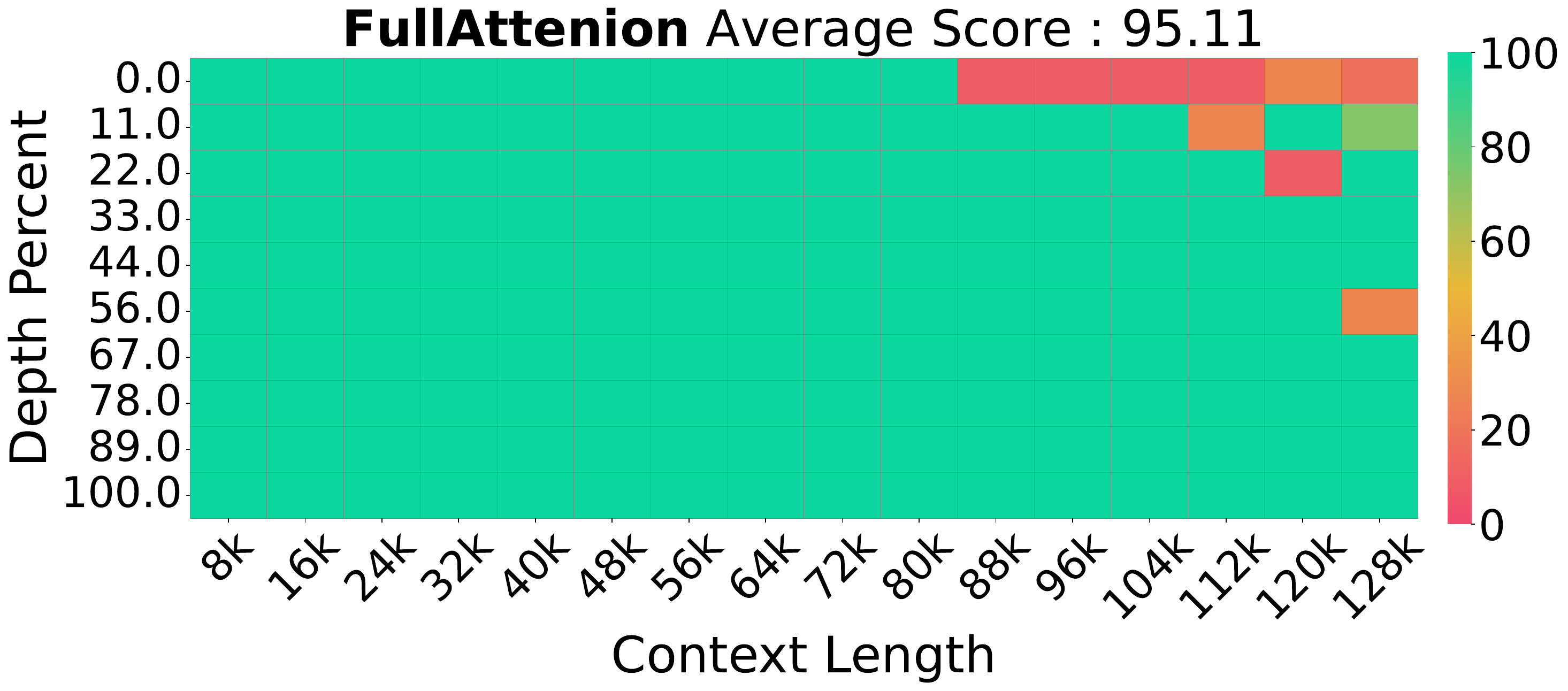}
\end{subfigure}
\begin{subfigure}{0.43\linewidth}
\centering
\includegraphics[width=\linewidth]{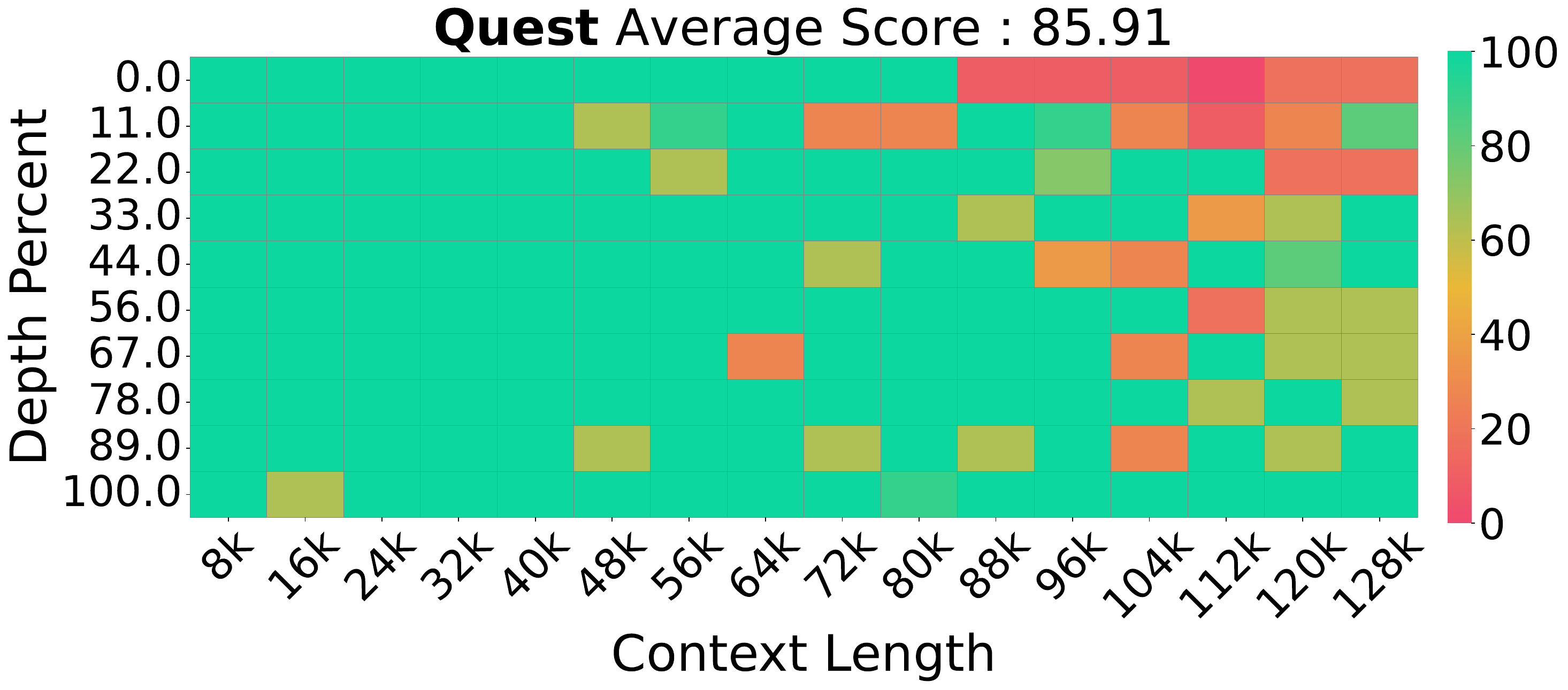}
\end{subfigure}
\begin{subfigure}{0.43\linewidth}
\centering
\includegraphics[width=\linewidth]{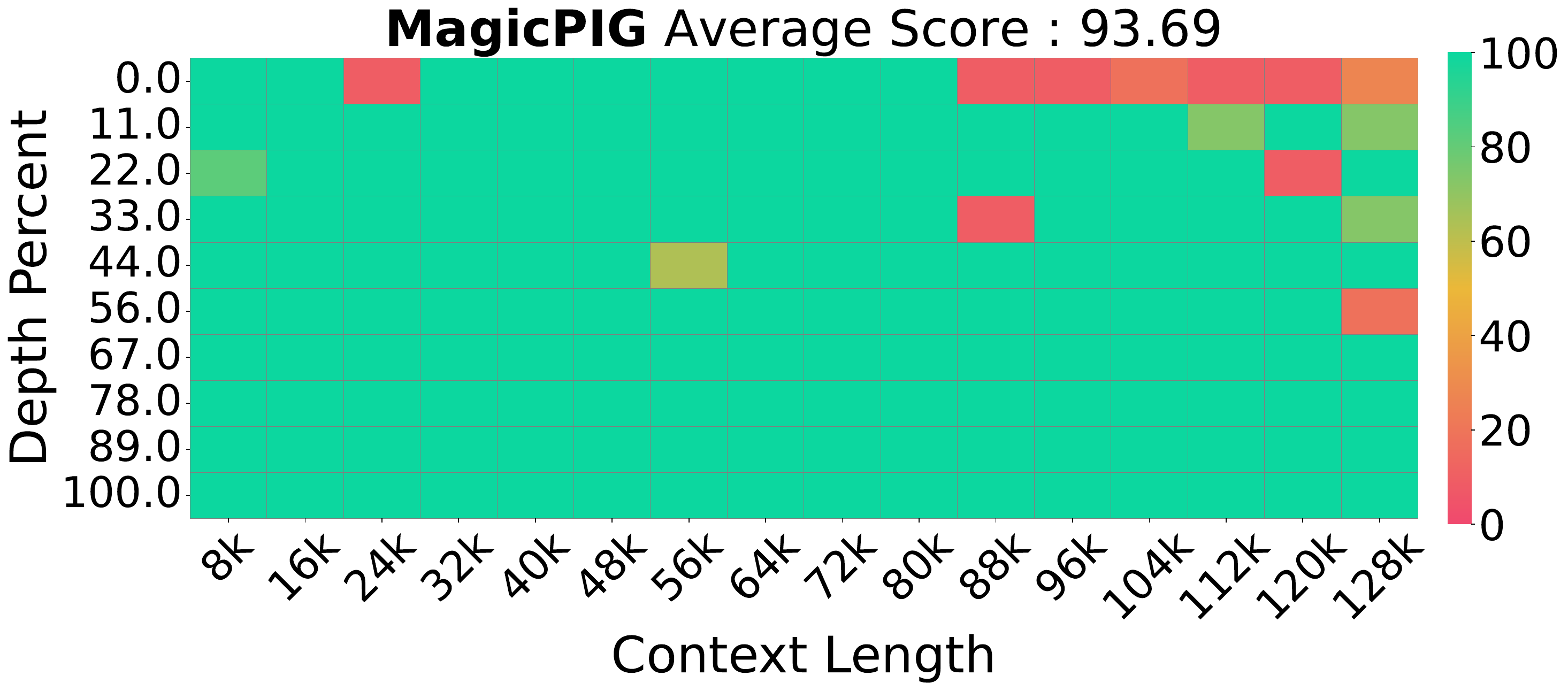}
\end{subfigure}
\begin{subfigure}{0.43\linewidth}
\centering
\includegraphics[width=\linewidth]{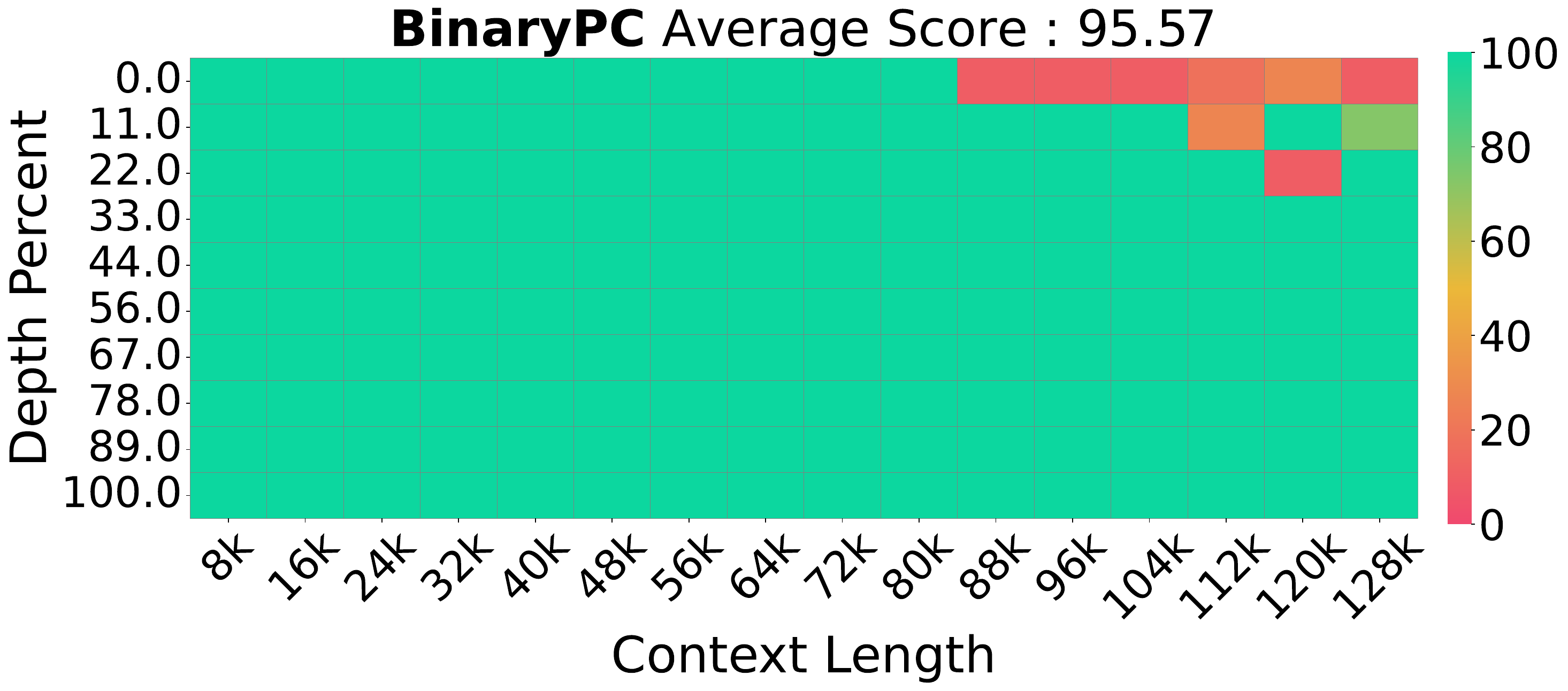}
\end{subfigure}
\caption{Evaluation on the NIAH benchmark~\cite{needle-in-haystack} using Llama-3.1-8B-Instruct.}
% across context lengths from 8K to 128K tokens.}
% 中文注释：在 Llama-3.1-8B-Instruct 上，跨 8K 到 128K 上下文长度对 NIAH 基准进行评估。 across context lengths from 8K to 128K tokens.
\label{fig:niah}
% \vspace{-0.15in}
\vspace{-0.1in}
\end{figure*}

% Medium-context evaluation on LongBench
\textbf{Medium-context.} We evaluate on LongBench~\cite{bai2024longbench}, a bilingual benchmark comprising 16 datasets across six task categories.
As shown in Table~\ref{tab:longbench_llama38b}, on Llama-3-8B, BinaryPC outperforms MagicPIG and matches accuracy with Spotlight in a training-free manner, while Spotlight requires substantial per-model training to learn complex hashing mappings.
On Llama-3.1-8B and Mistral-7B, BinaryPC with a 2\% budget matches full-attention performance and outperforms all sparse-attention baselines. Notably, with a 1K budget, BinaryPC even surpasses full attention on both models.
These findings demonstrate BinaryPC's strong generalization ability without model-specific tuning or auxiliary architectural modifications.
Table~\ref{tab:longbench} in the Appendix reports per-dataset performance across all 16 datasets for three model families, including additional results on Qwen2.5-7B.
% Appendix~\ref{sec:ultra_large_scale_appendix} further evaluates BinaryPC on Llama-3-70B-Instruct, including accuracy, memory overhead, and attention-layer latency.
Appendix~\ref{sec:ultra_large_scale_appendix} further shows that BinaryPC matches full attention on Llama-3-70B-Instruct while preserving the same relative hash-code memory overhead and similar single-layer attention latency.
\begin{table}[t]
  \centering
  \scriptsize
  \setlength{\tabcolsep}{2pt}
    \caption{RULER~\cite{hsieh2024ruler} scalability evaluation on Llama-3.1-8B-Instruct from 8K to 128K context length.}
    % \caption{RULER~\cite{hsieh2024ruler} benchmark results.
    % from 8K to 128K context length. 
    %PyramidKV, Cake, CompressKV, and Quest are evaluated under a 2K token budget.
    % }
    % 中文注释：RULER 基准任务在 8K 到 128K 的上下文长度范围上进行评测。PyramidKV、Cake、CompressKV 和 Quest 在 2K 的 token 预算下进行评估。
    \label{tab:ruler}
    \resizebox{\columnwidth}{!}{%
    \begin{tabular}{l|c|ccccc|c}
    \toprule
    Methods                &Token  & 8K   & 16K  & 32K  & 64K  & 128K & Avg. \\
    \midrule
    \textbf{Llama-3.1-8B}  &Full     &94.32 &93.96 &87.10 &85.19 &76.49 &87.41 \\
    TOPK                   &2\%      &91.53 &91.37 &88.80 &83.39 &73.62 &85.74 \\
    PyramidKV              &2K       &82.64 &79.39 &72.69 &68.34 &48.52 &70.32 \\
    Cake                   &2K       &91.48 &85.37 &78.14 &72.28 &61.61 &77.78 \\
    CompressKV             &2K       &89.52 &83.83 &78.83 &74.60 &65.32 &78.42 \\
    Quest                  &2K       &92.36 &90.52 &83.10 &78.45 &65.39 &81.96 \\
    MagicPIG               &Default  &91.28 &90.82 &85.26 &83.96 &72.07 &84.68 \\
    BinaryPC              &2\%      &91.67 &90.86 &87.73 &84.36 &73.23 &85.57 \\
    BinaryPC              &2K       &94.47 &93.98 &88.86 &85.09 &72.89 &\textbf{87.06} \\
    BinaryPC w/ OPC       &2\%      &90.56 &91.15 &87.36 &84.14 &73.27 &85.29 \\
    BinaryPC w/ OPC       &2K       &94.17 &93.88 &89.55 &84.77 &72.56 &\underline{86.99} \\
    % \midrule
    % \textbf{Qwen2.5-7B}    &93.00 &93.44 &91.80 &87.70 &83.29 &89.85 \\
    % PyramidKV              &78.45 &77.09 &74.06 &68.97 &61.86 &70.08 \\
    % Quest                  &86.97 &84.63 &80.60 &73.21 &69.12 &78.91 \\
    % MagicPIG               &37.57 &42.23 &46.13 &29.81 &28.54 &36.86 \\
    % BinaryPC              &90.68 &90.66 &88.10 &84.50 &81.66 &\textbf{87.12} \\
    % BinaryPC w/ OPC         &90.86 &90.08 &88.73 &84.78 &80.70 &\underline{87.03} \\
    \bottomrule
  \end{tabular}
    }
\end{table}

\textbf{Long-context.} We evaluate BinaryPC on InfiniteBench and LongBench v2. InfiniteBench~\cite{zhang2024infty} has an average sequence length exceeding 100K tokens, and we assess seven task types from this suite. LongBench v2~\cite{bai2025longbench} stratifies results by both difficulty and context length.
As shown in Table~\ref{tab:infinitebench}, BinaryPC delivers performance that is nearly equivalent to full attention across both benchmarks.
% Additional Qwen2.5-7B results are provided in the Appendix (Table~\ref{tab:infinitebench_qwen}).
Additional Qwen2.5-7B results in Appendix Table~\ref{tab:infinitebench_qwen} further show consistent gains over sparse baselines, demonstrating that the long-context robustness generalizes across model families.

\textbf{Scalability from 8K to 128K.} We assess the scalability and stability of BinaryPC across context lengths from 8K to 128K tokens using the RULER~\cite{hsieh2024ruler} benchmark and NIAH~\cite{needle-in-haystack}. On RULER (shown in Table~\ref{tab:ruler}), BinaryPC consistently outperforms other sparse methods and maintains stable performance across scaling regimes. Detailed per-task RULER results across different context lengths are provided in Appendix~\ref{sec:ruler_details}. 
As shown in Figure~\ref{fig:motivate-1}, BinaryPC achieves performance nearly identical to the oracle TOPK and closely aligns with full attention. In contrast, MagicPIG consistently underperforms the oracle TOPK baseline. 
% Notably, on Qwen2.5-7B, MagicPIG exhibits marked performance degradation, whereas BinaryPC maintains high accuracy without model-specific tuning, highlighting its superior stability.
For NIAH evaluation (Figure~\ref{fig:niah}), BinaryPC matches full-attention retrieval accuracy across all needle positions and context lengths, and even surpasses full attention in certain cases. More detailed NIAH results are included in the Appendix (Figures~\ref{fig:appendix_llama_images} and~\ref{fig:appendix_Qwen_images}) for a comprehensive comparison of more baselines and BinaryPC variants across models and context lengths.
Beyond retrieval accuracy, Appendix~\ref{sec:approximation_quality_appendix} further shows that BinaryPC provides a better approximation to full attention outputs than MagicPIG under comparable token budgets, achieving consistently higher cosine similarity across context lengths from 8K to 64K.
In summary, BinaryPC achieves superior performance compared to all baselines. 

\subsection{Efficiency Evaluation}
\label{sec:efficiency}

% \begin{figure}[t]
% \centering
% \begin{minipage}{0.48\linewidth}
%     \centering 
%     \includegraphics[width=\textwidth]{figures/efficiency/efficiency-ct-batch1.pdf}
% \end{minipage}
% \hfill
% \begin{minipage}{0.48\linewidth}
%     \centering
%     \includegraphics[width=\textwidth]{figures/efficiency/efficiency-ct-batch2.pdf}
% \end{minipage}
% \caption{End-to-end throughput comparison across context lengths for batch sizes of 1 (left) and 2 (right). }
% % 中文注释：在不同上下文长度下，批量大小为 1（左）和 2（右）的端到端吞吐量比较。
% \label{fig:efficiency_batch}
% \end{figure}

\begin{figure*}[t]
\centering
\begin{minipage}{0.24\textwidth}
    \centering 
    \includegraphics[width=\textwidth]{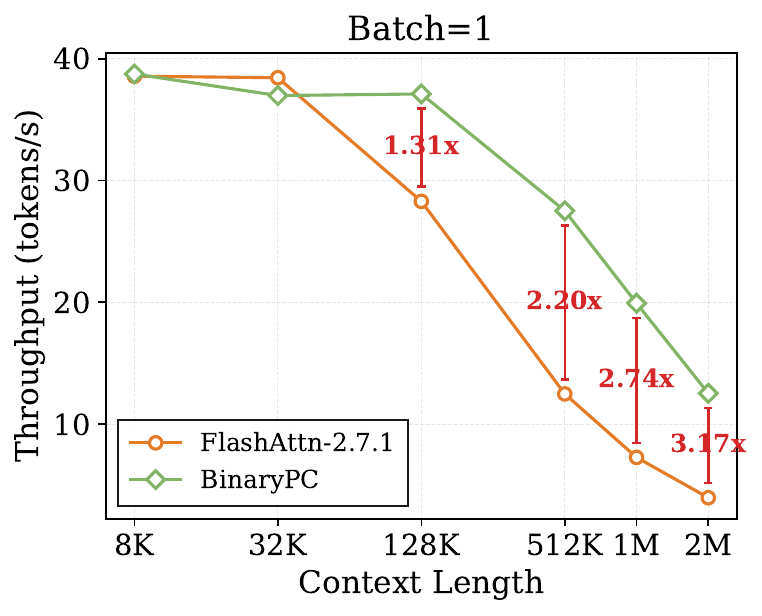}
\end{minipage}
\hfill
\begin{minipage}{0.24\textwidth}
    \centering
    \includegraphics[width=\textwidth]{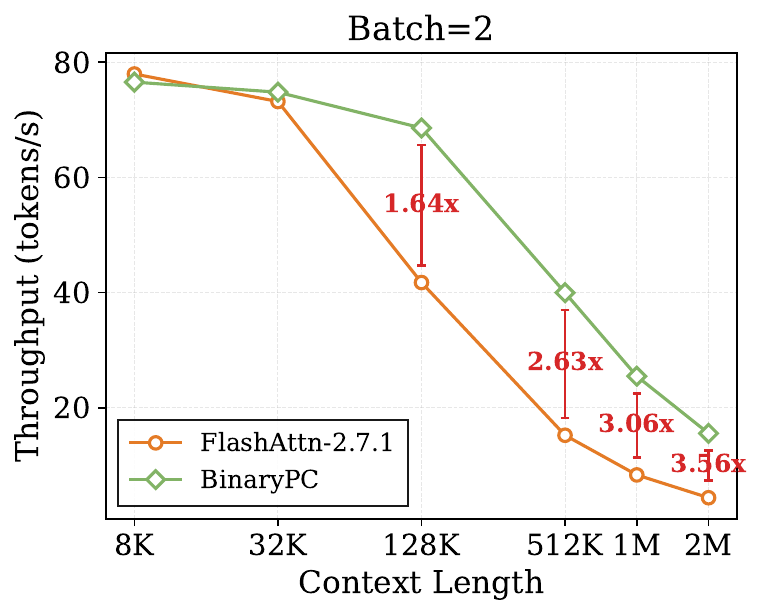}
\end{minipage}
\hfill
\begin{minipage}{0.24\textwidth}
    \centering 
    \includegraphics[width=\textwidth]{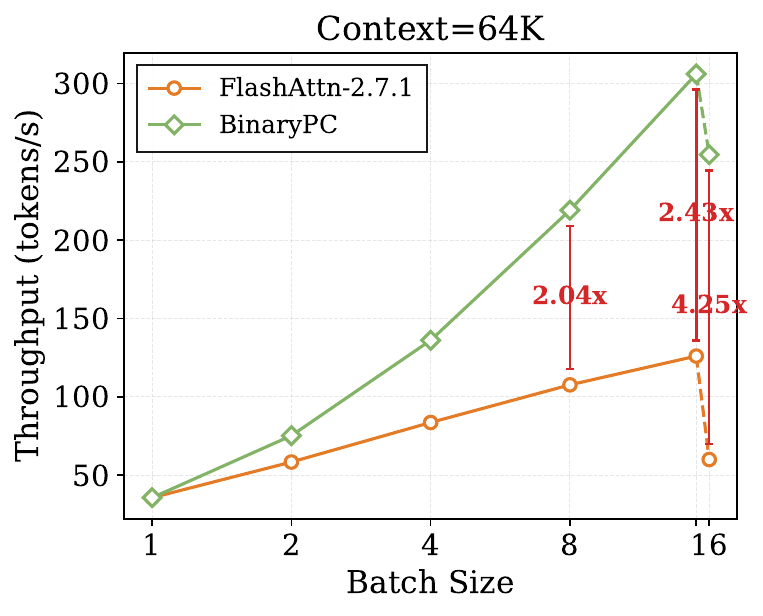}
\end{minipage}
\hfill
\begin{minipage}{0.24\textwidth}
    \centering
    \includegraphics[width=\textwidth]{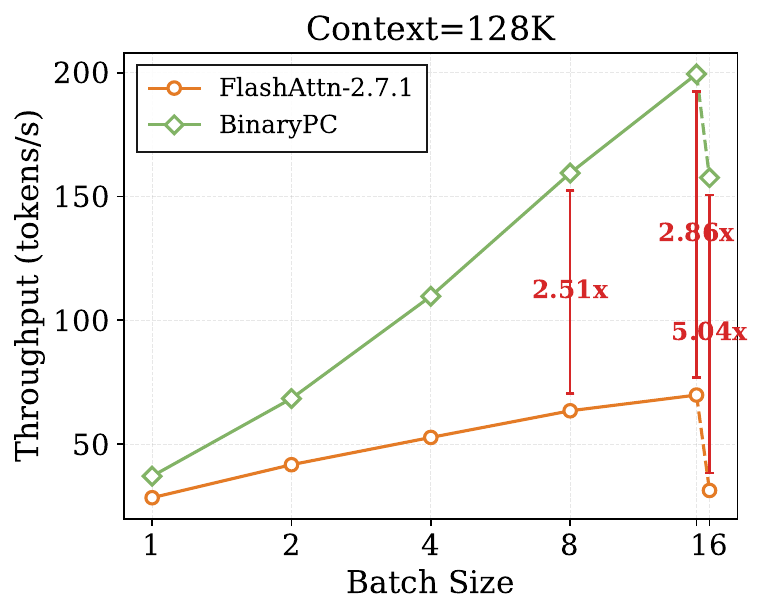}
\end{minipage}
\caption{End-to-end throughput comparison. \textbf{Left two:} throughput across context lengths (8K--2M) for batch sizes of 1 and 2. \textbf{Right two:} throughput across batch sizes (1--16) at fixed context lengths of 64K and 128K tokens.}
\label{fig:efficiency_batch}
% \vspace{-0.15in}
\end{figure*}

\textbf{Setup.} 
% All efficiency experiments were conducted on Llama-3.1-8B-Instruct~\cite{grattafiori2024llama} using eight 96 GB GPUs, evaluating the offline calibrated variant of BinaryPC.
% We employed the HuggingFace Transformers framework with pipeline parallelization and pre-allocated static KV cache to maximize throughput. The baseline method utilizes FlashAttention-2~\cite{dao2022flashattention,dao2023flashattention} with the native \textit{flash\_attention\_2} implementation. To evaluate model throughput at extended context lengths, we expanded positional encoding while disregarding output quality. 
% Alternative methods such as MagicPIG use hashing codes exceeding 1000 bits and rely on GPU–CPU collaborative decoding, resulting in notably lower throughput than FlashAttention-2. Therefore, FlashAttention-2 remains the primary baseline for efficiency evaluation.
% All efficiency experiments were conducted on Llama-3.1-8B-Instruct~\cite{grattafiori2024llama} using eight 96 GB GPUs, evaluating the offline-calibrated variant of BinaryPC. We employed the HuggingFace Transformers framework with pipeline parallelization and a pre-allocated static KV cache to maximize throughput. To assess model throughput at extended context lengths, we expanded positional encoding while disregarding output quality.
All efficiency experiments were conducted on Llama-3.1-8B-Instruct~\cite{grattafiori2024llama} using eight data-center GPUs, evaluating the offline-calibrated variant of BinaryPC. We employed the HuggingFace Transformers framework with pipeline parallelization and a pre-allocated static KV cache to maximize throughput. To assess model throughput at extended context lengths, we expanded positional encoding while disregarding output quality. 
% Alternative methods such as MagicPIG~\cite{chen2025magicpig} use hashing codes exceeding 1000 bits and rely on GPU–CPU collaborative decoding, which substantially reduces throughput. 
% As shown in Appendix Table~\ref{tab:magic_efficiency}, its throughput is consistently about half that of FlashAttention-2 across batch sizes and context lengths. We therefore adopt FlashAttention-2~\cite{dao2022flashattention,dao2024flashattention} as our baseline.
Among hashing-based baselines, MagicPIG~\cite{chen2025magicpig} uses hash codes exceeding 1000 bits and relies on GPU--CPU collaborative decoding, which substantially reduces throughput.
As shown in Appendix Table~\ref{tab:magic_efficiency}, MagicPIG reaches only 0.42--0.50$\times$ the throughput of FlashAttention-2 across batch sizes and context lengths, whereas BinaryPC achieves 1.05--1.69$\times$ speedup under the same single-GPU setting. We therefore adopt FlashAttention-2~\cite{dao2022flashattention,dao2024flashattention} as the primary efficiency baseline.
% GPU execution time was measured using CUDA events. 
We generated 64 consecutive tokens after a 32-step warm-up phase and averaged three runs per data point. 
% To validate the generality of our approach across different hardware, we performed efficiency experiments on eight NVIDIA RTX 3090 GPUs and analyzed the runtime of BinaryPC and its offline variant during the prefill stage, as reported in Appendix~\ref{sec:Efficiency_appendix}.
% We additionally conducted efficiency experiments on eight NVIDIA RTX 3090 GPUs and analyzed the runtime of BinaryPC and its offline variant during the prefill stage, as reported in Appendix~\ref{sec:Efficiency_appendix}.
To validate the generality of our approach across different hardware, we performed efficiency experiments on eight consumer-grade GPUs and analyzed the runtime of BinaryPC and its offline variant during the prefill stage, as reported in Appendix~\ref{sec:Efficiency_appendix}.
% We additionally conducted efficiency experiments on eight consumer-grade GPUs and analyzed the runtime of BinaryPC and its offline variant during the prefill stage, as reported in Appendix~\ref{sec:Efficiency_appendix}.

\begin{figure}[t]
\centering
\includegraphics[width=0.8\columnwidth]{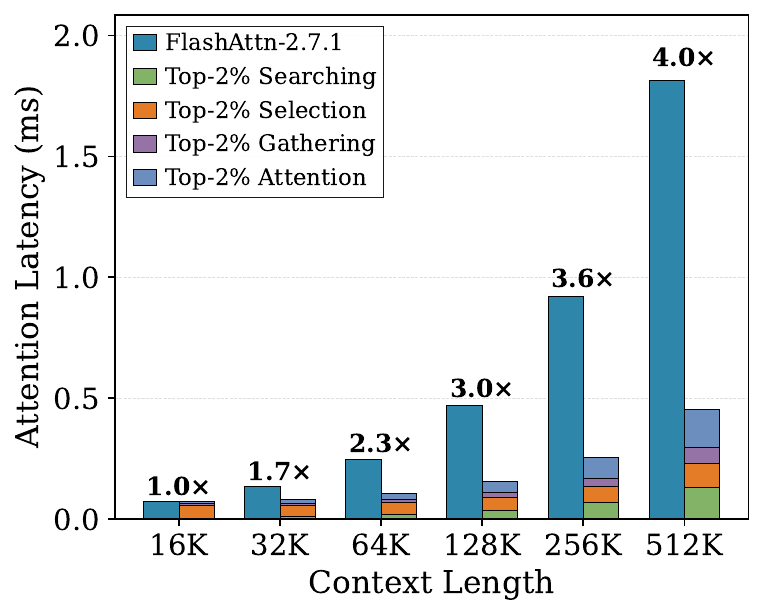}
\vspace{-0.05in}
\caption{Attention layer execution time breakdown on Llama-3.1-8B-Instruct under the default BinVortex configuration.}
% 中文注释：在 Llama-3.1-8B-Instruct 上采用 BinVortex 默认配置，报告了单请求解码下注意力层的执行时间分解。
\label{fig:efficiency_kernel}
% \vspace{-0.05in}
\end{figure}

\textbf{Kernel Evaluation.}
To fully exploit GPU hardware parallelism, we implement optimized CUDA kernels for binary hashing and hash score computation. Key vectors are transformed into 64-bit binary hash codes via constructed projection matrices.
For hash score calculation, each 8-bit quantized query element is decomposed into 8 bit-planes, which are packed as 64-bit words. The hash score calculation in Eq.~\eqref{eq:retrieval_expansion} is then computed by performing bitwise and bit count operations between each query bit-plane and the binary key hash via Algorithm~\ref{alg:hash_score}.
Figure~\ref{fig:efficiency_kernel} provides a detailed breakdown of the attention layer execution time across varying context lengths. The retrieval pipeline consists of three stages: (1)~\emph{Searching}, where a custom CUDA kernel computes the hash score between projected queries and binary key hash codes using bitwise and bit count operations; (2)~\emph{Selection}, which identifies the top-$k$ tokens via \textit{torch.topk}; and (3)~\emph{Gathering}, which retrieves the corresponding key-value pairs using \textit{torch.gather}.
% Detailed bit-packing and scoring-kernel optimizations are provided in Appendix~\ref{sec:cuda_kernel_details}.
Appendix~\ref{sec:cuda_kernel_details} details how BinaryPC packs each 64-bit key hash into one \texttt{int64} and computes hash scores using XOR and population-count operations.
Taking the 512K length setting as an example, these three stages introduce a combined overhead of 295\,$\mu$s, but BinaryPC can reduce the FlashAttention kernel invocation time from 1.814\,ms to 163\,$\mu$s, corresponding to a 4.0$\times$ speedup for a single attention layer. The overhead introduced by the retrieval stages is small relative to the total decoding latency, validating the efficiency of the proposed mechanism in long-sequence scenarios.

\textbf{End-to-end Throughput.}
Figure~\ref{fig:efficiency_batch} reports the end-to-end decoding throughput across context lengths from 8K to 2M tokens. Under small-batch settings, BinaryPC consistently outperforms the FlashAttention-2 baseline. For short contexts, BinaryPC attains comparable throughput to FlashAttention-2, while for long contexts it achieves up to 3.17$\times$ and 3.56$\times$ improvement, with the performance gap further widening as batch size and context length increase.

As the batch size increases at fixed context lengths of 64K and 128K, BinaryPC maintains an increasingly larger efficiency margin. Notably, on the data-center GPU setup, FlashAttention-2 exhibits a sharp throughput drop once the batch size exceeds 15, due to an internal heuristic that switches from the highly parallel Split-K kernel to the standard kernel.
% As the batch size increases at fixed context lengths of 64K and 128K, BinaryPC maintains an increasingly larger efficiency margin. Notably, on GPUs with 96 GB of VRAM, FlashAttention-2 exhibits a sharp throughput drop once the batch size exceeds 15, due to an internal heuristic that switches from the highly parallel Split-K kernel to the standard kernel.
While BinaryPC also invokes FlashAttention-2 kernels internally, its smaller FlashAttention input size substantially mitigates this throughput degradation. At a batch size of 16, BinaryPC achieves 4.25$\times$ and 5.04$\times$ improvements at context lengths of 64K and 128K, respectively.

% \subsection{Effect of EAS}
% \label{sec:Ablation}

% We conduct ablation study on InfiniteBench~\cite{zhang2024infty} using Llama-3.1-8B to analyze the effect of the EAS mechanism.
% As shown in Table~\ref{tab:ablation_infinite}, without EAS, the online variant BinaryPC exhibits a significant performance drop on the Passkey Retrieval (R.PK) task (from 99.32 to 55.76). 
% This degradation arises because the passkey is highly dissimilar from the surrounding context, causing it to behave as an outlier during online projection construction and making it difficult to be captured with binary codes. 
% %by  hinder effective online projection construction.
% In contrast, the offline variant exhibits greater resilience without EAS, benefiting from the diverse calibration data in offline projection construction.
% Enabling EAS provides a safeguard, allowing BinaryPC to achieve near-full-attention performance.

% \input{tables/ablation_infinite}

\subsection{Ablation Study}
\label{sec:Ablation}

\textbf{Effect of EAS.}
We first conduct an EAS budget ablation on InfiniteBench~\cite{zhang2024infty} using Llama-3.1-8B-Instruct with 100 samples per subtask, varying the EAS ratio over $\{0\%,2\%,5\%,10\%,15\%\}$.
As shown in Table~\ref{tab:ablation_eas_budget}, EAS is essential for standard BinaryPC: without EAS, R.PK drops to 64.00 and the average score falls to 41.30, while a minimal 2\% budget fully recovers R.PK to 99.00.
This degradation arises because the passkey is highly dissimilar from the surrounding context, causing it to behave as an outlier during online projection construction and making it difficult to be captured with binary codes.
Once EAS is activated, performance is insensitive to its budget ratio, with BinaryPC averages varying only from 47.42 to 47.63 across 2--15\%.
This indicates that EAS is a robust algorithmic component rather than a fragile hyperparameter.
OPC provides complementary robustness: with offline projection calibration, R.PK already reaches 99.00 even at 0\% EAS, while EAS still yields modest gains on other tasks such as En.Dia (15.00 to 18.00).
We therefore use 10\% as the default EAS budget, as it achieves near-optimal performance, introduces negligible overhead due to static query-agnostic selection, and provides a comfortable margin above the 2\% effectiveness threshold.

\begin{table}[!hptb]
  \centering
  \scriptsize
  \setlength{\tabcolsep}{2pt}
  \caption{Ablation study on the EAS budget ratio using InfiniteBench with Llama-3.1-8B-Instruct, evaluated with 100 samples per subtask.}
  \label{tab:ablation_eas_budget}
  \resizebox{\columnwidth}{!}{%
  \begin{tabular}{l|c|ccccccc|c}
    \toprule
    Methods                  & EAS   & En.Sum & En.QA & En.MC & En.Dia & Zh.QA & Math.F & R.PK & Avg. \\
    \midrule
    Full Attention           & --    & 32.85 & 27.41 & 72.00 & 20.00 & 36.17 & 48.00 & 99.00 & 47.92 \\
    \midrule
    \multirow{5}{*}{BinaryPC w/ OPC}
                             & 0\%   & 32.16 & 26.43 & 72.00 & 15.00 & 35.92 & 48.00 & 99.00 & 46.93 \\
                             & 2\%   & 31.85 & 26.32 & 72.00 & 18.00 & 36.18 & 48.00 & 99.00 & 47.34 \\
                             & 5\%   & 32.78 & 26.40 & 72.00 & 18.00 & 36.83 & 48.00 & 99.00 & \underline{47.57} \\
                             & 10\%  & 32.44 & 26.43 & 72.00 & 18.00 & 36.67 & 48.00 & 99.00 & 47.51 \\
                             & 15\%  & 32.39 & 26.35 & 72.00 & 18.00 & 36.16 & 48.00 & 99.00 & 47.41 \\
    \midrule
    \multirow{5}{*}{BinaryPC}
                             & 0\%   & 30.97 & 24.33 & 72.00 & 17.00 & 32.83 & 48.00 & 64.00 & 41.30 \\
                             & 2\%   & 32.50 & 26.18 & 72.00 & 18.00 & 36.28 & 48.00 & 99.00 & 47.42 \\
                             & 5\%   & 32.48 & 25.53 & 72.00 & 20.00 & 36.38 & 48.00 & 99.00 & \textbf{47.63} \\
                             & 10\%  & 32.46 & 26.17 & 72.00 & 18.00 & 36.24 & 48.00 & 99.00 & 47.41 \\
                             & 15\%  & 32.15 & 26.03 & 72.00 & 19.00 & 36.39 & 48.00 & 99.00 & 47.51 \\
    \bottomrule
  \end{tabular}
  }
% \vspace{-0.1in}
\end{table}

\textbf{Hash bit length.}
We further study the effect of hash bit length on NIAH~\cite{needle-in-haystack} using Llama-3.1-8B-Instruct.
As shown in Table~\ref{tab:ablation_bit_length}, 32-bit hashing exhibits a notable drop at 104K context length, reducing the average score to 86.37.
In contrast, 64-bit and 128-bit hashing perform similarly well, achieving average scores of 90.18 and 90.00, respectively.
From an implementation perspective, 64-bit hash codes fit into a single \texttt{int64} word and naturally support efficient bitwise operations, offering the best trade-off between accuracy and efficiency.

\begin{table}[!hptb]
  \centering
  \scriptsize
  \setlength{\tabcolsep}{4pt}
  \caption{Ablation study on hash bit length using NIAH with Llama-3.1-8B-Instruct.}
  \label{tab:ablation_bit_length}
  \begin{tabular}{l|ccccc|c}
    \toprule
    Hash Bits & 88K & 96K & 104K & 112K & 120K & Avg. \\
    \midrule
    32-bit  & 91.82 & 91.82 & 71.82 & 91.82 & 84.55 & 86.37 \\
    64-bit  & 90.91 & 91.82 & 92.73 & 90.91 & 84.55 & \textbf{90.18} \\
    128-bit & 91.82 & 91.82 & 90.91 & 91.82 & 83.64 & \underline{90.00} \\
    \bottomrule
  \end{tabular}
\end{table}

\textbf{OPC robustness under domain shift.}
We next examine whether OPC depends on calibration-data diversity.
To address this concern, we calibrate the OPC variant using only PG19~\cite{rae2020compressive}, a corpus of long-form literary text, and evaluate it on the full LongBench~\cite{bai2024longbench} benchmark spanning six diverse task categories, including code, synthetic reasoning, and multi-document QA, which are absent from the calibration data.
As shown in Table~\ref{tab:ablation_domain}, the PG19-only OPC variant achieves an average score of 30.43, comparable to both full attention (30.57) and the mixed-domain OPC variant (30.39).
It also shows no systematic degradation on out-of-domain tasks such as Code (67.06 vs. 67.77) or Synthetic (4.71 vs. 4.78), suggesting that calibration-data diversity has only marginal impact on OPC performance.

\begin{table}[!hptb]
  \centering
  \scriptsize
  \setlength{\tabcolsep}{2pt}
  \caption{Ablation study on OPC robustness under calibration-domain shift using LongBench~\cite{bai2024longbench}.}
  \label{tab:ablation_domain}
  \resizebox{\columnwidth}{!}{%
  \begin{tabular}{l|cccccc|c}
    \toprule
    Methods                 & S-Doc & M-Doc & Sum.  & F-shot & Syn.  & Code  & Avg. \\
    \midrule
    Full Attention          & 17.97 & 9.57  & 17.97 & 69.15  & 4.78  & 67.77 & 30.57 \\
    BinaryPC w/ OPC (mix)   & 17.78 & 9.66  & 17.50 & 69.10  & 5.44  & 66.66 & \underline{30.39} \\
    BinaryPC w/ OPC (PG19 only) & 18.26 & 9.47  & 17.78 & 68.92  & 4.71  & 67.06 & \textbf{30.43} \\
    \bottomrule
  \end{tabular}
  }
% \vspace{-0.1in}
\end{table}

\section{Conclusion}

\label{sec:conclusion}

We presented BinaryPC, a training-free and data-aware hashing-based sparse attention for efficient long-context LLM inference. By constructing compact hash codes that explicitly preserve the structural information of data under the hashing projection, BinaryPC enables efficient and accurate token retrieval using GPU-friendly bit-parallel operations, substantially reducing the computational cost of attention during decoding. An error-aware safeguard further ensures robust recall by explicitly retaining hard-to-hash tokens.
Extensive experiments across multiple model families and benchmarks demonstrate that BinaryPC preserves accuracy while reducing effective KV retrieval and attention computation. 
Compared to sparse attention baselines, BinaryPC consistently achieves superior task performance, while exhibiting strong scalability and stability from 8K to 128K context lengths. Moreover, BinaryPC delivers significant decoding speedups on modern GPUs without requiring model-specific training or architectural modifications.
Overall, these results highlight BinaryPC as a lightweight, practical, and scalable hashing-based approach for long-context LLM inference.

\newpage
\section*{Acknowledgements}
This work was supported by the National Key Research and Development Program of China (No. 2025YFE0113500), the National Science Fund for Distinguished Young Scholars (No. 62525605), and the National Natural Science Foundation of China (No. U25B2066, No. 62506313).

\section*{Impact Statement}

\label{sec:impact}

The deployment of long-context large language models (LLMs) is frequently constrained by substantial computational demands during decoding, limiting their accessibility for real-world applications that require processing extended sequences. This work introduces BinaryPC, a training-free sparse attention that addresses these constraints by enabling efficient KV cache retrieval through compact 64-bit binary hashing. By leveraging data-aware hashing projections and GPU-optimized bitwise operations, BinaryPC significantly reduces attention computation while preserving task accuracy.
% 长上下文大语言模型（LLM）的部署常常受到解码阶段巨大计算需求的制约，限制了其在需要处理长序列的实际应用中的可及性。本工作提出 BinaryPC——一种无需训练的稀疏注意力框架，通过紧凑的 64 位二值哈希实现高效的 KV 缓存检索，从而解决上述限制。通过利用数据感知的线性投影和 GPU 优化的位运算，BinaryPC 在保持任务准确率的同时，将注意力计算量显著降低至全注意力的约 2\%。

The societal implications of BinaryPC are multifaceted. By enabling efficient long-context inference on existing hardware, this work democratizes access to advanced LLM capabilities, empowering smaller organizations, independent researchers, and practitioners in resource-constrained settings. The ability to achieve up to 3.17$\times$ throughput improvements and 5.04$\times$ speedup in Flash Attention fallback scenarios translates directly to reduced energy consumption and lower operational costs for large-scale deployments. This contributes to the broader sustainability goals of AI by minimizing the environmental footprint associated with serving long-context workloads.
% BinaryPC 的社会意义是多方面的。通过在现有硬件上实现高效的长上下文推理，本工作使先进的 LLM 能力更加普惠，赋能小型组织、独立研究者以及资源受限环境下的从业者。高达 3.17 倍的吞吐量提升和 Flash Attention 回退场景下 5.04 倍的加速，直接转化为大规模部署中能耗的降低和运营成本的削减。这有助于实现 AI 领域更广泛的可持续发展目标，最大程度减少服务长上下文工作负载所带来的环境影响。
Furthermore, the training-free nature of BinaryPC eliminates the need for model-specific optimization or extensive calibration data, making it readily applicable across diverse model families without additional engineering overhead. This model agnosticism promotes equitable access to efficient inference techniques, ensuring that advancements in sparse attention are not confined to well-resourced institutions with the capacity for per-model fine-tuning.
% 此外，BinaryPC 无需训练的特性消除了对模型特定优化或大量校准数据的依赖，使其无需额外工程开销即可便捷地应用于各类模型家族。这种架构无关性促进了高效推理技术的公平获取，确保稀疏注意力领域的进展不局限于那些具备针对每个模型进行微调能力的资源丰富机构。

No specific ethical concerns or societal risks are associated with the proposed method. BinaryPC does not alter the fundamental capabilities or outputs of the underlying LLMs; it solely optimizes the computational efficiency of attention mechanisms. As such, it provides a pathway toward more equitable and responsible AI deployment, ensuring that advances in long-context LLM inference are accessible and sustainable across diverse sectors and applications.
% 本方法不涉及特定的伦理问题或社会风险。BinaryPC 不改变底层 LLM 的基本能力或输出，仅优化注意力机制的计算效率。因此，它为更加公平和负责任的 AI 部署提供了一条路径，确保长上下文 LLM 推理领域的进展能够在各行各业和多种应用场景中可获取且可持续。

\bibliography{refs}
\bibliographystyle{icml2026}

%%%%%%%%%%%%%%%%%%%%%%%%%%%%%%%%%%%%%%%%%%%%%%%%%%%%%%%%%%%%%%%%%%%%%%%%%%%%%%%
%%%%%%%%%%%%%%%%%%%%%%%%%%%%%%%%%%%%%%%%%%%%%%%%%%%%%%%%%%%%%%%%%%%%%%%%%%%%%%%
% APPENDIX
%%%%%%%%%%%%%%%%%%%%%%%%%%%%%%%%%%%%%%%%%%%%%%%%%%%%%%%%%%%%%%%%%%%%%%%%%%%%%%%
%%%%%%%%%%%%%%%%%%%%%%%%%%%%%%%%%%%%%%%%%%%%%%%%%%%%%%%%%%%%%%%%%%%%%%%%%%%%%%%

\newpage
\onecolumn
\appendix
\section{ Implementation Details}

\subsection{Baseline Methods}
\label{sec:Baseline_Methods}
\textbf{TOPK (Oracle).} TOPK selects the top-$k$ keys using exact full-precision attention scores under a fixed 2\% token budget, then restricts attention to this selected subset.

\textbf{PyramidKV.} PyramidKV employs a pyramid-shaped budget allocation strategy where shallow layers retain more KV pairs than deeper layers. We set the local window size to 8, pooling kernel size to 5 with average pooling, and the steepness parameter $\beta = 20$. The first two layers remain dense to preserve critical early-context information.

\textbf{Cake.} Cake dynamically allocates layer-wise budgets based on attention statistics from the prefill phase. We set the local window size to 8. Budget allocation parameters $\tau_1$ and $\tau_2$ follow the settings for Llama-3.1-8B-Instruct and Mistral-7B-Instruct-v0.3, with the eviction parameter $\gamma = 200.0$. The first two layers remain dense. Evaluations are conducted only on the models where detailed parameters are publicly available.

\textbf{CompressKV.} CompressKV utilizes a subset of important attention heads to estimate token importance. We set the local window size to 8 and pooling kernel size to 5. The number of active heads $k$ is set to 4. Head and layer weights are pre-computed offline using a calibration set. The first two layers remain dense. Evaluations are conducted only on the models where detailed parameters are publicly available.

\textbf{Quest.}  Quest performs query-aware token selection via chunk-based importance estimation. We set the chunk size to 16. During decoding, top-$k$ chunks are identified based on the maximum attention weight within each chunk. The first two layers remain dense.
% Unlike permanent compression, Quest retains the full KV cache and dynamically retrieves relevant chunks for each query.

\textbf{MagicPIG.} We adopt the default configuration, where LSH retrieval uses $K=10$ hash bits and $L=150$ tables, with a local window size of 64 and 4 sink tokens. The collision threshold is set to 2 and error correction is enabled via the anns\_es mode. The first and middle layers remain dense.

\textbf{Spotlight.} We adopt the official implementation with a 2\% token budget and a minimum retention of 20 tokens. The first two layers remain dense. Due to the training requirement, we conduct evaluations only on Llama-3-8B using the publicly released weights.

\subsection{Dataset Details}
\label{sec:dataset_details}
% 数据集详情
We evaluate BinaryPC across four context regimes using diverse benchmarks that comprehensively assess task accuracy, retrieval capability, and scalability.

\textbf{Short-context: LM-Eval-Harness.} We employ three representative tasks from LM-Eval-Harness~\citep{eval-harness}: GSM8K-CoT~\citep{cobbe2021training} for mathematical reasoning with chain-of-thought prompting, MMLU-Flan-CoT-Fewshot~\citep{hendrycks2021measuring} for multi-task language understanding, and CoQA~\citep{reddy2019coqa} for conversational question answering. Following the standard few-shot protocol, we sample 1000 instances per subtask and report accuracy for GSM8K and MMLU, and F1 for CoQA. 
% BinaryPC uses a 5\% token budget in this regime, with the number of retained tokens capped at 64.

\textbf{Medium-context: LongBench.} LongBench~\citep{bai2024longbench} is the first benchmark for bilingual, multitask, and comprehensive assessment of long context understanding capabilities. Featuring both Chinese and English languages, LongBench contains 16 datasets spanning six task categories: Single-Document QA (NrtvQA, Qasper, MF-en), Multi-Document QA (HotpotQA, 2WikiMQA, Musique), Summarization (GovReport, QMSum, MultiNews), Few-shot Learning (TREC, TriviaQA, SAMSum), Synthetic Retrieval (PCount, PRe), and Code Completion (Lcc, RB-P). The benchmark includes 14 English tasks, 5 Chinese tasks, and 2 code tasks, with the average length of most tasks ranging from 5K to 15K tokens. 
% BinaryPC uses a 2\% token budget in this regime, with at most 307 retained tokens, which is substantially smaller than the 1K token budgets commonly used by alternative methods.

\textbf{Long-context: InfiniteBench and LongBench v2.} InfiniteBench~\citep{zhang2024infty} is a cutting-edge benchmark tailored for evaluating the capabilities of language models to process, understand, and reason over super long contexts. It is designed to push the boundaries of language models by testing them against a context length of 100K+, which is 10 times longer than traditional datasets. We select seven task categories for evaluation: English summarization (En.Sum), English question answering (En.QA), English multiple-choice (En.MC), speaker identification (En.Dia), Chinese question answering (Zh.QA), special-integer retrieval (Math.F), and passkey retrieval (R.PK). LongBench v2~\citep{bai2025longbench} is designed to assess the ability of LLMs to handle long-context problems requiring deep understanding and reasoning across real-world multitasks. It consists of 503 challenging multiple-choice questions with context lengths ranging from 8K to 2M words, across six major task categories: single-document QA, multi-document QA, long in-context learning, long-dialogue history understanding, code repository understanding, and long structured data understanding. Results are stratified by difficulty (Easy/Hard) and context length (Short/Medium/Long). 
% For all evaluations, the input context is truncated to a maximum length of 128K tokens. Even at 128K, BinaryPC uses a 2\% token budget, which is comparable to the 2K token budgets used by alternative methods.

\textbf{Scalability from 8K to 128K: RULER and Needle-in-a-Haystack.} RULER~\citep{hsieh2024ruler} generates synthetic examples to evaluate long-context language models with configurable sequence length and task complexity. It consists of 13 complex tasks across 4 task categories, including retrieval, multi-hop tracing, aggregation, and question answering, evaluating long-context capabilities beyond simple in-context recall. We configure test sets scaling from 8K to 128K tokens to systematically evaluate performance degradation as context length increases. 
Needle-in-a-Haystack (NIAH)~\citep{needle-in-haystack} is a long-context retrieval benchmark that evaluates LLM performance with extended context windows, where relevant information is distributed at varying depths. 
% BinaryPC employs a 2\% token budget, retaining only 163 tokens at an 8K context length, while its budget scales to remain comparable with the fixed 2K tokens used by alternative methods at a 128K context length.

% \input{figures/Longbench_llama38b_bar.tex}

\begin{figure*}[t]
\centering
\begin{minipage}{0.32\linewidth}
    \centering 
    \includegraphics[width=\textwidth]{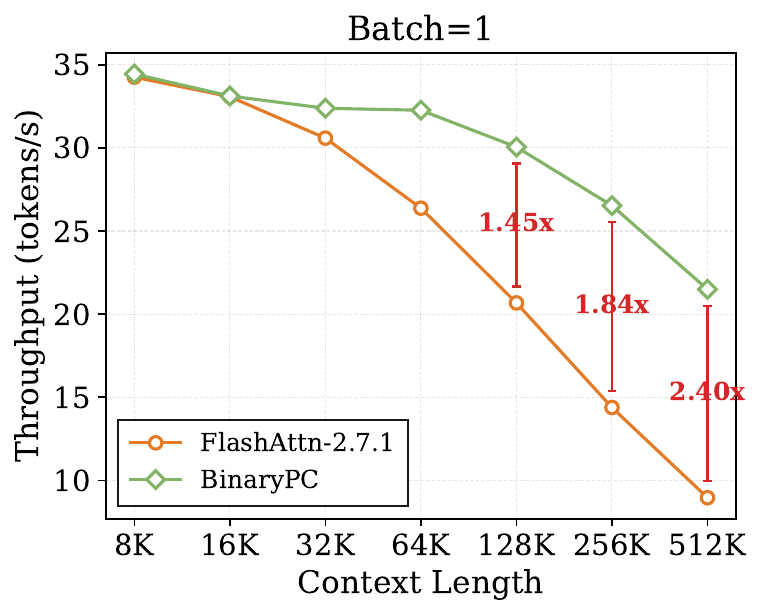}
\end{minipage}
\hfill
\begin{minipage}{0.32\linewidth}
    \centering
    \includegraphics[width=\textwidth]{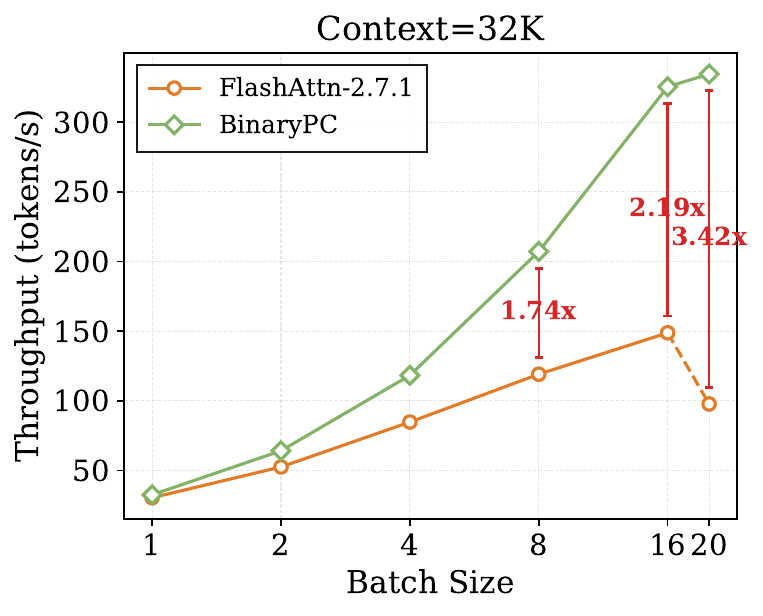}
\end{minipage}
\hfill
\begin{minipage}{0.32\linewidth}
    \centering
    \includegraphics[width=\textwidth]{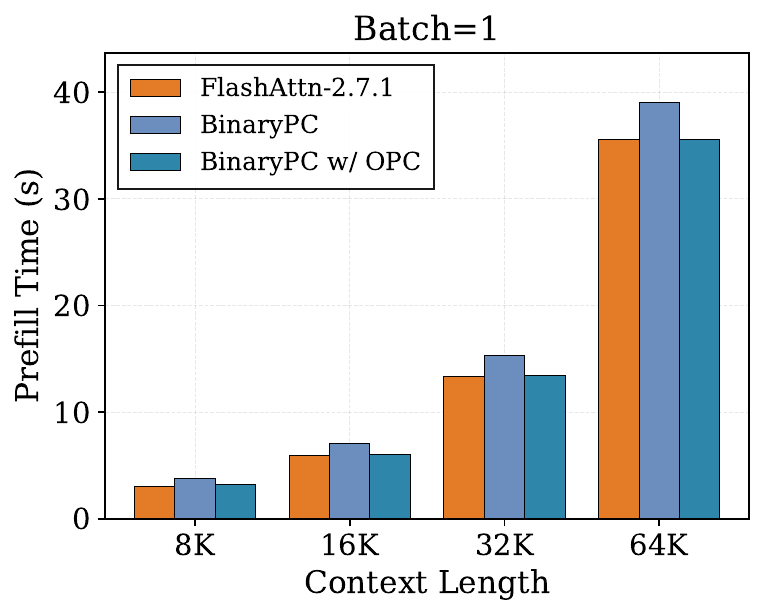}
\end{minipage}
% \caption{Efficiency evaluation on RTX 3090 GPUs. Left: Decoding throughput across context lengths from 8K to 512K tokens at batch size 1. Middle: Throughput scaling with batch sizes from 1 to 20 at a fixed 32K context length. Right: Prefill stage latency comparison between BinVortex variants and FlashAttention-2.}
\caption{Efficiency evaluation on consumer-grade GPUs. Left: Decoding throughput across context lengths from 8K to 512K tokens at batch size 1. Middle: Throughput scaling with batch sizes from 1 to 20 at a fixed 32K context length. Right: Prefill stage latency comparison between BinVortex variants and FlashAttention-2.}

\label{fig:efficiency_3090}
\end{figure*}

\begin{table*}[t]
  \centering
  \scriptsize
  \setlength{\tabcolsep}{2pt}

    \caption{Performance on InfiniteBench~\cite{zhang2024infty} (left) and LongBench v2~\cite{bai2025longbench} (right).}
  \label{tab:infinitebench_qwen}
  \resizebox{\textwidth}{!}{%
  \begin{tabular}{l|c|ccccccc|c|ccccc|c}
    \toprule
        \multirow{2}{*}{Methods} & & \multicolumn{8}{c|}{InfiniteBench} & \multicolumn{6}{c}{LongBench v2} \\
         &Token      &En.Sum  &En.QA &En.MC &En.Dia&Zh.QA &Math.F&R.PK & Avg.                         & Easy & Hard & Short& Medium& Long& Avg. \\
          \midrule
      \textbf{Qwen2.5-7B}  &Full       &34.02   &12.14 &68.12 &19.00 &9.90  &40.29 &34.58 &31.15                        &34.9  &27.7  &37.8  &26.5  &25.9  &30.4   \\
              PyramidKV    &2K         &27.48   &11.17 &67.69 &10.00 &9.20  &39.43 &34.58 &28.50                        &35.4  &27.3  &37.2  &26.0  &27.8  &\underline{30.4}   \\
              Quest        &2K         &15.57   &11.60 &67.69 &13.00 &10.56 &40.00 &34.58 &27.57                        &31.8  &25.7  &37.8  &22.3  &23.1  &28.0   \\
              MagicPIG     &Default    &27.31   &11.35 &69.00 &14.00 &10.45 &29.71 &10.17 &24.57                        &32.8  &25.7  &31.7  &28.4  &23.1  &28.4   \\
              BinaryPC    &2\%        &31.40   &11.43 &67.25 &14.00 &10.45 &39.71 &34.58 &\textbf{29.83}               &35.9  &28.3  &38.9  &27.4  &25.9  &\textbf{31.2}   \\
              BinaryPC w/ OPC &2\%    &31.07   &11.57 &69.00 &15.50 &10.22 &33.43 &34.41 &\underline{29.31}            &35.4  &27.3  &38.9  &27.0  &23.1  &\underline{30.4}   \\
        \bottomrule
  \end{tabular}
  }% end resizebox
\end{table*}

\begin{table*}[t]
    \centering
    \scriptsize
    \setlength{\tabcolsep}{2.5pt}
    \caption{LongBench~\cite{bai2024longbench} evaluation across 16 datasets.}
    \label{tab:longbench}
    \resizebox{\textwidth}{!}{%
    \begin{tabular}{lc*{17}{c}}
        \toprule
        & & \multicolumn{3}{c}{Single-Document QA}
        & \multicolumn{3}{c}{Multi-Document QA}
        & \multicolumn{3}{c}{Summarization}
        & \multicolumn{3}{c}{Few-shot Learning}
        & \multicolumn{2}{c}{Synthetic}
        & \multicolumn{2}{c}{Code}
        & \\ % 最后一列是 Avg，不分组
        \cmidrule(lr){3-5}\cmidrule(lr){6-8}\cmidrule(lr){9-11}%
        \cmidrule(lr){12-14}\cmidrule(lr){15-16}\cmidrule(lr){17-18}
        Method & Token &
        \rotatebox[origin=c]{30}{NrtvQA} &
        \rotatebox[origin=c]{30}{Qasper} &
        \rotatebox[origin=c]{30}{MF-en} &
        \rotatebox[origin=c]{30}{HotpotQA} &
        \rotatebox[origin=c]{30}{2WikiMQA} &
        \rotatebox[origin=c]{30}{Musique} &
        \rotatebox[origin=c]{30}{GovReport} &
        \rotatebox[origin=c]{30}{QMSum} &
        \rotatebox[origin=c]{30}{MultiNews} &
        \rotatebox[origin=c]{30}{TREC} &
        \rotatebox[origin=c]{30}{TriviaQA} &
        \rotatebox[origin=c]{30}{SAMSum} &
        \rotatebox[origin=c]{30}{PCount} &
        \rotatebox[origin=c]{30}{PRe} &
        \rotatebox[origin=c]{30}{Lcc} &
        \rotatebox[origin=c]{30}{RB-P} &
        Avg \\
        \midrule
                                         % NrtvQA  Qasper  MF-en   Hotpot  2Wiki   Musi    Gov     QMSum   Multi   TREC    Trivi   SAMSum   PCount PRe     Lcc     RB-P    Avg
            \textbf{Llama-3-8B}  &Full   & 16.94 & 14.18 & 23.62 &  9.97 & 11.75 &  7.15 & 29.85 & 23.05 &  1.26 & 70.50 & 91.20 & 45.26 &  1.70 &  8.29 & 70.83 & 64.49 & 30.63 \\
            MagicPIG             &Default& 15.54 & 14.20 & 21.13 &  9.45 & 11.30 &  6.78 & 25.84 & 23.45 &  0.72 & 69.50 & 90.97 & 44.13 &  1.95 &  8.38 & 70.06 & 63.04 & 29.78 \\
            Spotlight            &2\%    & 15.44 & 14.94 & 24.39 &  9.12 & 11.81 &  6.64 & 29.34 & 22.47 &  2.57 & 72.00 & 91.20 & 44.61 &  2.50 &  7.62 & 67.37 & 63.01 & \underline{30.31} \\
            BinaryPC             &2\%    & 15.39 & 14.10 & 21.25 &  9.27 & 12.27 &  6.99 & 29.92 & 23.27 &  1.54 & 70.50 & 91.20 & 44.69 &  2.00 &  7.04 & 69.85 & 63.61 & 30.18 \\
            BinaryPC w/ OPC      &2\%    & 16.92 & 13.09 & 22.81 &  9.60 & 12.01 &  7.48 & 29.90 & 22.81 &  0.97 & 70.50 & 91.20 & 45.36 &  1.62 &  7.67 & 70.28 & 63.45 & \textbf{30.35} \\
            \midrule
                                  % NrtvQA  Qasper  MF-en   Hotpot  2Wiki   Musi    Gov     QMSum   Multi   TREC    Trivi   SAMSum   PCount PRe     Lcc     RB-P    Avg
            \textbf{Llama-3.1-8B}&Full & 29.56 & 44.70 & 55.93 & 57.82 & 48.95 & 32.61 & 34.45 & 25.51 & 26.88 & 72.50 & 91.15 & 44.10 & 11.50 & 99.50  & 62.97 & 56.16 & 49.64 \\
            PyramidKV            &1K& 30.45 & 41.75 & 55.38 & 57.13 & 49.25 & 30.68 & 26.69 & 23.85 & 25.41 & 70.50 & 91.18 & 42.71 & 11.25 & 99.50  & 61.84 & 52.85 & 48.15 \\
            Cake                 &1K& 29.99 & 41.69 & 56.82 & 57.16 & 48.06 & 32.00 & 27.66 & 24.31 & 25.63 & 70.50 & 91.52 & 43.33 & 10.92 & 100.00 & 61.53 & 55.10 & 48.51 \\
            CompressKV           &1K& 29.61 & 43.52 & 57.21 & 57.60 & 48.47 & 31.93 & 28.18 & 24.43 & 25.67 & 71.00 & 91.18 & 43.70 & 11.25 & 99.50  & 62.59 & 55.34 & 48.82 \\
            Quest                &1K& 30.08 & 43.60 & 53.66 & 57.45 & 48.75 & 32.88 & 34.46 & 25.22 & 26.83 & 72.25 & 89.99 & 42.78 & 11.10 & 99.50  & 60.85 & 53.44 & 48.93 \\
            MagicPIG             &Default& 30.47 & 42.57 & 56.13 & 57.26 & 48.33 & 32.80 & 32.94 & 25.16 & 26.24 & 72.50 & 90.86 & 42.93 & 11.17 & 98.50  & 60.80 & 55.43 & 49.01 \\
            BinaryPC             &2\%& 29.09 & 45.10 & 56.57 & 58.14 & 48.84 & 32.00 & 34.92 & 25.13 & 26.96 & 72.50 & 90.92 & 44.06 & 11.00 & 99.50  & 62.09 & 55.12 & \textbf{49.50} \\
            BinaryPC w/ OPC      &2\%& 30.06 & 43.81 & 56.87 & 58.31 & 48.46 & 31.95 & 34.57 & 25.20 & 26.45 & 72.50 & 91.12 & 43.16 & 11.25 & 99.50  & 62.71 & 56.12 & \textbf{49.50} \\
            \midrule
                                  % NrtvQA  Qasper  MF-en   Hotpot  2Wiki   Musi    Gov     QMSum   Multi   TREC    Trivi   SAMSum   PCount PRe     Lcc     RB-P    Avg
            \textbf{Mistral-7B}   &Full& 27.35 & 38.03 & 50.50 & 51.41 & 38.89 & 28.69 & 34.05 & 25.32 & 26.51 & 76.00 & 88.89 & 47.33 &  6.50 & 97.50 & 59.03 & 60.97 & 47.31 \\
            PyramidKV             &1K& 26.18 & 34.89 & 49.80 & 49.78 & 36.98 & 26.09 & 27.04 & 23.54 & 25.18 & 74.50 & 89.86 & 45.97 &  4.00 & 96.00 & 57.63 & 58.12 & 45.35 \\
            Cake                  &1K& 26.74 & 36.04 & 50.28 & 50.29 & 37.07 & 27.08 & 28.56 & 24.43 & 26.17 & 74.00 & 89.31 & 47.00 &  5.50 & 96.50 & 58.74 & 60.01 & 46.11 \\
            CompressKV            &1K& 27.26 & 37.10 & 51.14 & 50.10 & 38.20 & 27.95 & 28.98 & 24.80 & 25.79 & 76.00 & 89.57 & 46.46 &  5.50 & 97.00 & 58.66 & 60.22 & 46.55 \\
            Quest                 &1K& 26.35 & 36.80 & 49.42 & 47.56 & 38.15 & 26.95 & 32.26 & 24.44 & 26.40 & 75.50 & 89.08 & 45.29 &  5.32 & 93.75 & 58.00 & 58.71 & 45.87 \\
            MagicPIG              &Default& 26.78 & 37.51 & 51.31 & 50.76 & 37.80 & 28.18 & 33.40 & 25.67 & 26.33 & 76.00 & 89.09 & 47.18 &  6.00 & 96.50 & 59.64 & 60.84 & 47.06 \\
            BinaryPC             &2\%& 26.84 & 37.65 & 50.46 & 50.65 & 37.90 & 27.42 & 34.58 & 24.89 & 26.64 & 75.50 & 89.41 & 46.56 &  6.50 & 97.50 & 58.84 & 61.80 & \underline{47.07} \\
            BinaryPC w/ OPC        &2\%& 27.53 & 39.57 & 50.60 & 50.30 & 37.64 & 28.69 & 34.27 & 25.12 & 26.40 & 76.00 & 89.86 & 47.27 &  5.50 & 97.50 & 59.10 & 60.45 & \textbf{47.24} \\
            \midrule
                                  % NrtvQA  Qasper  MF-en   Hotpot  2Wiki   Musi    Gov     QMSum   Multi   TREC    Trivi   SAMSum   PCount PRe     Lcc     RB-P    Avg
            \textbf{Qwen2.5-7B}   &Full& 28.80 & 47.88 & 50.38 & 58.74 & 54.09 & 33.80 & 33.37 & 23.51 & 24.28 & 76.50 & 84.07 & 44.61 &  7.00 & 100.00 & 46.19 & 39.24 & 47.03 \\
            PyramidKV             &1K& 28.29 & 44.58 & 49.01 & 56.87 & 52.69 & 31.66 & 25.96 & 20.71 & 21.99 & 74.50 & 84.08 & 44.11 &  7.00 & 100.00 & 43.94 & 35.32 & 45.04 \\
            Quest                 &1K& 29.54 & 46.86 & 48.97 & 58.07 & 53.96 & 33.69 & 33.17 & 22.82 & 23.81 & 75.50 & 82.13 & 43.48 &  7.50 & 89.50  & 45.22 & 36.24 & 45.65 \\
            MagicPIG              &Default& 28.96 & 42.84 & 45.61 & 52.92 & 51.06 & 29.27 & 30.95 & 22.63 & 22.65 & 73.50 & 82.97 & 44.89 &  6.50 & 96.50  & 44.39 & 38.05 & 44.61 \\
            BinaryPC             &2\%& 29.94 & 45.99 & 49.60 & 58.01 & 53.93 & 32.23 & 33.50 & 23.35 & 24.72 & 77.00 & 85.50 & 44.28 &  7.50 & 100.00 & 42.01 & 36.14 & \textbf{46.48} \\
            BinaryPC w/ OPC       &2\% & 28.35 & 46.52 & 49.79 & 57.10 & 53.39 & 32.12 & 34.20 & 23.36 & 24.60 & 77.00 & 85.90 & 44.25 &  8.00 & 100.00 & 39.94 & 36.20 & \underline{46.30} \\
        \bottomrule
    \end{tabular}
    }% end resizebox
\end{table*}

\subsection{Procedure Overview of BinaryPC}
\label{sec:binarypc_procedure_overview}

\begin{figure*}[t]
  \centering
  \includegraphics[width=0.92\textwidth]{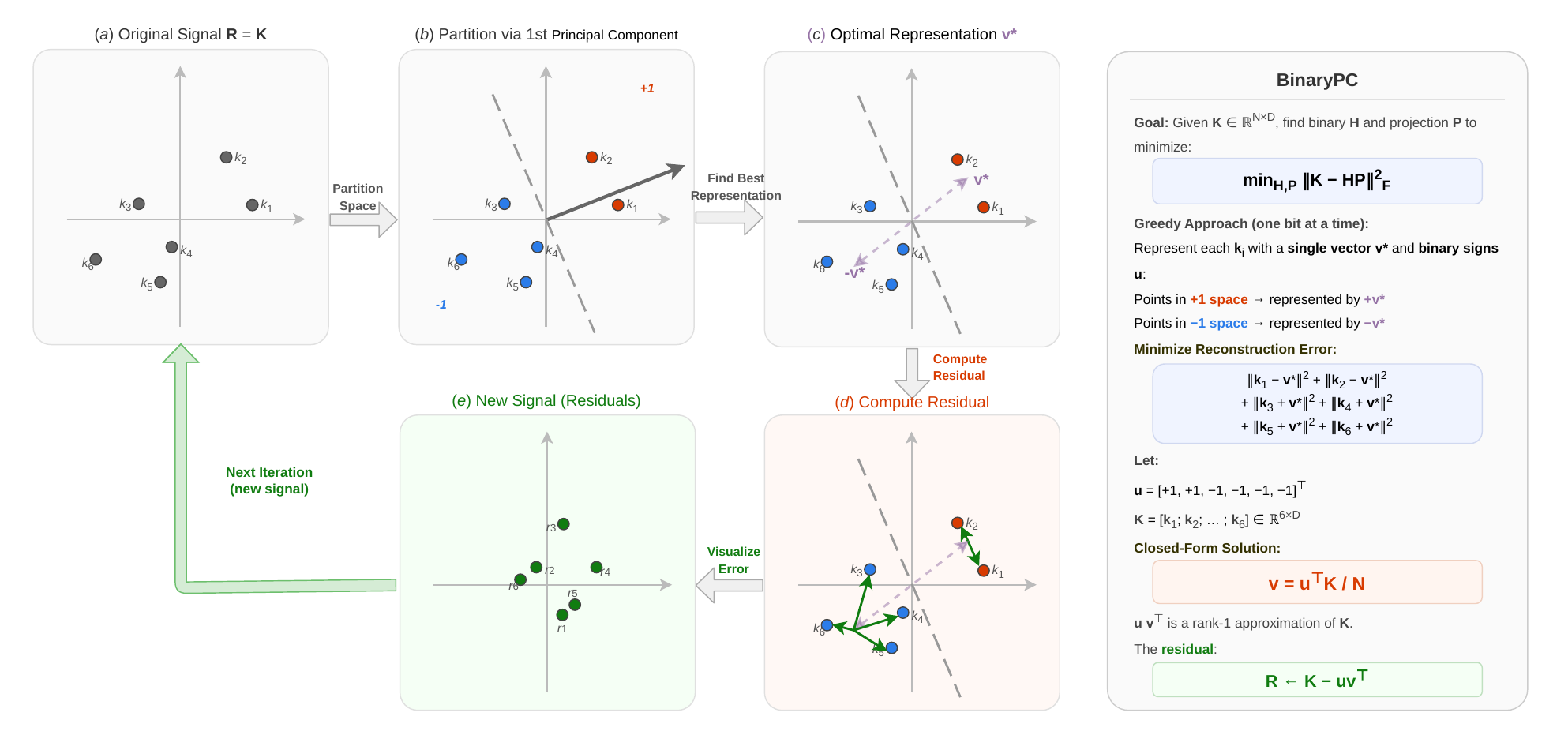}
  % \caption{Procedure overview of BinaryPC. Starting from the residual $\mR=\mK$, each iteration maps every key vector to a binary sign $u_i \in \{+1,-1\}$ through data-aware space partitioning and computes a shared projection vector $\vv$. For a fixed binary vector $\vu$, $\vv=\vu^\top\mR/N$ minimizes the rank-1 approximation error $\|\mR-\vu\vv^\top\|_F^2$. The approximated component is then removed from the residual, and the next iteration repeats this procedure to obtain one additional bit.}
  \caption{Procedure overview of BinaryPC. Each iteration partitions the current residual into binary signs, computes one shared projection component, removes the corresponding rank-1 approximation, and passes the remaining residual to the next iteration to generate the next bit.}
  \label{fig:binarypc_process}
  
\end{figure*}

Figure~\ref{fig:binarypc_process} provides an intuitive overview of the BinaryPC procedure. Unlike LSH, which partitions the space using data-independent random hyperplanes, BinaryPC derives hashing directions from the principal directions of the key vectors, aligning partitions with the intrinsic data structure and enabling more faithful similarity preservation.

BinaryPC iteratively solves the reconstruction objective $\min_{\mH,\mP}\|\mK-\mH\mP\|_F^2$ one bit at a time. Starting from the residual $\mR=\mK$, each iteration maps every data point $\vk_i \in \mK$ to a binary sign $u_i \in \{+1,-1\}$ through space partitioning and computes a shared projection vector $\vv$. For a fixed binary vector $\vu$, the optimal projection vector has the closed-form solution $\vv = \vu^\top\mR/N$, where $N$ is the number of data points. 
This creates a rank-1 approximation $\vu\vv^\top$. The algorithm then updates the residual as $\mR \leftarrow \mR-\vu\vv^\top$, which is passed to the next iteration as the new signal.

\subsection{CUDA Optimizations on Bitwise Operations}
\label{sec:cuda_kernel_details}

We implement three CUDA kernels for BinaryPC.

\textbf{Hashing kernel.}
Each key vector ($D=128$, BF16) is projected to 64 binary values and directly packed into a single \texttt{int64} word.
Let $\vp = \vq \mP^\top$ for notation simplicity.

\textbf{Packing kernel.}
Each 64-entry quantized vector $\vp$ with values in $[-127,127]$ is packed into one \texttt{int64} $\vs$ storing the signs and seven \texttt{int64} values $\vm[j]$ storing the magnitude bits. 

\textbf{Scoring kernel.}
Using one \texttt{int64} hash code and the eight packed \texttt{int64} values ($\vs$ and $\vm[j]$), this kernel computes hash scores for all query-key pairs per head.
Under this 8-\texttt{int64} representation, each $p_i$ is expressed as
\begin{equation}
    p_i = s_i \sum_{j=1}^{7} 2^j m[j]_i.
\end{equation}
The inner product between $\vp$ and $\vh$ in Eq.~\eqref{eq:retrieval_expansion} can be written as
\begin{equation}
    \sum_{i=1}^{64} h_i s_i \sum_{j=1}^{7} 2^j m[j]_i,
\end{equation}
which can be arranged as
\begin{equation}
    \sum_{j=1}^{7} 2^j \sum_{i=1}^{64} h_i s_i m[j]_i.
\end{equation}
Since each $h_i$ is either $+1$ or $-1$, we use $\vs$ XOR $\vh$ to determine all $h_i s_i$ in a single CUDA instruction; denote the resulting \texttt{int64} data as $\vw$. Then, each $\sum_{i=1}^{64} h_i s_i m[j]_i$ can be computed via
\begin{equation}
    \texttt{\_\_popcll}(\vm[j] \mathbin{\&} \vw)
    -
    \texttt{\_\_popcll}(\vm[j] \mathbin{\&} ({\sim}\vw)).
\end{equation}
This counts the number of positive $(h_i s_i m[j]_i)$ contributions and subtracts the number of negative ones. To optimize memory I/O, the eight \texttt{int64} values are shared within each CUDA block to reduce data movement.

\begin{table}[t]
  \centering
  \footnotesize
  \setlength{\tabcolsep}{8pt}
  \caption{Average cosine similarity between sparse and full attention outputs at the first decoding step on Llama-3.1-8B-Instruct.}
  \label{tab:attention_output_similarity}
  \begin{tabular}{lcccc}
    \toprule
    Method & 8K & 16K & 32K & 64K \\
    \midrule
    BinaryPC (2.0\%) & 0.9835 & 0.9811 & 0.9870 & 0.9895 \\
    MagicPIG (2.2\%) & 0.9388 & 0.9651 & 0.9595 & 0.9725 \\
    \bottomrule
  \end{tabular}
\end{table}

\section{Additional Experiment Results}

\subsection{Binary Representation and Approximation Quality}
\label{sec:approximation_quality_appendix}

BinaryPC does not simply binarize vectors by taking the sign. Instead, it greedily minimizes the reconstruction objective $\|\mK - \mH\mP\|_F^2$ via iterative rank-1 binary decomposition. The rows of the projection matrix $\mP$ represent components, and each key vector is represented as a weighted sum of these components, where the weights are $-1$ or $+1$. Therefore, the product $\mH\mP$ reconstructs both directional and magnitude information.

\textbf{Approximation error of $\mK$.}
As shown in Figure~\ref{fig:residual_norms}, as the number of binary principal components increases, the residual $\|\mK - \mH\mP\|_F$ decreases progressively, demonstrating the effectiveness of our reconstruction-based approach.

\textbf{Retrieval quality via Needle-in-a-Haystack.}
Figures~\ref{fig:niah}, \ref{fig:appendix_Qwen_images}, and~\ref{fig:appendix_llama_images} provide retrieval-accuracy comparisons across needle depths and context lengths. BinaryPC achieves near-perfect retrieval across all needle depths and context lengths, whereas competing methods exhibit systematic failures at specific depth ranges.

\textbf{Approximation error of attention output.}
We measure the average cosine similarity between sparse and full attention outputs at the first decoding step on Llama-3.1-8B-Instruct. As shown in Table~\ref{tab:attention_output_similarity}, under comparable token budgets of approximately 2\%, BinaryPC consistently achieves higher cosine similarity than MagicPIG across all context lengths. This demonstrates that BinaryPC's reconstruction-based hashing provides superior quality for token selection.

\subsection{Additional Efficiency Evaluation}
\label{sec:Efficiency_appendix}

% To validate the generality of BinaryPC across different hardware configurations, we conduct additional efficiency experiments on eight NVIDIA RTX 3090 GPUs (24 GB VRAM each) and analyze the runtime during the prefill stage.
To validate the generality of BinaryPC across different hardware configurations, we conduct additional efficiency experiments on eight consumer-grade GPUs and analyze the runtime during the prefill stage.

% \textbf{Decoding Throughput.} Figure~\ref{fig:efficiency_3090} (left and middle) reports the decoding throughput on RTX 3090 GPUs. Under batch size 1, BinaryPC achieves comparable throughput to FlashAttention-2 at shorter context lengths, while attaining up to 2.4$\times$ speedup at 512K tokens. At a fixed 32K context length, BinaryPC achieves 2.19$\times$ improvement at batch size 16. Notably, RTX 3090 GPUs have 82 Streaming Multiprocessors, and when batch size exceeds 16, FlashAttention-2 switches from the highly parallel Split-K kernel to the standard kernel, resulting in throughput degradation. At batch size 20, BinaryPC achieves 3.42$\times$ speedup. These results confirm the generality of BinaryPC across different hardware configurations.
\textbf{Decoding Throughput.} Figure~\ref{fig:efficiency_3090} (left and middle) reports the decoding throughput on consumer-grade GPUs. Under batch size 1, BinaryPC achieves comparable throughput to FlashAttention-2 at shorter context lengths, while attaining up to 2.4$\times$ speedup at 512K tokens. At a fixed 32K context length, BinaryPC achieves 2.19$\times$ improvement at batch size 16. On this hardware, when batch size exceeds 16, FlashAttention-2 switches from the highly parallel Split-K kernel to the standard kernel, resulting in throughput degradation. At batch size 20, BinaryPC achieves 3.42$\times$ speedup. These results confirm the generality of BinaryPC across different hardware configurations.

\textbf{Prefill Stage Analysis.} Figure~\ref{fig:efficiency_3090} (right) presents the runtime breakdown during the prefill stage. The prefill phase computes attention over the entire input context and constructs the binary hash codes for subsequent decoding. The offline calibrated variant (BinaryPC w/ OPC) achieves nearly identical prefill latency to FlashAttention-2, as it leverages pre-calibrated projection matrices without additional computation during inference. In contrast, the online variant (BinaryPC) incurs modest overhead for computing projection matrices from the input context. This one-time cost is amortized over multiple decoding steps, making it negligible for generation tasks with substantial output lengths.

\subsection{Additional Long-context Evaluation}
\label{sec:Long_context_appendix}

Table~\ref{tab:infinitebench_qwen} reports additional results on InfiniteBench~\cite{zhang2024infty} and LongBench v2~\cite{bai2025longbench} using Qwen2.5-7B-Instruct-1M~\cite{yang2025qwen2}. BinaryPC consistently outperforms all sparse attention baselines on both benchmarks, achieving near-full-attention accuracy without model-specific tuning.

\subsection{Ultra-Large-Scale Model Evaluation}
\label{sec:ultra_large_scale_appendix}

\textbf{Accuracy.}
We further evaluate BinaryPC on Llama-3-70B-Instruct~\cite{grattafiori2024llama} using LongBench~\cite{bai2024longbench}. As shown in Table~\ref{tab:llama70b_scalability}, BinaryPC matches the accuracy of full attention across LongBench task categories, demonstrating full-attention-level accuracy at the 70B scale.

\textbf{Memory overhead.}
The additional memory introduced by BinaryPC comes from binary hash codes and projection matrices. Each hash code uses 8 bytes per token per KV head when stored as an \texttt{int64}, while the $128 \times 64$ projection matrix is negligible for long sequences. In comparison, the BF16 KV cache uses 512 bytes per token per KV head, so the hash-code overhead is only $1/64$, or approximately 1.56\%, of the KV-cache size. Moving from Llama-3.1-8B-Instruct to Llama-3-70B-Instruct, the total number of KV-head instances across layers increases by 2.5$\times$ ($8 \times 32$ to $8 \times 80$), while model size grows by 8.75$\times$. Thus, the absolute BinaryPC memory overhead grows more slowly than model size, and the relative overhead with respect to the KV cache remains unchanged.

\textbf{Efficiency.}
From Llama-3.1-8B-Instruct to Llama-3-70B-Instruct, the GQA group size doubles from 4 to 8, while the number of KV heads per layer remains unchanged. Since both regular attention and BinaryPC are primarily bottlenecked by KV-cache access and lightweight hashing overhead, the single-attention-layer latency remains similar across the two model scales. Table~\ref{tab:llama70b_scalability} reports this comparison using a 2048-token BinaryPC budget. Although the model size increases by 8.75$\times$, the number of layers, and therefore the total attention runtime, increases by only 2.5$\times$ from 32 to 80 layers.

\begin{table}[t]
  \centering
  \footnotesize
  \setlength{\tabcolsep}{6pt}
  \caption{Ultra-large-scale evaluation of BinaryPC. The upper block reports LongBench~\cite{bai2024longbench} category-level accuracy on Llama-3-70B-Instruct~\cite{grattafiori2024llama}; the lower block reports single-attention-layer latency.}
  \label{tab:llama70b_scalability}
  \begin{tabular}{lcccccc}
    \toprule
    \multicolumn{7}{c}{\textbf{Accuracy on LongBench}} \\
    \midrule
    Method & S-Doc & M-Doc & Sum. & F-shot & Syn. & Code \\
    \midrule
    Full Attention-70B & 42.18 & 47.51 & 27.31 & 70.83 & 38.50 & 54.24 \\
    BinaryPC-70B       & 42.06 & 47.78 & 26.98 & 70.96 & 38.25 & 55.22 \\
    \addlinespace[0.35em]
    \midrule
    \multicolumn{7}{c}{\textbf{Single-attention-layer latency (ms)}} \\
    \midrule
    Method & Budget & 32K & 64K & 128K & 256K & Avg. \\
    \midrule
    Full Attention-8B  & Full & 0.181 & 0.343 & 0.667 & 1.287 & 0.620 \\
    Full Attention-70B & Full & 0.181 & 0.343 & 0.654 & 1.289 & 0.617 \\
    BinaryPC-8B        & 2048 & 0.245 & 0.243 & 0.246 & 0.242 & 0.244 \\
    BinaryPC-70B       & 2048 & 0.243 & 0.243 & 0.243 & 0.245 & 0.244 \\
    \bottomrule
  \end{tabular}
\end{table}

\subsection{Efficiency Comparison with MagicPIG}
\label{sec:Efficiency_Comparison_appendix}

% Due to kernel limitations in MagicPIG~\cite{chen2025magicpig}, which relies on GPU-CPU collaborative decoding and is incompatible with multi-GPU parallelization, we conduct a direct comparison on a single 96 GB GPU using Llama-3.1-8B-Instruct~\cite{grattafiori2024llama}.
Due to kernel limitations in MagicPIG~\cite{chen2025magicpig}, which relies on GPU-CPU collaborative decoding and is incompatible with multi-GPU parallelization, we conduct a direct comparison on a single data-center GPU using Llama-3.1-8B-Instruct~\cite{grattafiori2024llama}.
% MagicPIG employs hashing codes exceeding 1000 bits and offloads hash table lookups to the CPU, resulting in substantial communication overhead that diminishes throughput gains from sparse attention.
As shown in Table~\ref{tab:magic_efficiency}, MagicPIG consistently achieves lower throughput than the FlashAttention-2 baseline across all tested configurations, with speedup ratios ranging from 0.42$\times$ to 0.50$\times$. In contrast, BinaryPC surpasses FlashAttention-2 in all settings, achieving speedups of 1.05$\times$ to 1.69$\times$. The performance gap widens as batch size increases: at 64K context length with batch size 4, BinaryPC achieves 148.46 tokens/s compared to MagicPIG's 42.05 tokens/s, representing a 3.53$\times$ improvement over MagicPIG.
% These results demonstrate that BinaryPC's lightweight binary hashing mechanism enables practical speedups without the communication bottleneck inherent in CPU-offloading approaches.

\begin{table}[t]
\centering
% \caption{Throughput comparison (tokens/s) between MagicPIG~\cite{chen2025magicpig} and BinaryPC on a single 96 GB GPU. The gray text in brackets denotes batch size. Gain is computed relative to FlashAttention-2~\cite{dao2024flashattention}.}
\caption{Throughput comparison (tokens/s) between MagicPIG~\cite{chen2025magicpig} and BinaryPC on a single data-center GPU. The gray text in brackets denotes batch size. Gain is computed relative to FlashAttention-2~\cite{dao2024flashattention}.}

\label{tab:magic_efficiency}
\begin{tabular}{c|c|cc|cc}
\toprule
Context & FlashAttn-2.7.1 & MagicPIG & Gain & BinaryPC & Gain \\
\midrule
32K \textcolor{gray}{(1)} & 45.31 & 20.28 & 0.45$\times$ & \textbf{47.61} & 1.05$\times$ \\
32K \textcolor{gray}{(2)} & 79.77 & 39.59 & 0.50$\times$ & \textbf{91.66} & 1.15$\times$ \\
32K \textcolor{gray}{(4)} & 123.98 & 57.06 & 0.46$\times$ & \textbf{167.72} & 1.35$\times$ \\
% 32K \textcolor{gray}{(8)} & 178.05 & 90.25 & 0.51$\times$ & \textbf{295.33} & 1.66$\times$ \\
\midrule
64K \textcolor{gray}{(1)} & 38.57 & 16.38 & 0.42$\times$ & \textbf{45.95} & 1.19$\times$ \\
64K \textcolor{gray}{(2)} & 62.40 & 29.09 & 0.47$\times$ & \textbf{84.64} & 1.36$\times$ \\
64K \textcolor{gray}{(4)} & 87.62 & 42.05 & 0.48$\times$ & \textbf{148.46} & 1.69$\times$ \\
% 64K \textcolor{gray}{(8)} & OOM & 57.37 & -- & OOM & -- \\
\bottomrule
\end{tabular}
\end{table}

\subsection{Efficiency of Additional Baselines}
\label{sec:additional_efficiency_baselines}

The runtime efficiency results for the most relevant hashing-based baseline, MagicPIG~\cite{chen2025magicpig}, are reported in Table~\ref{tab:magic_efficiency}. As shown there, MagicPIG consistently achieves lower throughput than both FlashAttention-2~\cite{dao2024flashattention} and BinaryPC across all tested sequence lengths and batch sizes. In contrast, Spotlight~\cite{li2025spotlight} does not provide publicly reproducible efficiency benchmarking code. Although CUDA kernels are released, no runnable instructions or examples are provided, making a fair latency comparison difficult.

% Accordingly, we use FlashAttention-2 as the efficiency baseline for evaluating practical decoding efficiency, since it provides a strong and reproducible reference. Nevertheless, we made our best effort to further profile additional sparse-attention baselines for a more comprehensive comparison. Some methods could not be benchmarked without substantial engineering effort. The results we were able to obtain are shown in Table~\ref{tab:additional_efficiency_baselines}. We report throughput (tokens/s), which is inversely proportional to latency. All experiments use Llama-3.1-8B-Instruct on eight RTX 3090 GPUs. FlashAttention-2 is measured with StaticCache for better efficiency, while other baselines follow their official implementations. BinaryPC achieves competitive throughput while retaining the full KV cache.
Accordingly, we use FlashAttention-2 as the efficiency baseline for evaluating practical decoding efficiency, since it provides a strong and reproducible reference. Nevertheless, we made our best effort to further profile additional sparse-attention baselines for a more comprehensive comparison. Some methods could not be benchmarked without substantial engineering effort. The results we were able to obtain are shown in Table~\ref{tab:additional_efficiency_baselines}. We report throughput (tokens/s), which is inversely proportional to latency. All experiments use Llama-3.1-8B-Instruct on eight consumer-grade GPUs. FlashAttention-2 is measured with StaticCache for better efficiency, while other baselines follow their official implementations. BinaryPC achieves competitive throughput while retaining the full KV cache.

\begin{table}[t]
  \centering
  \footnotesize
  \setlength{\tabcolsep}{7pt}
  % \caption{Throughput comparison (tokens/s) of additional sparse-attention baselines on Llama-3.1-8B-Instruct using eight RTX 3090 GPUs. FlashAttention-2 is measured with StaticCache, while other baselines follow their official implementations.}
    \caption{Throughput comparison (tokens/s) of additional sparse-attention baselines on Llama-3.1-8B-Instruct using eight consumer-grade GPUs. FlashAttention-2 is measured with StaticCache, while other baselines follow their official implementations.}

  \label{tab:additional_efficiency_baselines}
  \begin{tabular}{l|c|cccc}
    \toprule
    Method & Token Budget & 16K & 32K & 64K & 128K \\
    \midrule
    FlashAttention-2 & Full KV & 30.97 & 29.94 & 26.57 & 21.31 \\
    CakeKV           & 2,048   & 28.01 & 27.92 & 24.56 & 20.47 \\
    PyramidKV        & 2,048   & 28.88 & 28.30 & 27.74 & 28.14 \\
    CompressKV       & 2,048   & 28.61 & 28.66 & 28.30 & 29.01 \\
    BinaryPC         & 2,048   & 30.13 & 28.60 & 28.12 & 27.66 \\
    \bottomrule
  \end{tabular}
\end{table}

\subsection{LongBench Evaluation Details}
\label{sec:LongBench_appendix}

Table~\ref{tab:longbench} provides comprehensive per-dataset results on LongBench~\cite{bai2024longbench} across four models: Llama-3-8B, Llama-3.1-8B-Instruct~\cite{grattafiori2024llama}, Mistral-7B-Instruct-v0.3~\cite{jiang2023mistral7b}, and Qwen2.5-7B-Instruct-1M~\cite{yang2025qwen2}. 
Across all four models, BinaryPC consistently achieves competitive or superior performance compared to baseline methods.
These results demonstrate that BinaryPC maintains strong task-level accuracy across diverse task categories and model architectures, validating its effectiveness as a general-purpose sparse attention mechanism for medium-context scenarios.

\subsection{NIAH Evaluation Details}
% NIAH 评估详情

\begin{figure*}[htbp]
% Row 1: Full Attention, PyramidKV 2048
\begin{subfigure}{0.48\textwidth}
\includegraphics[width=\linewidth]{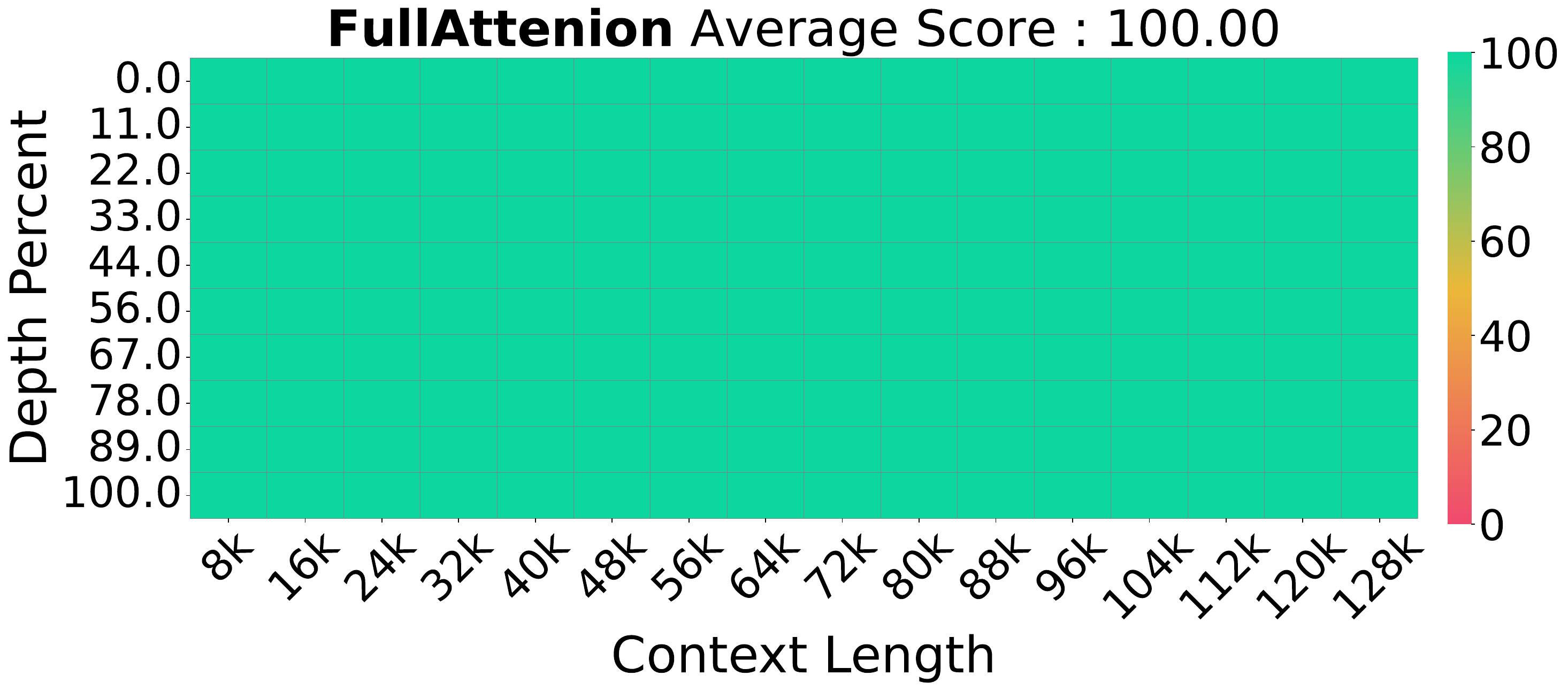}
\label{fig:appendix_Qwen_origin}
\end{subfigure}
\hfill
\begin{subfigure}{0.48\textwidth}
\includegraphics[width=\linewidth]{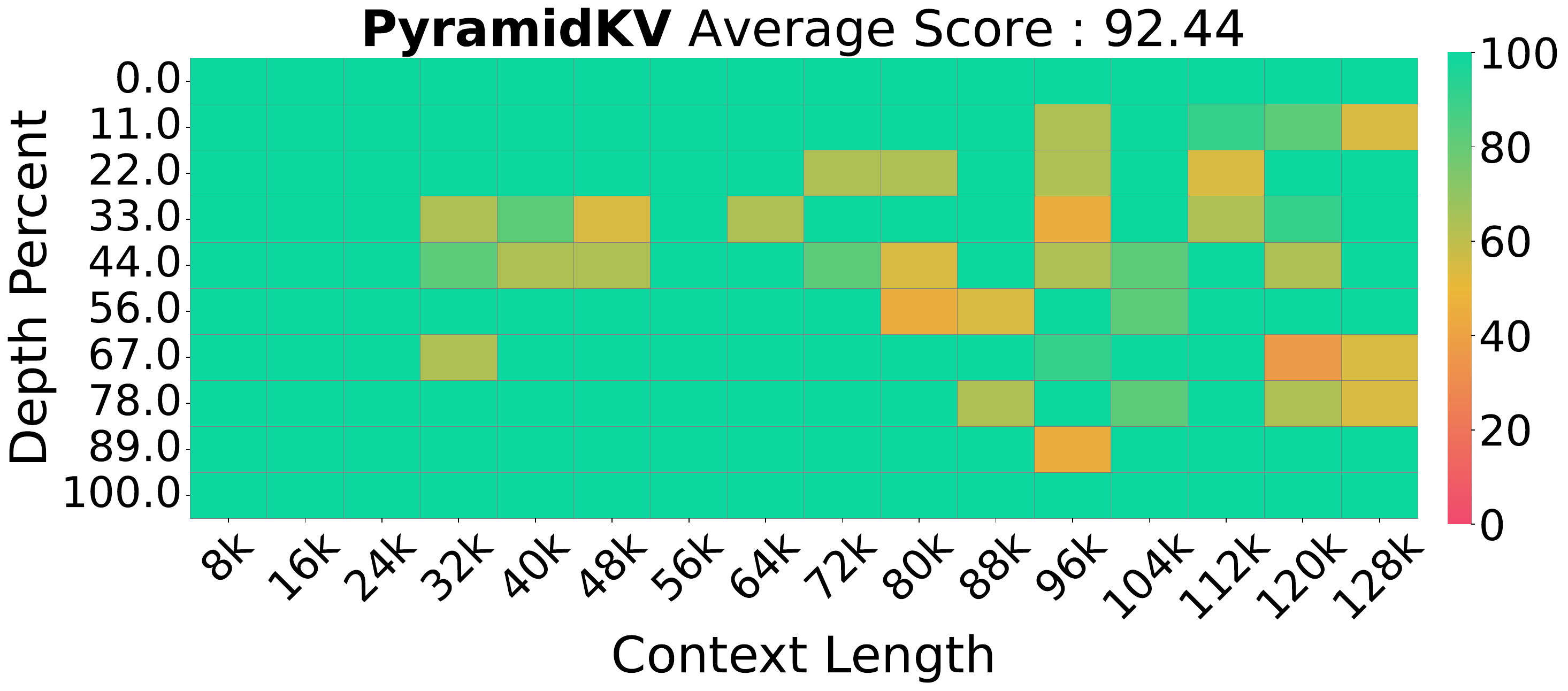}
\label{fig:appendix_Qwen_pyramidkv_2048}
\end{subfigure}
% Row 3: Quest 2048, MagicPIG
\begin{subfigure}{0.48\textwidth}
\includegraphics[width=\linewidth]{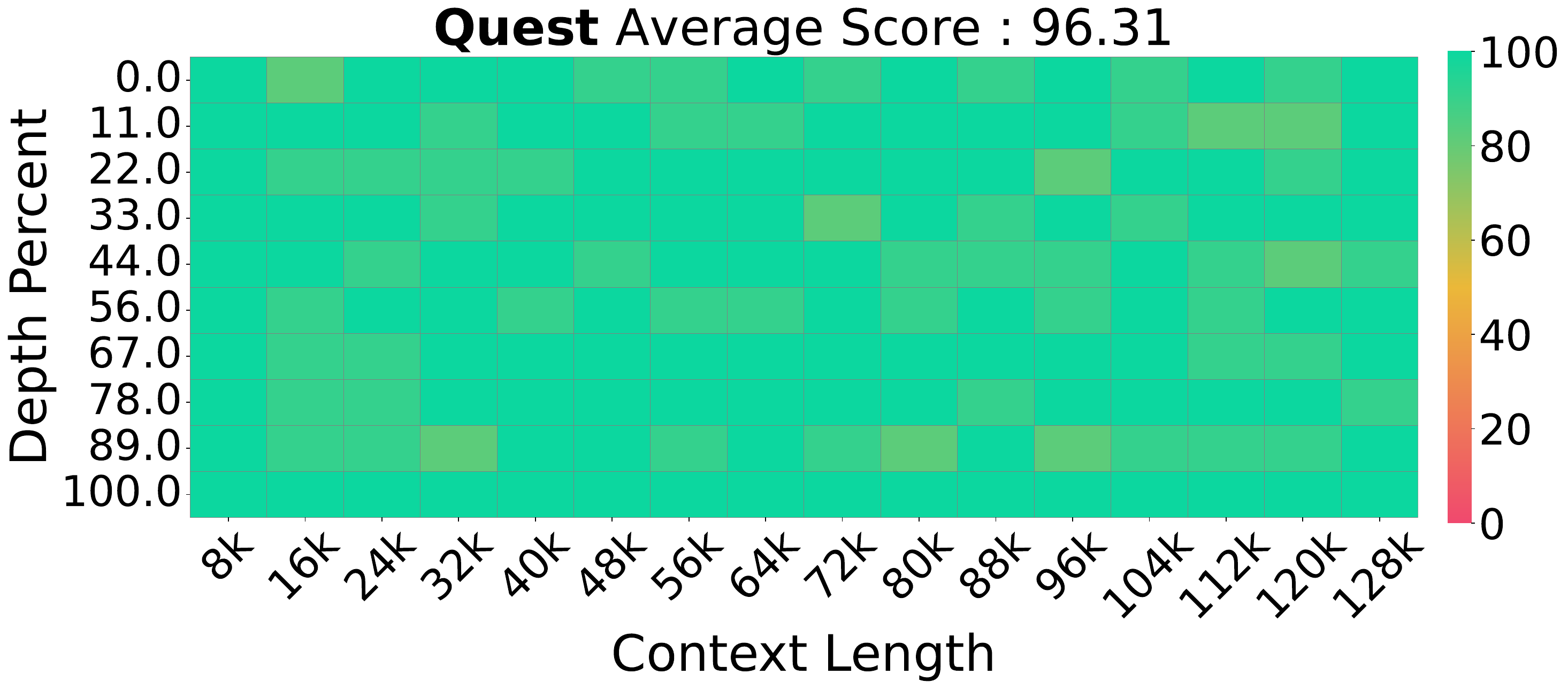}
\label{fig:appendix_Qwen_quest_2048}
\end{subfigure}
\hfill
\begin{subfigure}{0.48\textwidth}
\includegraphics[width=\linewidth]{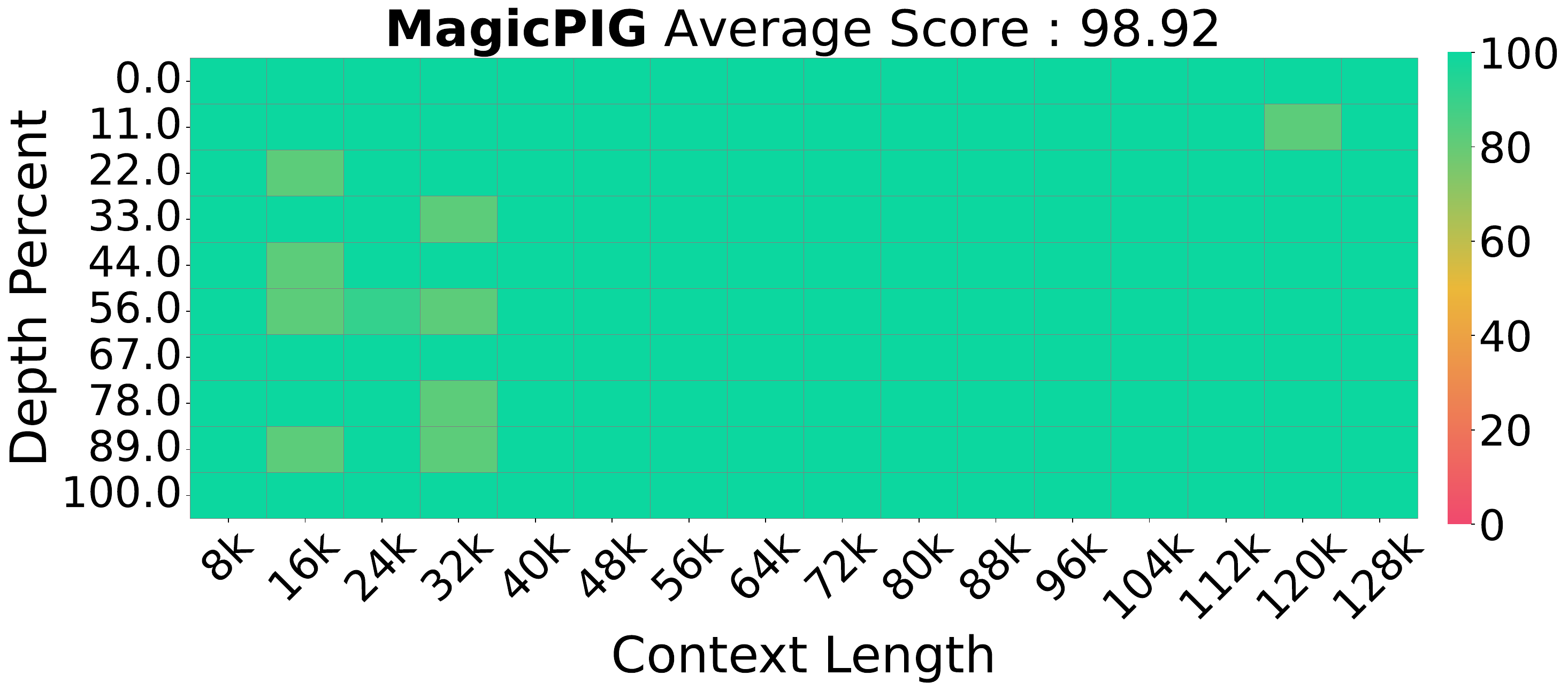}
\label{fig:appendix_Qwen_magicpig}
\end{subfigure}
% Row 4: BinVortex, BinVortex Offline
\begin{subfigure}{0.48\textwidth}
\includegraphics[width=\linewidth]{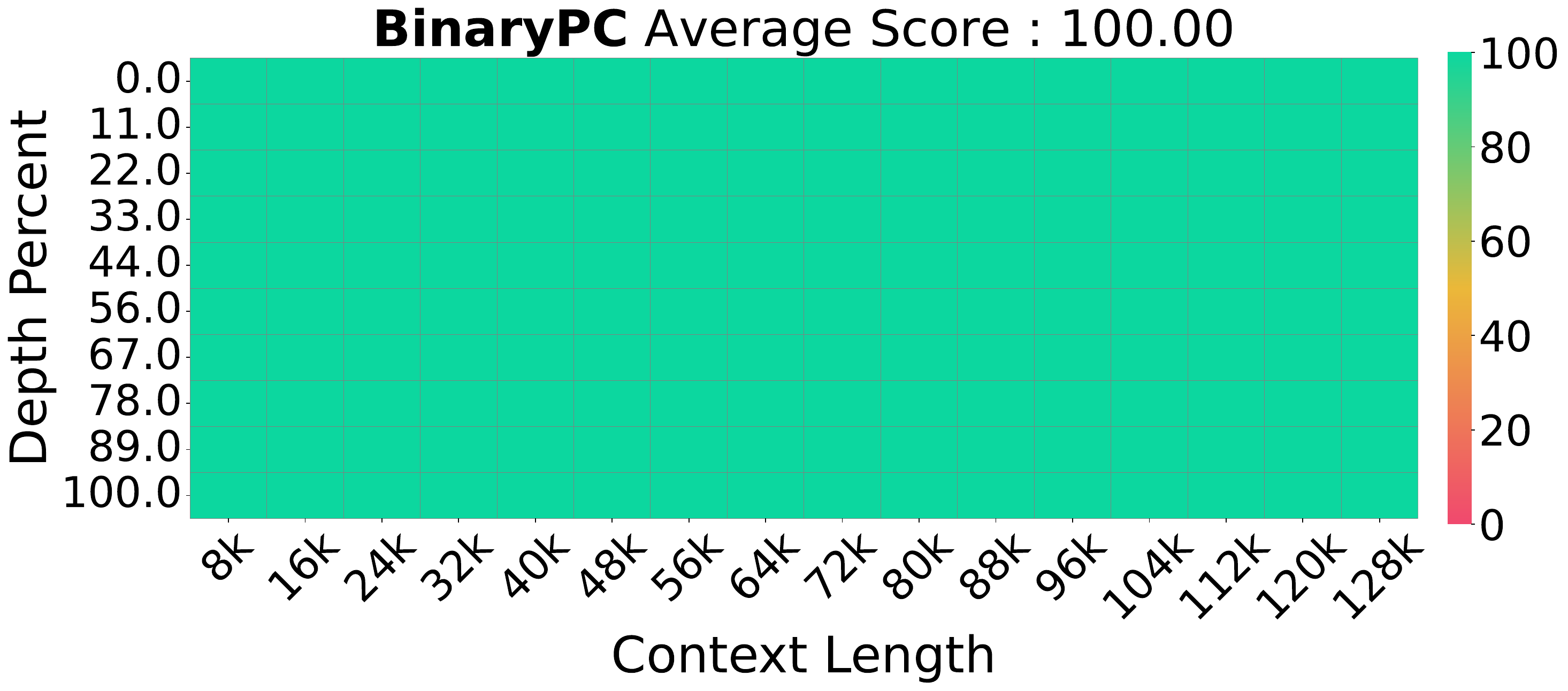}
\label{fig:appendix_Qwen_binvortex}
\end{subfigure}
\hfill
\begin{subfigure}{0.48\textwidth}
\includegraphics[width=\linewidth]{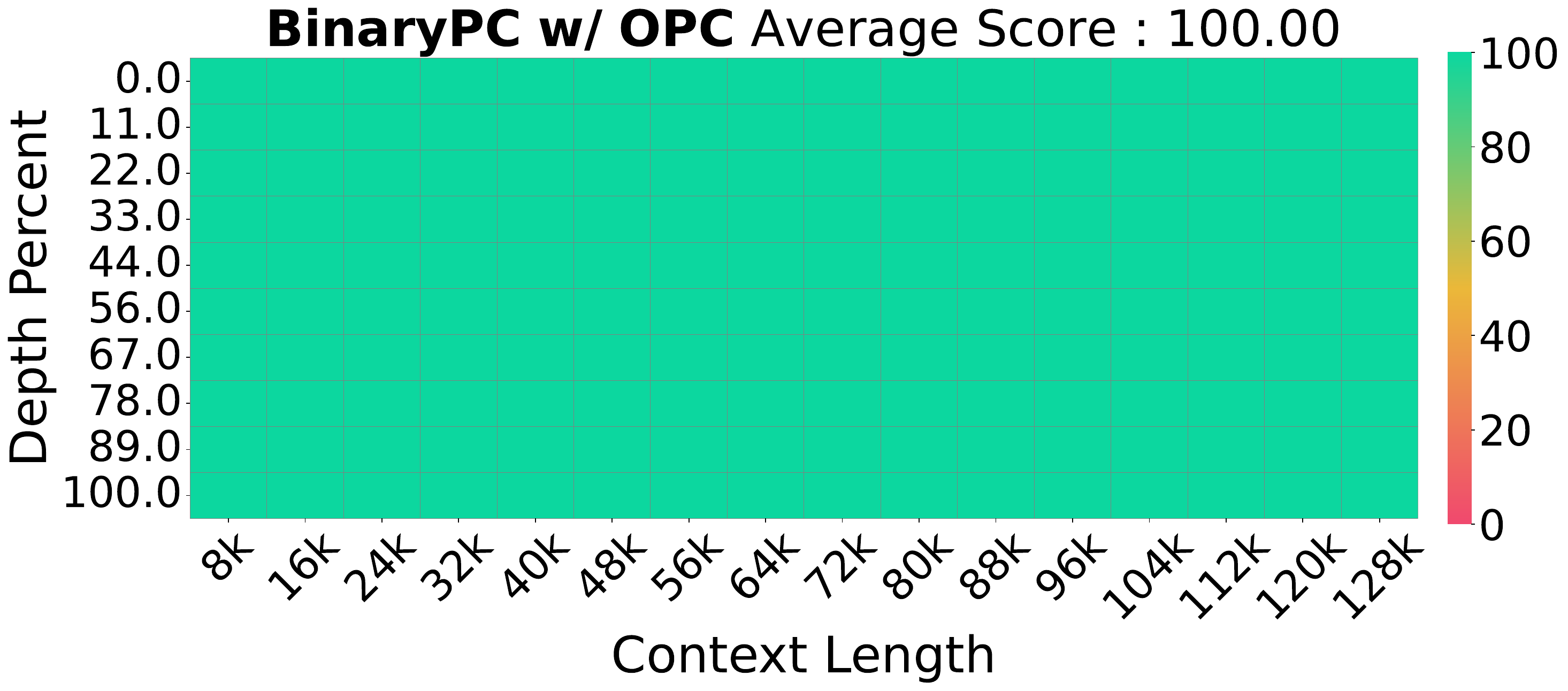}
\label{fig:appendix_Qwen_binvortex_offline}
\end{subfigure}
% \caption{NIAH~\cite{needle-in-haystack} evaluation on Qwen2.5-7B-Instruct-1M. PyramidKV and Quest use a 2K token budget; MagicPIG uses default settings.}
\caption{NIAH~\cite{needle-in-haystack} evaluation on Qwen2.5-7B-Instruct-1M. PyramidKV and Quest use a 2K token budget; MagicPIG uses default settings; BinaryPC and its offline-calibrated variant use a 2\% token budget.}

\label{fig:appendix_Qwen_images}
\end{figure*}

\begin{figure}[htbp]
% Row 1: Full Attention, PyramidKV 2048
\begin{subfigure}{0.48\textwidth}
\includegraphics[width=\linewidth]{figures/Llama-3.1-8B-Instruct/Llama-3.1-8B-Instruct_origin.pdf}
\label{fig:appendix_llama_origin}
\end{subfigure}
\hfill
\begin{subfigure}{0.48\textwidth}
\includegraphics[width=\linewidth]{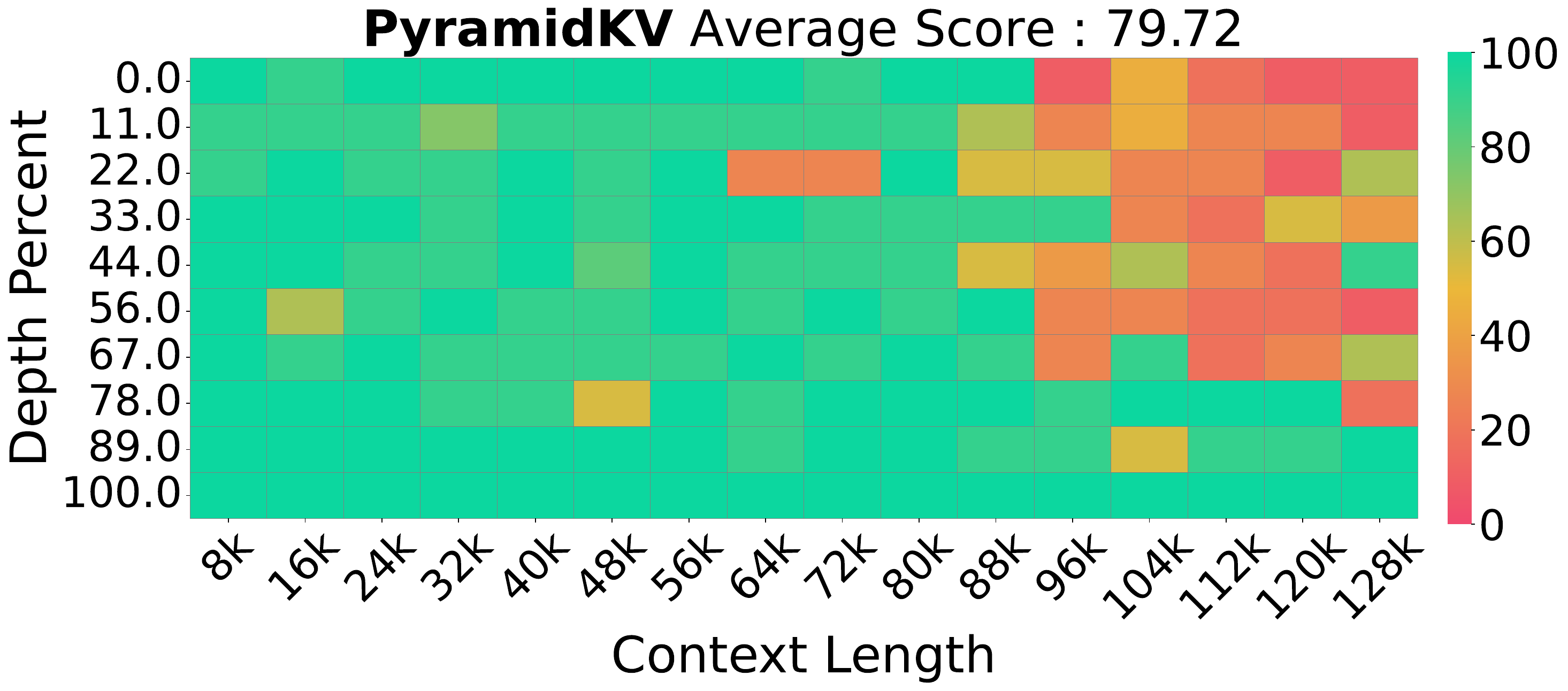}
\label{fig:appendix_llama_pyramidkv_2048}
\end{subfigure}
% Row 2: CakeKV 2048, CompressKV 2048
\begin{subfigure}{0.48\textwidth}
\includegraphics[width=\linewidth]{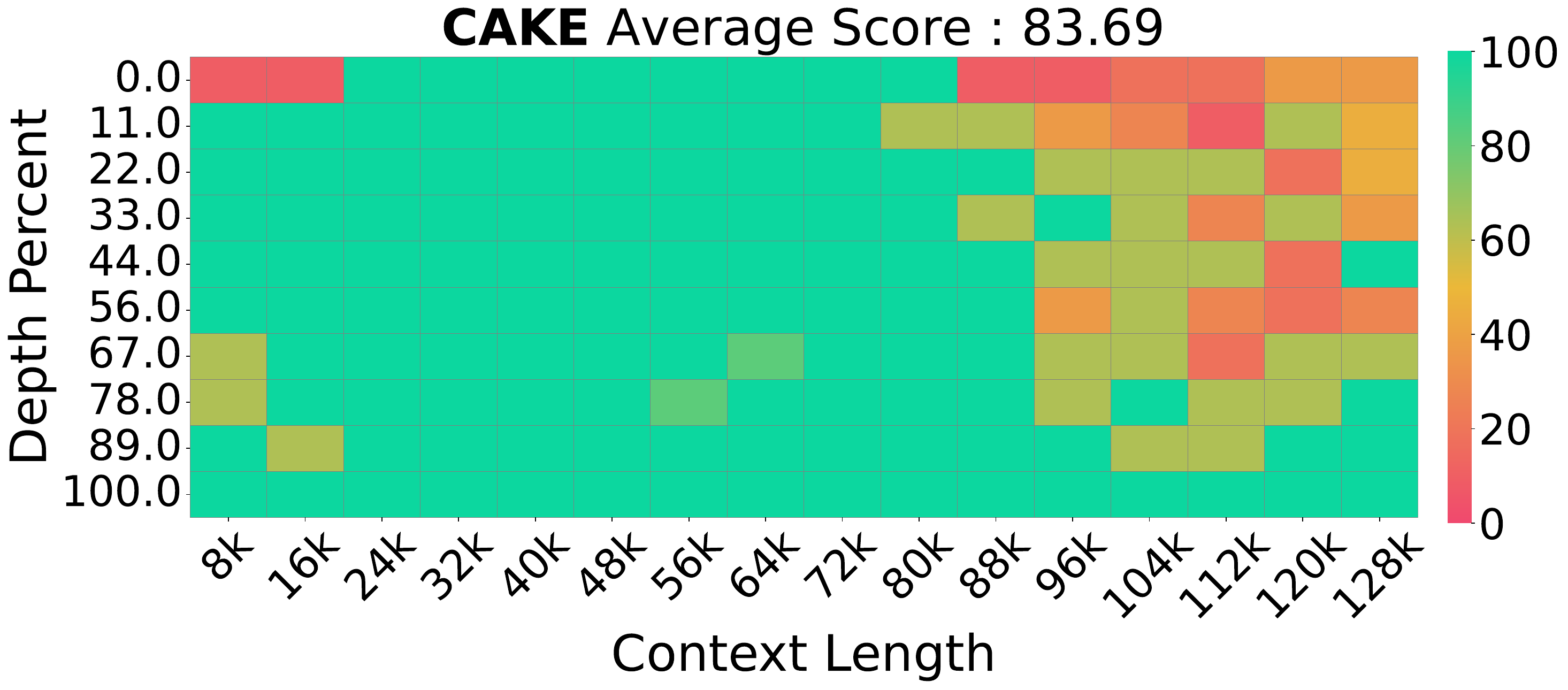}
\label{fig:appendix_llama_cakekv_2048}
\end{subfigure}
\hfill
\begin{subfigure}{0.48\textwidth}
\includegraphics[width=\linewidth]{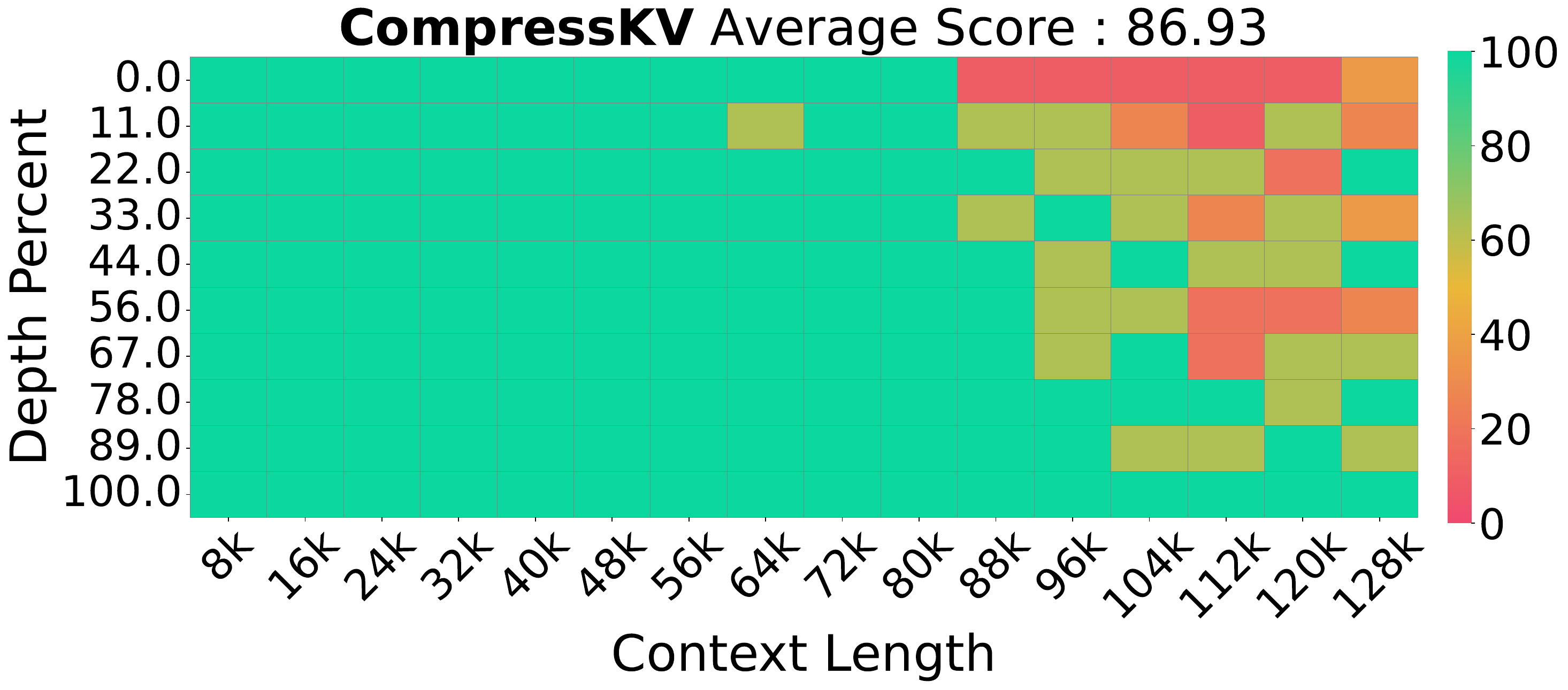}
\label{fig:appendix_llama_compresskv_2048}
\end{subfigure}
% Row 3: Quest 2048, MagicPIG
\begin{subfigure}{0.48\textwidth}
\includegraphics[width=\linewidth]{figures/Llama-3.1-8B-Instruct/Llama-3.1-8B-Instruct_quest_2048.pdf}
\label{fig:appendix_llama_quest_2048}
\end{subfigure}
\hfill
\begin{subfigure}{0.48\textwidth}
\includegraphics[width=\linewidth]{figures/Llama-3.1-8B-Instruct/Llama-3.1-8B-Instruct_magicpig.pdf}
\label{fig:appendix_llama_magicpig}
\end{subfigure}
% Row 4: BinVortex, BinVortex Offline
\begin{subfigure}{0.48\textwidth}
\includegraphics[width=\linewidth]{figures/Llama-3.1-8B-Instruct/Llama-3.1-8B-Instruct_binvortex.pdf}
\label{fig:appendix_llama_binvortex}
\end{subfigure}
\hfill
\begin{subfigure}{0.48\textwidth}
\includegraphics[width=\linewidth]{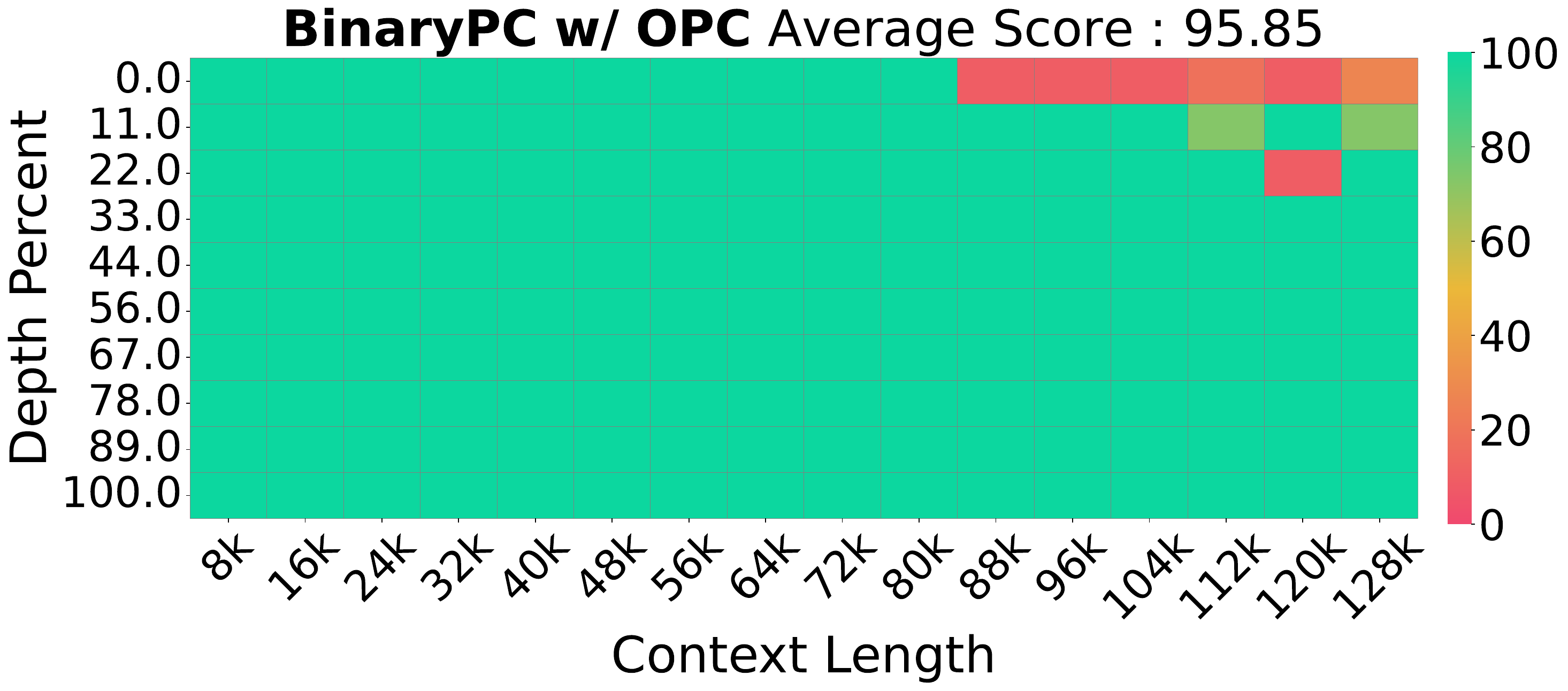}
\label{fig:appendix_llama_binvortex_offline}
\end{subfigure}
\caption{NIAH~\cite{needle-in-haystack} evaluation on Llama-3.1-8B-Instruct. PyramidKV, Cake, CompressKV, and Quest use a 2K token budget; MagicPIG uses default settings; BinaryPC and its offline-calibrated variant use a 2\% token budget.}
\label{fig:appendix_llama_images}
\end{figure}

Figure~\ref{fig:appendix_Qwen_images} and Figure~\ref{fig:appendix_llama_images} present comprehensive Needle-in-a-Haystack~\cite{needle-in-haystack} evaluations on Qwen2.5-7B-Instruct-1M~\cite{yang2025qwen2} and Llama-3.1-8B-Instruct~\cite{grattafiori2024llama}, respectively. Each heatmap visualizes retrieval accuracy as a function of needle depth (vertical axis) and context length from 8K to 128K tokens (horizontal axis). Static methods exhibit systematic failures at specific depth ranges due to their inability to adapt to query-dependent information needs. Query-aware and hashing-based methods improve upon static baselines but still show degradation at longer contexts or certain depth configurations. In contrast, BinaryPC and its offline variant achieve near-perfect retrieval across all positions and context lengths.

\subsection{RULER Evaluation Details}
\label{sec:ruler_details}

Tables~\ref{tab:ruler_8k} to~\ref{tab:ruler_128k} provide the comprehensive per-task results on the RULER~\cite{hsieh2024ruler} benchmark for context lengths ranging from 8K to 128K tokens. BinaryPC maintains the most stable performance across different context lengths and subtasks, consistently achieving competitive results with full attention compared to other sparse methods.


\begin{table}[t]
  \centering
  \caption{RULER~\cite{hsieh2024ruler} benchmark task-level evaluation at 8K context length.}
  \label{tab:ruler_8k}
  \begin{small}
      \resizebox{\columnwidth}{!}{
        \begin{tabular}{l|c|ccccccccccccc|c}
          \toprule
          Methods               &Token& N-MK1 & N-MK2 & N-MK3 & N-MQ  & N-MV  & N-S1  & N-S2  & N-S3  & CWE   & FWE   & QA-1  & QA-2  & VT    & Avg. \\
          \midrule
          \textbf{Llama-3.1-8B} &Full& 99.00 & 99.00 & 100.0 & 99.50 & 100.0 & 100.0 & 100.0 & 100.0 & 96.30 & 91.33 & 61.00 & 80.17 & 99.80 & 94.32 \\
          TOPK                  &2\%& 99.00 & 99.00 & 100.0 & 99.75 & 100.0 & 100.0 & 100.0 & 100.0 & 75.40 & 75.00 & 61.00 & 81.17 & 99.60 & 91.53 \\
          PyramidKV             &2K& 99.00 & 98.00 & 2.00  & 99.25 & 100.0 & 100.0 & 100.0 & 99.00 & 56.40 & 77.33 & 62.00 & 81.50 & 99.80 & 82.64 \\
          Cake                  &2K& 97.00 & 98.00 & 84.00 & 99.50 & 100.0 & 100.0 & 99.00 & 97.00 & 88.80 & 85.00 & 60.00 & 81.17 & 99.80 & 91.48 \\
          CompressKV            &2K& 99.00 & 98.00 & 55.00 & 99.00 & 99.75 & 100.0 & 100.0 & 100.0 & 87.70 & 83.33 & 61.00 & 81.17 & 99.80 & 89.52 \\
          Quest                 &2K& 99.00 & 98.00 & 93.00 & 99.00 & 99.75 & 100.0 & 100.0 & 98.00 & 90.60 & 86.00 & 59.00 & 81.17 & 97.20 & \textbf{92.36}\\
          MagicPIG              &Default& 98.00 & 97.00 & 93.00 & 97.75 & 95.25 & 98.00 & 95.00 & 93.00 & 94.10 & 87.00 & 61.00 & 80.50 & 97.00 & 91.28 \\
          \rowcolor{RowColor}
          BinaryPC              &2\%& 99.00 & 100.0 & 100.0 & 99.75 & 100.0 & 99.00 & 100.0 & 100.0 & 75.30 & 82.67 & 61.00 & 79.17 & 95.80 & \underline{91.67} \\
          \rowcolor{RowColor}
          BinaryPC w/ OPC       &2\%& 100.0 & 99.00 & 94.00 & 98.50 & 99.75 & 100.0 & 100.0 & 100.0 & 70.80 & 77.00 & 60.00 & 80.17 & 98.00 & 90.56 \\
          % \midrule
          % \textbf{Qwen2.5-7B}   & 100.0 & 100.0 & 99.00 & 100.0 & 99.75 & 100.0 & 100.0 & 100.0 & 89.50 & 84.00 & 56.00 & 81.17 & 99.60 & 93.00 \\
          % PyramidKV             & 100.0 & 90.00 & 14.00 & 98.75 & 98.25 & 100.0 & 100.0 & 98.00 & 32.60 & 57.00 & 54.00 & 78.42 & 98.80 & 78.45 \\
          % Quest                 & 97.00 & 94.00 & 58.00 & 99.00 & 99.50 & 100.0 & 100.0 & 90.00 & 80.80 & 78.67 & 55.00 & 80.42 & 98.20 & 86.97 \\
          % MagicPIG              & 32.00 & 27.00 & 3.00  & 20.25 & 18.75 & 27.00 & 33.00 & 0.00  & 76.10 & 75.67 & 51.00 & 68.08 & 56.60 & 37.57 \\
          % \rowcolor{RowColor}
          % BinaryPC             & 100.0 & 100.0 & 98.00 & 100.0 & 99.75 & 100.0 & 100.0 & 99.00 & 74.40 & 80.67 & 57.00 & 79.42 & 90.60 & 90.68 \\
          % \rowcolor{RowColor}
          % BinaryPC w/ OPC        & 100.0 & 100.0 & 98.00 & 100.0 & 99.75 & 100.0 & 100.0 & 100.0 & 73.20 & 81.33 & 60.00 & 76.50 & 92.40 & 90.86 \\
          \bottomrule
        \end{tabular}
      }
  \end{small}
\end{table}

\begin{table}[t]
  \centering
  \caption{RULER~\cite{hsieh2024ruler} benchmark task-level evaluation at 16K context length.}
  \label{tab:ruler_16k}
  \begin{small}
      \resizebox{\columnwidth}{!}{
        \begin{tabular}{l|c|ccccccccccccc|c}
          \toprule
          Methods               &Token& N-MK1 & N-MK2 & N-MK3 & N-MQ  & N-MV  & N-S1  & N-S2  & N-S3  & CWE   & FWE   & QA-1  & QA-2  & VT    & Avg. \\
          \midrule
          \textbf{Llama-3.1-8B} &Full& 100.0 & 100.0 & 99.00 & 99.75 & 100.0 & 100.0 & 100.0 & 100.0 & 89.80 & 94.67 & 58.00 & 80.50 & 99.80 & 93.96 \\
          TOPK                  &2\%& 100.0 & 100.0 & 99.00 & 99.00 & 99.50 & 100.0 & 100.0 & 100.0 & 81.50 & 73.67 & 56.00 & 79.50 & 99.60 & \textbf{91.37} \\
          PyramidKV             &2K& 100.0 & 97.00 & 0.00  & 99.50 & 99.00 & 100.0 & 100.0 & 84.00 & 29.00 & 88.00 & 55.00 & 80.83 & 99.80 & 79.39 \\
          Cake                  &2K& 100.0 & 100.0 & 26.00 & 98.50 & 98.50 & 100.0 & 99.00 & 95.00 & 66.90 & 91.33 & 55.00 & 80.17 & 99.40 & 85.37 \\
          CompressKV            &2K& 100.0 & 100.0 & 14.00 & 92.00 & 97.50 & 100.0 & 99.00 & 99.00 & 62.70 & 92.67 & 54.00 & 80.50 & 98.40 & 83.83 \\
          Quest                 &2K& 100.0 & 97.00 & 89.00 & 97.00 & 99.00 & 100.0 & 100.0 & 99.00 & 74.20 & 92.00 & 55.00 & 80.50 & 94.00 & 90.52 \\
          MagicPIG              &Default& 100.0 & 98.00 & 94.00 & 94.75 & 97.00 & 99.00 & 100.0 & 93.00 & 84.20 & 89.33 & 55.00 & 79.83 & 96.60 & 90.82 \\
          \rowcolor{RowColor}
          BinaryPC             &2\%& 100.0 & 100.0 & 98.00 & 99.50 & 99.50 & 96.00 & 100.0 & 100.0 & 75.30 & 84.00 & 56.00 & 80.50 & 92.40 & 90.86 \\
          \rowcolor{RowColor}
          BinaryPC w/ OPC        &2\%& 100.0 & 100.0 & 93.00 & 99.25 & 99.50 & 100.0 & 100.0 & 98.00 & 72.40 & 87.00 & 57.00 & 79.17 & 99.60 & \underline{91.15} \\
          % \midrule
          % \textbf{Qwen2.5-7B}   & 100.0 & 99.00 & 98.00 & 98.75 & 98.00 & 100.0 & 100.0 & 100.0 & 83.80 & 96.33 & 62.00 & 80.42 & 98.40 & 93.44 \\
          % PyramidKV             & 100.0 & 77.00 & 0.00  & 91.25 & 91.75 & 100.0 & 100.0 & 97.00 & 22.70 & 85.00 & 60.00 & 81.42 & 96.00 & 77.09 \\
          % Quest                 & 100.0 & 89.00 & 48.00 & 99.50 & 93.50 & 100.0 & 100.0 & 79.00 & 67.40 & 91.33 & 58.00 & 80.08 & 94.40 & 84.63 \\
          % MagicPIG              & 41.00 & 42.00 & 3.00  & 20.50 & 19.50 & 39.00 & 37.00 & 2.00  & 62.90 & 84.67 & 61.00 & 68.17 & 68.20 & 42.23 \\
          % \rowcolor{RowColor}
          % BinaryPC             & 100.0 & 98.00 & 95.00 & 100.0 & 96.75 & 100.0 & 100.0 & 99.00 & 66.20 & 92.00 & 60.00 & 79.67 & 92.00 & 90.66 \\
          % \rowcolor{RowColor}
          % BinaryPC w/ OPC        & 100.0 & 99.00 & 93.00 & 100.0 & 97.50 & 100.0 & 100.0 & 99.00 & 62.70 & 90.67 & 59.00 & 79.00 & 91.20 & 90.08 \\
          \bottomrule
        \end{tabular}
      }
  \end{small}
\end{table}

\begin{table}[t]
  \centering
  \caption{RULER~\cite{hsieh2024ruler} benchmark task-level evaluation at 32K context length.}
  \label{tab:ruler_32k}
  \begin{small}
      \resizebox{\columnwidth}{!}{
        \begin{tabular}{l|c|ccccccccccccc|c}
          \toprule
          Methods               &Token& N-MK1 & N-MK2 & N-MK3 & N-MQ  & N-MV  & N-S1  & N-S2  & N-S3  & CWE   & FWE   & QA-1  & QA-2  & VT    & Avg. \\
          \midrule
          \textbf{Llama-3.1-8B} &Full& 100.0 & 99.00 & 97.00 & 99.00 & 100.0 & 100.0 & 100.0 & 100.0 & 12.50 & 93.00 & 55.00 & 77.17 & 99.60 & 87.10 \\
          TOPK                  &2\%& 100.0 & 100.0 & 97.00 & 98.00 & 98.25 & 100.0 & 100.0 & 100.0 & 56.20 & 74.00 & 55.00 & 77.17 & 98.80 & \textbf{88.80} \\
          PyramidKV             &2K& 99.00 & 90.00 & 0.00  & 94.75 & 95.00 & 100.0 & 100.0 & 57.00 & 3.10  & 76.67 & 53.00 & 77.50 & 99.00 & 72.69 \\
          Cake                  &2K& 100.0 & 99.00 & 6.00  & 98.25 & 98.25 & 100.0 & 100.0 & 89.00 & 15.20 & 80.67 & 55.00 & 76.50 & 98.00 & 78.14 \\
          CompressKV            &2K& 100.0 & 98.00 & 2.00  & 96.00 & 96.75 & 100.0 & 100.0 & 97.00 & 25.10 & 82.00 & 55.00 & 76.50 & 96.40 & 78.83 \\
          Quest                 &2K& 99.00 & 99.00 & 63.00 & 98.25 & 98.25 & 100.0 & 100.0 & 95.00 & 15.60 & 84.67 & 56.00 & 77.75 & 93.80 & 83.10 \\
          MagicPIG              &Default& 99.00 & 96.00 & 91.00 & 98.00 & 96.25 & 100.0 & 100.0 & 98.00 & 8.50  & 93.33 & 55.00 & 75.08 & 98.20 & 85.26 \\
          \rowcolor{RowColor}
          BinaryPC             &2\%& 100.0 & 100.0 & 96.00 & 97.25 & 99.75 & 100.0 & 100.0 & 100.0 & 28.80 & 89.00 & 55.00 & 76.50 & 98.20 & \underline{87.73} \\
          \rowcolor{RowColor}
          BinaryPC w/ OPC        &2\%& 100.0 & 99.00 & 96.00 & 97.50 & 99.00 & 100.0 & 100.0 & 100.0 & 26.50 & 85.67 & 55.00 & 78.17 & 98.80 & 87.36 \\
          % \midrule
          % \textbf{Qwen2.5-7B}   & 100.0 & 100.0 & 97.00 & 100.0 & 98.25 & 100.0 & 100.0 & 100.0 & 71.30 & 97.00 & 56.00 & 79.42 & 94.40 & 91.80 \\
          % PyramidKV             & 99.00 & 82.00 & 1.00  & 81.25 & 87.25 & 100.0 & 100.0 & 97.00 & 17.40 & 77.00 & 52.00 & 76.83 & 92.00 & 74.06 \\
          % Quest                 & 89.00 & 83.00 & 34.00 & 98.25 & 94.00 & 100.0 & 91.00 & 91.00 & 56.10 & 85.67 & 54.00 & 82.75 & 89.00 & 80.60 \\
          % MagicPIG              & 52.00 & 54.00 & 14.00 & 42.25 & 37.25 & 48.00 & 48.00 & 6.00  & 54.80 & 67.33 & 51.00 & 72.00 & 53.00 & 46.13 \\
          % \rowcolor{RowColor}
          % BinaryPC             & 99.00 & 100.0 & 97.00 & 99.75 & 94.25 & 100.0 & 98.00 & 100.0 & 59.90 & 86.00 & 54.00 & 77.75 & 79.60 & 88.10 \\
          % \rowcolor{RowColor}
          % BinaryPC w/ OPC        & 99.00 & 100.0 & 95.00 & 99.50 & 95.75 & 100.0 & 100.0 & 100.0 & 56.50 & 90.33 & 53.00 & 79.75 & 84.60 & 88.73 \\
          \bottomrule
        \end{tabular}
      }
  \end{small}
\end{table}

\begin{table}[t]
  \centering
  \caption{RULER~\cite{hsieh2024ruler} benchmark task-level evaluation at 64K context length.}
  \label{tab:ruler_64k}
  \begin{small}
      \resizebox{\columnwidth}{!}{
        \begin{tabular}{l|c|ccccccccccccc|c}
          \toprule
          Methods               &Token& N-MK1 & N-MK2 & N-MK3 & N-MQ  & N-MV  & N-S1  & N-S2  & N-S3  & CWE   & FWE   & QA-1  & QA-2  & VT    & Avg. \\
          \midrule
          \textbf{Llama-3.1-8B} &Full& 100.0 & 99.00 & 98.00 & 99.75 & 98.25 & 100.0 & 100.0 & 100.0 & 0.70 & 89.33 & 52.00 & 73.83 & 96.60 & 85.19 \\
          TOPK                  &2\%& 100.0 & 99.00 & 98.00 & 99.75 & 98.00 & 100.0 & 100.0 & 100.0 & 4.20 & 66.00 & 52.00 & 74.50 & 92.60 & 83.39 \\
          PyramidKV             &2K& 97.00 & 78.00 & 0.00  & 86.75 & 85.00 & 100.0 & 99.00 & 51.00 & 0.00 & 70.00 & 51.00 & 75.83 & 94.80 & 68.34 \\
          Cake                  &2K& 100.0 & 96.00 & 2.00  & 93.25 & 88.25 & 100.0 & 98.00 & 76.00 & 0.10 & 70.67 & 50.00 & 73.83 & 91.60 & 72.28 \\
          CompressKV            &2K& 100.0 & 97.00 & 0.00  & 95.00 & 88.75 & 100.0 & 100.0 & 95.00 & 2.40 & 71.00 & 54.00 & 74.50 & 92.20 & 74.60 \\
          Quest                 &2K& 100.0 & 98.00 & 29.00 & 98.75 & 96.50 & 100.0 & 99.00 & 92.00 & 0.20 & 90.00 & 53.00 & 74.75 & 88.60 & 78.45 \\
          MagicPIG              &Default& 100.0 & 96.00 & 90.00 & 99.00 & 96.50 & 100.0 & 100.0 & 97.00 & 1.30 & 89.00 & 50.00 & 77.50 & 95.20 & 83.96 \\
          \rowcolor{RowColor}
          BinaryPC             &2\%& 100.0 & 100.0 & 95.00 & 98.50 & 98.00 & 100.0 & 100.0 & 99.00 & 1.70 & 84.33 & 53.00 & 74.50 & 92.60 & \textbf{84.36} \\
          \rowcolor{RowColor}
          BinaryPC w/ OPC        &2\%& 100.0 & 100.0 & 91.00 & 98.75 & 98.00 & 100.0 & 100.0 & 100.0 & 2.90 & 79.33 & 54.00 & 75.83 & 94.00 & \underline{84.14} \\
          % \midrule
          % \textbf{Qwen2.5-7B}   & 98.00 & 100.0 & 95.00 & 100.0 & 95.25 & 100.0 & 99.00 & 100.0 & 45.20& 85.33 & 53.00 & 77.17 & 92.20 & 87.70 \\
          % PyramidKV             & 100.0 & 71.00 & 0.00  & 83.50 & 86.25 & 100.0 & 100.0 & 65.00 & 9.60 & 65.00 & 50.00 & 75.83 & 90.40 & 68.97 \\
          % Quest                 & 95.00 & 80.00 & 12.00 & 93.75 & 87.50 & 100.0 & 98.00 & 72.00 & 34.20& 68.00 & 49.00 & 76.83 & 85.40 & 73.21 \\
          % MagicPIG              & 18.00 & 33.00 & 4.00  & 15.25 & 11.25 & 17.00 & 9.00  & 0.00  & 38.20& 70.00 & 49.00 & 64.08 & 58.80 & 29.81 \\
          % \rowcolor{RowColor}
          % BinaryPC             & 98.00 & 100.0 & 95.00 & 100.0 & 90.50 & 100.0 & 98.00 & 100.0 & 45.90& 74.33 & 51.00 & 73.75 & 72.00 & 84.50 \\
          % \rowcolor{RowColor}
          % BinaryPC w/ OPC        & 100.0 & 100.0 & 90.00 & 100.0 & 91.50 & 100.0 & 99.00 & 100.0 & 42.80& 72.67 & 50.00 & 76.83 & 79.40 & 84.78 \\
          \bottomrule
        \end{tabular}
      }
  \end{small}
\end{table}

\begin{table}[t]
  \centering
  \caption{RULER~\cite{hsieh2024ruler} benchmark task-level evaluation at 128K context length.}
  \label{tab:ruler_128k}
  \begin{small}
      \resizebox{\columnwidth}{!}{
        \begin{tabular}{l|c|ccccccccccccc|c}
          \toprule
          Methods               &Token& N-MK1 & N-MK2 & N-MK3 & N-MQ  & N-MV  & N-S1  & N-S2  & N-S3  & CWE   & FWE   & QA-1  & QA-2  & VT    & Avg. \\
          \midrule
          \textbf{Llama-3.1-8B} &Full& 97.00 & 83.00 & 70.00 & 99.00 & 94.00 & 100.0 & 100.0 & 100.0 & 0.10 & 74.67 & 42.00 & 72.58 & 62.00 & 76.49 \\
          TOPK                  &2\%& 97.00 & 82.00 & 59.00 & 99.00 & 92.50 & 100.0 & 100.0 & 100.0 & 0.10 & 55.33 & 42.00 & 72.58 & 57.60 & \textbf{73.62} \\
          PyramidKV             &2K& 96.00 & 28.00 & 0.00  & 42.75 & 33.25 & 100.0 & 98.00 & 13.00 & 0.20 & 60.33 & 42.00 & 69.17 & 48.00 & 48.52 \\
          Cake                  &2K& 95.00 & 73.00 & 0.00  & 90.75 & 75.00 & 100.0 & 100.0 & 49.00 & 0.20 & 40.33 & 43.00 & 72.58 & 62.00 & 61.61 \\
          CompressKV            &2K& 96.00 & 73.00 & 0.00  & 93.50 & 80.25 & 100.0 & 100.0 & 75.00 & 0.20 & 51.67 & 42.00 & 72.50 & 65.00 & 65.32 \\
          Quest                 &2K& 95.00 & 66.00 & 0.00  & 92.00 & 85.25 & 100.0 & 100.0 & 53.00 & 0.20 & 63.00 & 43.00 & 65.83 & 63.40 & 63.59 \\
          MagicPIG              &Default& 96.00 & 75.00 & 42.00 & 96.25 & 86.00 & 100.0 & 98.00 & 94.00 & 0.10 & 75.00 & 42.00 & 70.92 & 61.60 & 72.07 \\
          \rowcolor{RowColor}
          BinaryPC            &2\% & 97.00 & 80.00 & 45.00 & 98.75 & 92.50 & 100.0 & 100.0 & 100.0 & 0.10 & 71.00 & 43.00 & 73.58 & 51.00 & 73.23 \\
          \rowcolor{RowColor}
          BinaryPC w/ OPC        &2\%& 96.00 & 79.00 & 44.00 & 97.75 & 92.50 & 100.0 & 100.0 & 100.0 & 0.10 & 71.33 & 43.00 & 72.58 & 56.20 & \underline{73.27} \\
          % \midrule
          % \textbf{Qwen2.5-7B}   & 99.00 & 99.00 & 85.00 & 100.0 & 88.50 & 99.00 & 99.00 & 98.00 & 20.30& 83.33 & 47.00 & 68.42 & 96.20 & 83.29 \\
          % PyramidKV             & 100.0 & 75.00 & 0.00  & 55.00 & 73.75 & 99.00 & 97.00 & 36.00 & 4.60 & 59.33 & 48.00 & 66.08 & 90.40 & 61.86 \\
          % Quest                 & 96.00 & 84.00 & 7.00  & 94.25 & 76.25 & 100.0 & 97.00 & 60.00 & 20.60& 69.00 & 39.00 & 64.08 & 91.40 & 69.12 \\
          % MagicPIG              & 17.00 & 39.00 & 4.00  & 15.50 & 11.00 & 21.00 & 21.00 & 1.00  & 14.90& 56.67 & 42.00 & 56.42 & 71.60 & 28.54 \\
          % \rowcolor{RowColor}
          % BinaryPC             & 99.00 & 97.00 & 82.00 & 99.75 & 78.50 & 100.0 & 94.00 & 100.0 & 28.30& 76.67 & 50.00 & 68.75 & 87.60 & 81.66 \\
          % \rowcolor{RowColor}
          % BinaryPC w/ OPC        & 99.00 & 97.00 & 71.00 & 99.75 & 76.25 & 100.0 & 98.00 & 99.00 & 33.70& 73.00 & 48.00 & 66.75 & 87.60 & 80.70 \\
          \bottomrule
        \end{tabular}
      }
  \end{small}
\end{table}

%%%%%%%%%%%%%%%%%%%%%%%%%%%%%%%%%%%%%%%%%%%%%%%%%%%%%%%%%%%%%%%%%%%%%%%%%%%%%%%
%%%%%%%%%%%%%%%%%%%%%%%%%%%%%%%%%%%%%%%%%%%%%%%%%%%%%%%%%%%%%%%%%%%%%%%%%%%%%%%

\end{document}